\documentclass[times,referee,twocolumn,final,authoryear]{elsarticle}

\usepackage{ycviu}
\usepackage{framed,multirow}

\usepackage{amssymb}
\usepackage{latexsym}

\usepackage{mathtools}
\usepackage{animate}
\usepackage[breaklinks=true,colorlinks,urlcolor=blue,citecolor=blue,linkcolor=blue,bookmarks=false]{hyperref}
\usepackage{booktabs}
\usepackage{subcaption}
\usepackage{paralist}

\usepackage{url}
\usepackage{xcolor}
\definecolor{newcolor}{rgb}{.8,.349,.1}

\newcommand{\first}[1]{{\color{blue}\textbf{#1}}}
\newcommand{\second}[1]{{\color{red}\underline{#1}}}

\newcommand{\revised}[1]{{{#1}}}

\newcommand{\etal}{\textit{et~al}\mbox{.}}

\newcommand{\heading}[1]{\noindent\textbf{#1}}

\newlength\paramargin
\newlength\figmargin
\newlength\secmargin
\newlength\figcapmargin
\newlength\twoimg
\newlength\threeimg
\newlength\fourimg
\newlength\fiveimg
\newlength\siximg
\newlength\sevenimg
\newlength\eightimg

\newcommand{\mpage}[2]
{
\begin{minipage}{#1\textwidth}\centering
#2
\end{minipage}
}

\journal{Computer Vision and Image Understanding}

\begin{document}

\begin{frontmatter}

\title{Self-Attentive 3D Human Pose and Shape Estimation from Videos}

\author[1]{Yun-Chun \snm{Chen}} 
\author[2]{Marco \snm{Piccirilli}}
\author[3]{Robinson \snm{Piramuthu}}
\author[4]{Ming-Hsuan \snm{Yang}\corref{cor1}}
\cortext[cor1]{Corresponding author: 
  Tel.: +1-209-228-4677.}
\ead{mhyang@ucmerced.edu}

\address[1]{Department of Computer Science, University of Toronto, ON, Canada}
\address[2]{eBay Inc., San Jose, CA, USA}
\address[3]{Amazon, Oakland, CA, USA}
\address[4]{School of Engineering, University of California at Merced, CA, USA}

\received{1 May 2013}
\finalform{10 May 2013}
\accepted{13 May 2013}
\availableonline{15 May 2013}
\communicated{S. Sarkar}

\begin{abstract}
We consider the task of estimating 3D human pose and shape from videos. 
While existing frame-based approaches have made significant progress, these methods are independently applied to each image, thereby often leading to inconsistent predictions.
In this work, we present a video-based learning algorithm for 3D human pose and shape estimation.
The key insights of our method are two-fold.
First, to address the inconsistent temporal prediction issue, we exploit temporal information in videos and propose a self-attention module that jointly considers short-range and long-range dependencies across frames, resulting in temporally coherent estimations.
Second, we model human motion with a forecasting module that allows the transition between adjacent frames to be smooth.
We evaluate our method on the 3DPW, MPI-INF-3DHP, and Human3.6M datasets. 
Extensive experimental results show that our algorithm performs favorably against the state-of-the-art methods.
\end{abstract}


\end{frontmatter}

%\linenumbers

\section{Introduction}

3D human pose and shape estimation~\citep{HMR,SPIN,SMPLify} is an active research topic in computer vision and computer graphics that finds numerous applications~\citep{xu2019mo,liu2018neural}. 
The inherent under-constrained nature where multiple 3D meshes can explain the same 2D projection makes this problem very challenging.
While frame-based methods~\citep{HMR,SPIN,SMPLify} and video-based approaches~\citep{VIBE,lee2018propagating,rayat2018exploiting,TemporalHMR,PHD} have been developed to recover human pose in the literature, numerous issues remain to be addressed. 
First, existing approaches employ recurrent neural networks (RNNs) to model temporal information for consistent predictions.
However, it is difficult to train RNNs to capture long-range dependencies~\citep{transformer,pascanu2013difficulty}.
On the other hand, one recent approach employing RNNs does not consistently render smooth predictions across frames~\citep{VIBE}.
\begin{figure}[t]
  \begin{center}
  \animategraphics[autoplay,loop,width=\linewidth]{30}{gif/}{000000}{000234}
  \vspace{-5.5mm}
  \caption{
  \textbf{3D human pose and shape estimation.} 
  The results are generated by our method without prior information of camera or manual initialization. The embedded video can be viewed using Adobe Acrobat.
  }
  \label{fig:teaser}
  \end{center}
  \vspace{-8.0mm}
\end{figure}
Second, as most real-world datasets do not contain ground-truth camera parameter annotations, existing methods typically reproject the predicted 3D joints onto the 2D space using the estimated camera parameters, followed by a loss enforced between the reprojected 2D joints and the corresponding ground-truth 2D joints.
Nevertheless, such regularization terms are still insufficient to account for complex scenes. 
Third, existing methods~\citep{VIBE,TemporalHMR,PHD,HMR} do not perform well for humans under heavy occlusion or out-of-view, as there is no explicit constraint enforced on the invisible regions.

In this paper, we propose the Self-attentive Pose and Shape Network (SPS-Net) for 3D human pose and shape estimation from videos.
Our key insights are two-fold.
First, motivated by the attention models in neural machine translation~\citep{transformer} and image generation~\citep{SAGAN} tasks, we develop a self-attention module to exploit temporal cues in videos for coherent predictions.
For each input frame, our self-attention module derives a visual representation by observing past and future frames and predicting the associated attention weights.
Second, motivated by the autoregressive models in human motion prediction~\citep{TemporalHMR,PHD}, we develop a forecasting module that leverages visual cues from human motion to encourage our model to generate temporally smooth predictions.
By jointly considering both features, our SPS-Net is able to estimate accurate and temporally coherent human pose and shape (see Figure~\ref{fig:teaser}).

\revised{In addition to coherent and smooth predictions, we also address the issues with ground-truth camera parameters and heavy occlusion for robust 3D human pose and shape estimation from videos.}
To account for images without ground-truth camera parameter annotations, we exploit the property that the camera parameters for the overlapped frames of two segments from the same video should be the same. 
We enforce this constraint with a camera parameter consistency loss.
Furthermore, we address the occlusion and out-of-view issues by masking out some regions of the video frames.
Our core idea is to leverage the predictions of the original video frames to \emph{supervise} those of the synthesized occluded or partially visible data, making our model more robust to the occlusion and out-of-view issues.
We demonstrate the effectiveness of the proposed SPS-Net on three standard benchmarks, including the 3DPW~\citep{3DPW}, MPI-INF-3DHP~\citep{MPII}, and Human3.6M~\citep{human36m} datasets.

Our main contributions can be summarized as follows:

\begin{compactitem}
  \item We present a video-based learning algorithm for 3D human pose and shape estimation.
  \item We propose a camera parameter consistency loss that provides additional supervisory signals for model training, resulting in more accurate camera parameter predictions.
  \item Our model learns to predict plausible estimations when occlusion or out-of-view occurs in a self-supervised fashion.
  \item Extensive evaluations on three challenging benchmarks demonstrate that our method achieves the state-of-the-art performance against existing approaches.
\end{compactitem}

\vspace{-3.5mm}
\section{Related Work}
\vspace{-1.5mm}

\heading{3D human pose and shape estimation.}
Existing methods for 3D human pose and shape estimation can be broadly categorized as frame-based and video-based.
Frame-based methods typically use an off-the-shelf keypoint detector (e.g., DeepCut~\citep{DeepCut}) to fit the SMPL~\citep{SMPL} body model~\citep{SMPLify}, leverage silhouettes and keypoints for model fitting~\citep{lassner2017unite}, or directly regress the parameters for the SMPL~\citep{SMPL} body model from pixels using neural networks~\citep{SPIN,HMR,CMR}.
While these frame-based approaches are able to recover 3D poses from a single image, independently applying these algorithms to each video frame often leads to temporally inconsistent predictions.
Video-based methods, on the other hand, usually adopt RNN-based models to generate temporally coherent predictions.
These approaches either focus on estimating the human body of the current frame~\citep{Temporal3DKinetics,sun2019human,VIBE} or predicting the past and future motions~\citep{TemporalHMR,PHD}.

Our algorithm differs from these video-based methods in three aspects.
First, in contrast to adopting RNN-based models, we develop a self-attention module to aggregate temporal information and a forecasting module to model human motion for predicting temporally coherent estimations.
Second, we enforce a consistency loss on the prediction of camera parameters to regularize model learning.
Third, we address the occlusion and out-of-view issues with a self-supervised learning scheme to generate plausible human pose and shape predictions.

\heading{Attention models.}
Attention models have been shown effective in neural machine translation~\citep{transformer} and image generation problems~\citep{SAGAN,parmar2018image}.
For machine translation, employing self-attention models~\citep{transformer} helps capture short-range and long-range correlations between tokens in the sentence for improving the translation quality.
In image generation, the Image Transformer~\citep{parmar2018image} and SAGAN~\citep{SAGAN} show that leveraging self-attention mechanisms facilitates the models to generate realistic images.
In 3D human pose and shape estimation, the VIBE~\citep{VIBE} method adopts a self-attention scheme in the discriminator for feature aggregation, allowing the discriminator to better distinguish the motions of attended video frames between the real sequences and generated ones.

We adopt self-attention modules in both the SPS-Net and discriminator.
Our method differs from the VIBE~\citep{VIBE} in that our self-attention module aims to derive a representation for each frame that contains temporal information by jointly considering short-range and long-range dependencies across video frames, whereas the VIBE~\citep{VIBE} method aims to derive a single representation for the entire pose sequence.

\heading{Future human pose predictions.}
Predicting future poses from videos has been studied by a few approaches in the literature.
Existing algorithms estimate 2D poses from pixels~\citep{denton2017unsupervised,finn2016unsupervised}, optical flow~\citep{walker2016uncertain}, or 2D poses~\citep{walker2017pose}, or predict 3D outputs based on 3D inputs~\citep{butepage2017deep,fragkiadaki2015recurrent,jain2016structural,li2017auto,villegas2018neural}.
Other approaches learn 3D pose prediction from 2D inputs~\citep{PHD,TemporalHMR}.

Similar to the HMMR~\citep{TemporalHMR} and PHD~\citep{PHD} methods, we leverage visual cues from human motion to predict temporally smooth predictions.
Our method differs from them in that our self-attention module helps capture short-range and long-range dependencies across video frames in the input video, while the 1D convolution in the temporal encoder and autoregressive module of these methods does not have such ability.

\begin{figure*}[t]
  \begin{center}
    \includegraphics[width=1.0\linewidth]{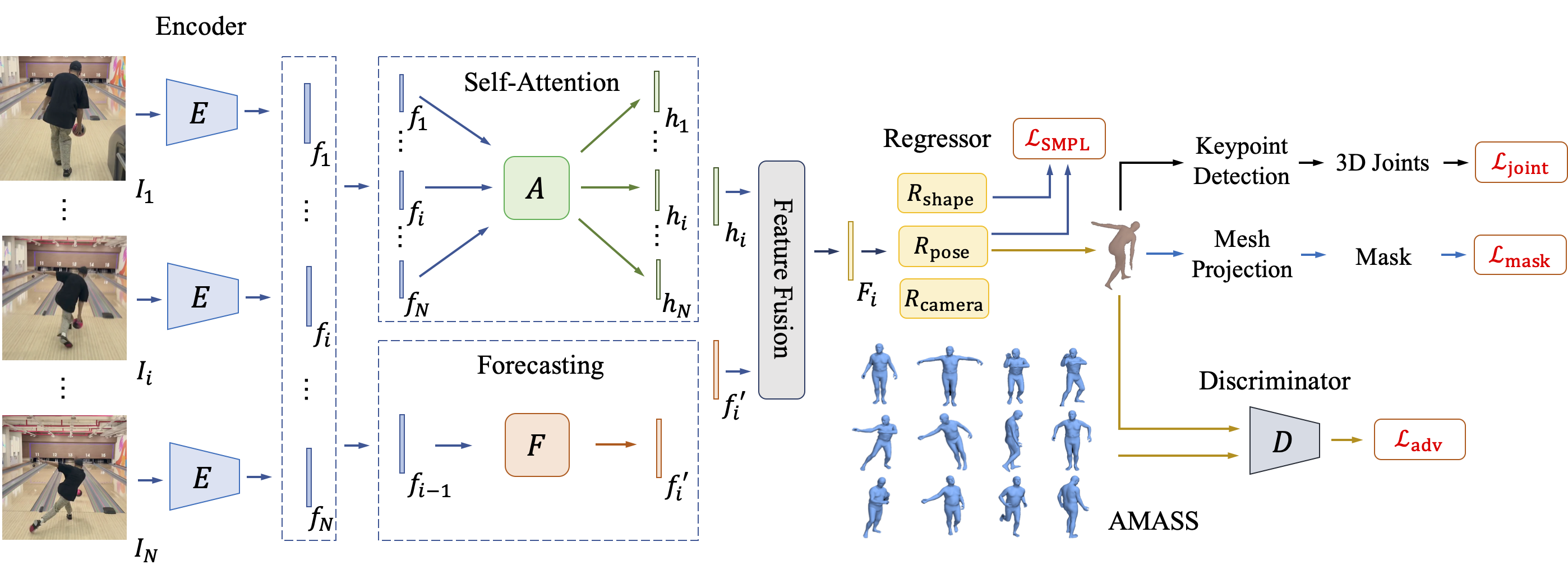}
  \end{center}
  \vspace{-6.0mm}
  \caption{
  \textbf{Overview of the Self-attentive Pose and Shape Network (SPS-Net).} 
  Our SPS-Net is composed of four main components: a feature encoder $E$ (highlighted in blue), a self-attention module $A$ (highlighted in green), a forecasting module $F$ (highlighted in orange), and three parameter regressors $R_\mathrm{shape}$, $R_\mathrm{pose}$, and $R_\mathrm{camera}$ (highlighted in yellow).
  The feature encoder extracts features from input video frames.
  The encoded features are then passed to the self-attention module to produce latent representations that contain temporal information of past and future frames and to the forecasting module to predict the features of the next time step.
  The latent representations and the predicted features of the same time step are forwarded to the feature fusion module for feature aggregation.
  Finally, the fused representations are passed to three parameter regressors to predict the corresponding shape, pose, and camera parameters, respectively.
  }
  \label{fig:model}
  \vspace{-5.0mm}
\end{figure*}

\heading{Consistency constraints for visual learning.} 
Exploiting consistency constraints to regularize model learning has been shown effective in numerous applications, including semantic matching~\citep{FlowWeb}, optical flow estimation~\citep{UnFlow}, depth prediction~\citep{gordon2019depth}, and image-to-image translation~\citep{zhu2017unpaired,DRIT,MUNIT}.
Other methods exploit consistency constraints across multiple network outputs, including depth and optical flow estimation~\citep{zou2018df}, joint semantic matching and object co-segmentation~\citep{chen2020show}, ego-motion~\citep{zhou2017unsupervised}, and domain adaptation~\citep{CrDoCo}.
In our work, we show that enforcing consistency constraints on the prediction of camera parameters for the overlapped video frames of two segments from the same video results in performance improvement.

\vspace{-3.5mm}
\section{Proposed Algorithm}

In this section, we first provide an overview of our approach.
Next, we describe the details of the self-attention and forecasting modules, followed by formulating the proposed camera parameter consistency loss.
We then motivate the self-supervised learning scheme for addressing the occlusion and out-of-view issues.

\vspace{-1.5mm}
\subsection{Algorithmic overview}

Given an input video $V = \{I_i\}_{i=1}^N$ of length $N$ containing a single person, our goal is to learn a model that recovers the 3D human body of each frame.
We present the Self-attentive Pose and Shape Network (SPS-Net), comprising four components: 1) feature encoder $E$, 2) self-attention module $A$, 3) forecasting module $F$, and 4) three parameter regressors $R_\mathrm{shape}$, $R_\mathrm{pose}$, and $R_\mathrm{camera}$.

As shown in Figure~\ref{fig:model}, we first apply the encoder $E$ to each frame $I_i \in V$ to extract the feature $f_i = E(I_i) \in \mathbb{R}^{d}$, where $d$ denotes the number of channels of the feature $f_i$.
Next, the self-attention module $A$ takes all the encoded features $\{f_i\}_{i=1}^N$ as input and outputs the corresponding latent representations $\{h_i\}_{i=1}^N$, where $h_i \in \mathbb{R}^{d}$ denotes the latent representation for $I_i$, containing temporal information of past and future frames.
The forecasting module $F$ takes each encoded feature $f_i$ as input and forecasts the feature of the next time step $f_{i+1}' = F(f_i) \in \mathbb{R}^{d}$.
The latent representations $\{h_i\}_{i=1}^N$ and the predicted features $\{f_{i+1}'\}_{i=1}^{N}$ of the same time step (e.g., $h_i$ and $f_i'$) are passed to a feature fusion module to derive the fused representations $\{F_i\}_{i=1}^N$, where $F_i \in \mathbb{R}^{d}$ contains both global temporal and local motion information. 
The pose parameter regressor $R_\mathrm{pose}$ takes each fused representation $F_i$ as input and renders the pose parameters $\theta_i$ for each frame $I_i$, where $\theta_i = R_\mathrm{pose}(F_i) \in \mathbb{R}^{72}$.
The shape parameter regressor $R_\mathrm{shape}$, on the other hand, takes all the fused representations $\{F_i\}_{i=1}^N$ as input and regresses the shape parameters $\beta \in \mathbb{R}^{10}$ of the input video $V$.

\heading{3D human body representation.}
Similar to the state-of-the-art methods~\citep{HMR,SPIN,VIBE}, we adopt the SMPL~\citep{SMPL} body model to describe the human body using a 3D mesh representation.
The SMPL~\citep{SMPL} model is described by the pose $\theta \in \mathbb{R}^{72}$ and shape $\beta \in \mathbb{R}^{10}$ parameters.
The pose parameters $\theta$ contain the global body rotation and the relative 3D rotation of $23$ joints in axis-angle format.
The shape parameters $\beta$ are parameterized by the first $10$ linear coefficients of a PCA shape space.
We use a gender-neutral shape model as in previous work~\citep{HMR,SPIN,VIBE}.
The differentiable SMPL~\citep{SMPL} body model takes the pose $\theta$ and shape $\beta$ parameters as input and outputs a triangular mesh $M(\theta, \beta) \in \mathbb{R}^{6890 \times 3}$ consisting of $6,890$ mesh vertices by shaping a template body mesh based on forward kinematics.
The 3D keypoints $X \in \mathbb{R}^{k \times 3}$ of $k$ body joints can be obtained by applying a pre-trained linear regressor $W$ to the 3D mesh $M(\theta, \beta)$, and is defined as $X = W M(\theta, \beta)$.

\begin{figure*}[h]
  \begin{center}
    \includegraphics[width=0.9\linewidth]{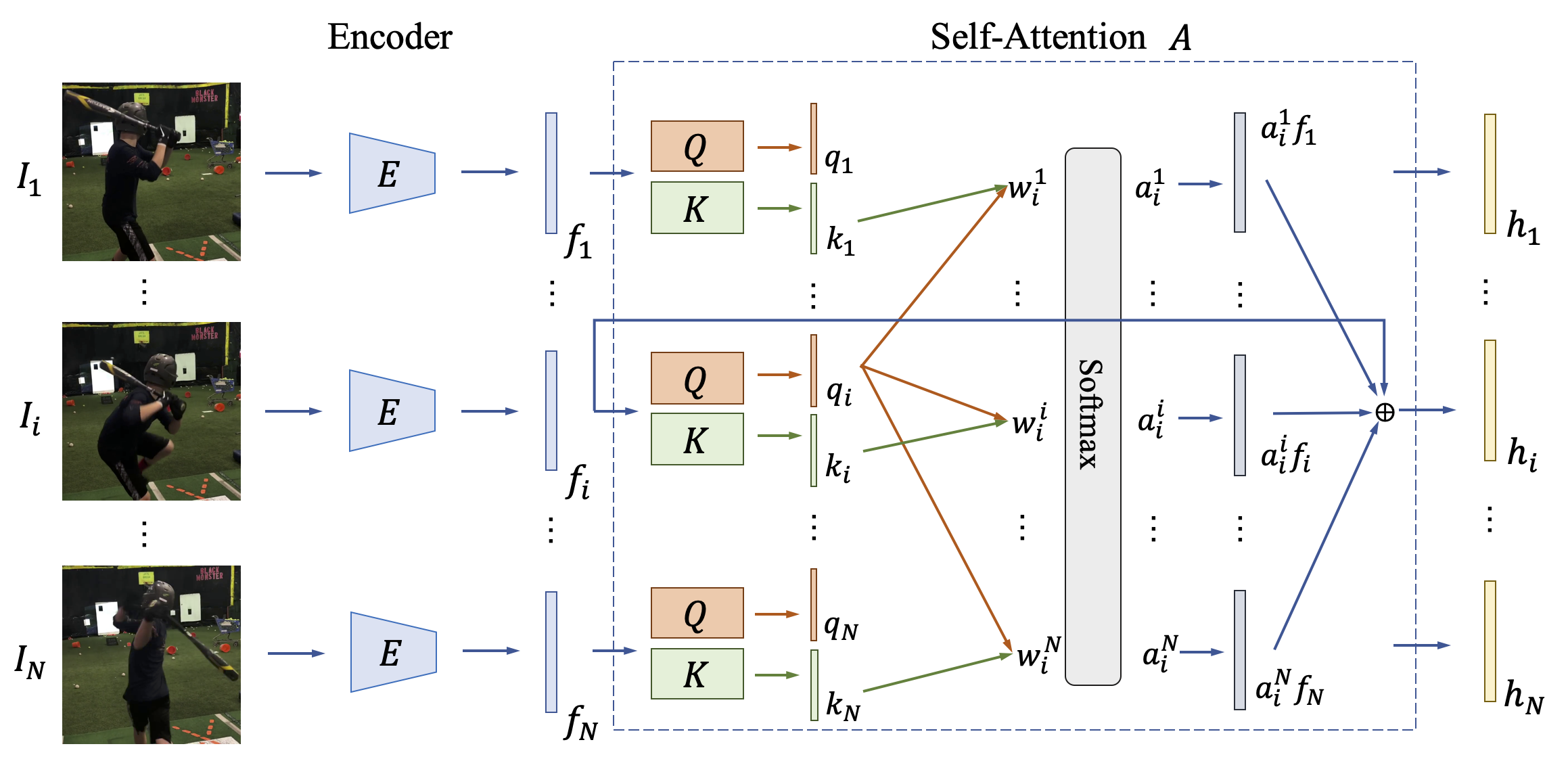}
  \end{center}
  \vspace{-6.0mm}
  \caption{
  \textbf{Overview of the self-attention module $A$.}
  \revised{
  Our self-attention module $A$ is composed of an attention network $Q$ and an attention network $K$.
  Given a sequence of input features, the self-attention module first predicts the attention vector $q$ and the attention vector $k$ for each frame.
  Next, we compute the inner product between each attention vector $q$ and all attention vectors $\{k_j\}_{j=1}^N$, followed by normalizing the weights using a softmax function.
  The input features are first fused using the associated weights and then summed with the skipped input features to derive the final latent representations as output.
  }
  }
  \label{fig:self-attention}
  \vspace{-4.0mm}
\end{figure*}

\heading{Camera model.}
Similar to existing approaches~\citep{HMR,SPIN,VIBE}, we use a weak-perspective camera model in this work.
By estimating the camera parameters $\{s, R, t\}$ using the regressor $R_\mathrm{camera}$, where $s \in \mathbb{R}$ denotes the scale, $R \in \mathbb{R}^{3 \times 3}$ is the global rotation in axis-angle format, and $t \in \mathbb{R}^{2}$ denotes the translation, the 2D projection $x \in \mathbb{R}^{k \times 2}$ of the 3D keypoints $X$ can be obtained by $x = s \Pi (RX(\theta, \beta)) + t$, where $\Pi$ is an orthographic projection. 

\vspace{-2.5mm}
\subsection{Self-attention module}

Given a sequence of features $\{f_i\}_{i=1}^{N}$ encoded by the encoder $E$, our goal is to leverage temporal cues in the input video to provide more information that helps regularize the estimation of human pose and shape.
Existing methods exploit temporal information by resorting to an RNN-based model, e.g., GRU~\citep{VIBE} or LSTM~\citep{lee2018propagating,rayat2018exploiting}.
However, training RNN-based models is difficult to capture long-range dependencies~\citep{transformer,pascanu2013difficulty}.

Motivated by the attention models~\citep{transformer,SAGAN,parmar2018image} which have been shown effective to jointly capture short-range and long-range dependencies while being more parallelizable to train~\citep{transformer}, we develop a self-attention module to learn latent representations $h$ that jointly observe past and future video frames for producing temporally consistent pose and shape predictions.
\revised{
In this work, we aim to exploit the idea that the occluded frames can benefit from the information of the non-occluded frames, while the non-occluded frames do not have to depend on the information from the occluded frames (i.e., anti-symmetric attention of humans that can be occluded and un-occluded between frames in either direction).
To achieve this, we have an attention network $Q$ and an attention network $K$ in our self-attention module $A$.
}

\revised{
As shown in Figure~3, for each feature $f_i$, we first apply the attention network $Q$ and the attention network $K$ to encode an attention vector $q_i = Q(f_i) \in \mathbb{R}^{d}$ and an attention vector $k_i = K(f_i) \in \mathbb{R}^{d}$, respectively.
To consider the dependency between two input frames $I_i$ and $I_j$, we compute the inner product between the attention vector $q_i$ of frame $I_i$ and the attention vector $k_j$ of frame $I_j$, i.e., $w_i^j = q_i \cdot k_j \in \mathbb{R}$.
To derive the latent representation $h_i$ for frame $I_i$, we first apply a softmax layer to all the weights $\{w_i^l\}_{l=1}^{N}$ computed between the attention vector $q_i$ of frame $I_i$ and all attention vectors $\{k_l\}_{l=1}^N$ for normalization to derive the attention weights.
}
The attention weights $\{a_i^l\}_{l=1}^N$ are computed by
\vspace{-1.5mm}
\begin{equation}
  a_i^l = \frac{\exp(w_i^l)}{\sum_{n=1}^{N}\exp(w_i^n)}.
  \label{eq:attention-weight}
  \vspace{-1.0mm}
\end{equation}
We then apply a weighted sum layer to sum over all input features $\{f_l\}_{l=1}^N$ with the associated attention weights $\{a_i^l\}_{l=1}^N$.
In addition, we add a residual connection~\citep{ResNet} to pass the input feature $f_i$ to the output of the self-attention module.
Specifically, the latent representation $h_i$ is described by
\vspace{-1.5mm}
\begin{equation}
  h_i = f_i + \sum_{l=1}^{N} a_i^l \cdot f_l.
  \label{eq:latent-feature}
  \vspace{-1.0mm}
\end{equation}

\vspace{-6.0mm}
\subsection{Forecasting module}

In addition to considering global temporal information as in the self-attention module $A$, we exploit visual cues from human motion to encourage our model to generate temporally smooth predictions.
Motivated by methods that focus on tackling human motion prediction~\citep{TemporalHMR,PHD}, we develop a forecasting module $F$ that takes each encoded feature $f_i$ as input and forecasts the feature of the next time step $f_{i+1}'$.
As the feature of the next time step is available (given by the encoder), we train the forecasting module $F$ in a self-supervised fashion with a feature regression loss:
\vspace{-1.5mm}
\begin{equation}
  \mathcal{L}_\mathrm{feature} = \sum_{i=1}^{N-1}\|f_{i+1} - f_{i+1}' \|_2.
  \label{eq:feature-loss}
  \vspace{-1.0mm}
\end{equation}
We note that since the feature of the next time step of $f_N$ is not available, we do not compute the feature regression loss on $f_{N+1}'$.
\revised{
The feature regression loss $\mathcal{L}_\mathrm{feature}$ allows the forecasting module $F$ to forecast the feature of the next time step for each input feature by exploiting visual cues from human motion that provide more temporal context for generating temporally smooth predictions.
}

% and allows our model to predict features that make the local transition between adjacent video frames smooth.

\vspace{-2.5mm}
\subsection{3D human pose and shape estimation}

To jointly consider the latent representations $\{h_i\}_{i=1}^N$ that contain global temporal information and the predicted features $\{f_{i+1}'\}_{i=1}^{N-1}$ that contain local motion information for predicting the parameters for 3D human pose and shape estimation, we have a feature fusion module that fuses $\{h_i\}_{i=1}^N$ and $\{f_{i+1}'\}_{i=1}^{N-1}$ at the same time step to derive the fused representations $\{F_i\}_{i=1}^{N}$.
We note that since our encoder $E$ is pre-trained on single-image pose and shape estimation task and fixed during training as in prior work~\citep{HMR,VIBE}, the feature $f_i$ encoded by the encoder $E$ is \emph{static} and does not contain motion information.
Therefore, we use the predicted feature $f_i'$ from the forecasting module $F$ that contains motion information for feature fusion.

\begin{figure}[t]
  \begin{center}
    \includegraphics[width=1.0\linewidth]{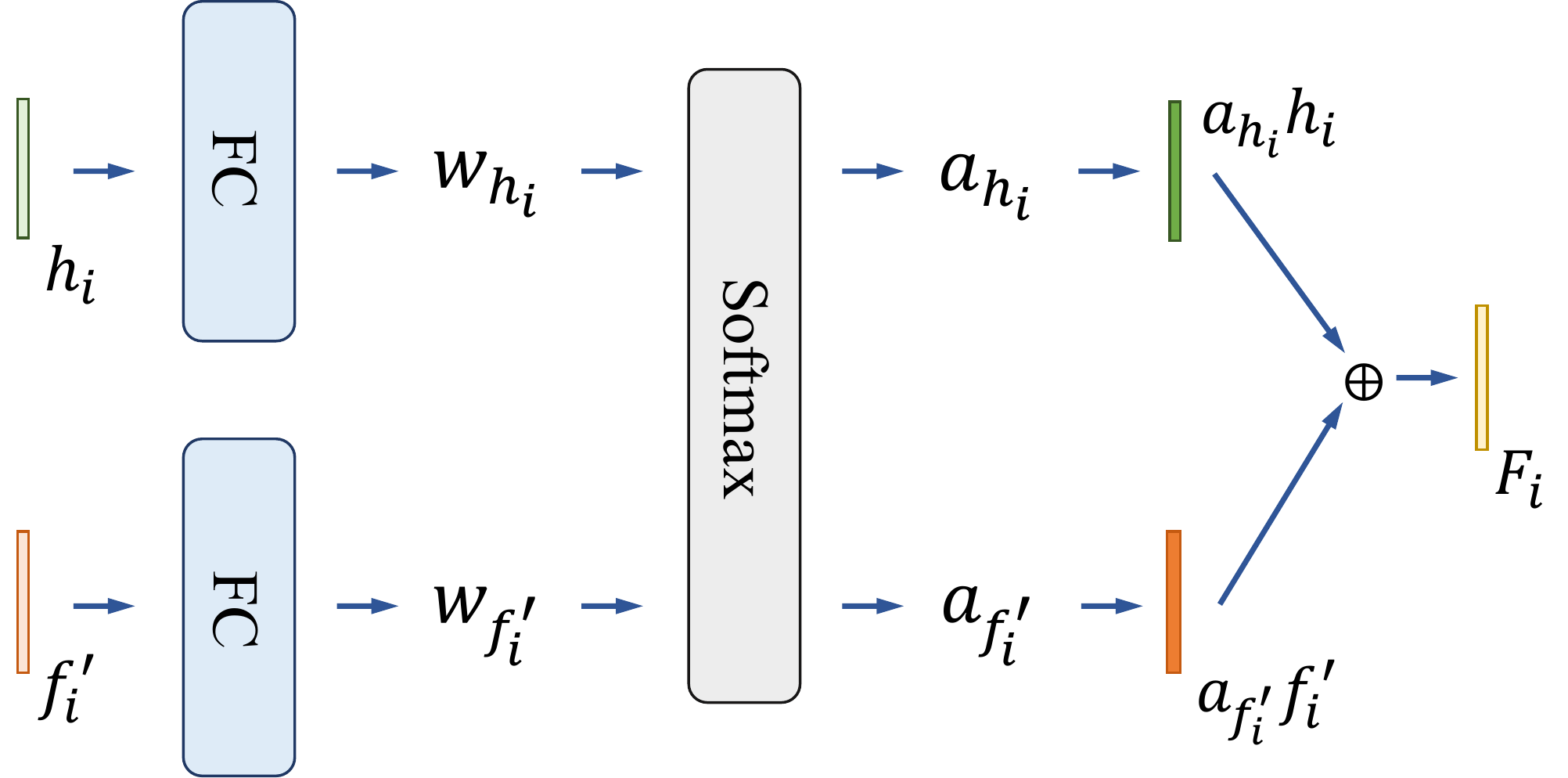}
  \end{center}
  \vspace{-5.0mm}
  \caption{
  \textbf{Overview of the feature fusion module.}
  Our feature fusion module consists of a shared fully connected  layer and a softmax layer.
  We first apply the FC layer to each input feature to predict a weight.
  We then apply a softmax layer to normalize the predicted weights.
  The input features are combined with the normalized weights to generate $F_i$.
  }
  \label{fig:fusion}
  \vspace{-5.0mm}  
\end{figure}

As shown in Figure~\ref{fig:fusion}, our feature fusion module is composed of a fully connected (FC) layer, followed by a softmax layer.
Given a latent representation $h_i$ and a predicted feature $f_i'$, we first apply the FC layer to each input feature to predict a weight.
The predicted weights are then normalized using a softmax layer.
The two input features are then fused by $F_i = a_{h_i} \cdot h_i + a_{f_i'} \cdot f_i' \in \mathbb{R}^d$.
We note that since $f_1'$ is not available, we define $F_1 = h_1$.

Next, we pass all the fused features $\{F_i\}_{i=1}^N$ to the shape $R_\mathrm{shape}$, pose $R_\mathrm{pose}$, and camera $R_\mathrm{camera}$ parameter regressors to predict the corresponding parameters, respectively.
Similar to one prior work~\citep{HMR}, we adopt an iterative error feedback scheme to regress the parameters.
To train the proposed SPS-Net, we impose a SMPL parameter regression loss $\mathcal{L}_\mathrm{SMPL}$ on the estimated pose $\{\hat{\theta}_i\}_{i=1}^N$ and shape $\hat{\beta}$ parameters, a 3D joint loss $\mathcal{L}_\mathrm{joint}^{3D}$ on the predicted 3D joints $\{\hat{X}_i\}_{i=1}^N$, and a 2D joint loss $\mathcal{L}_\mathrm{joint}^{2D}$ on the reprojected 2D joints $\{\hat{x}_i\}_{i=1}^N$~\citep{HMR,VIBE}.
Specifically, the SMPL parameter regression loss $\mathcal{L}_\mathrm{SMPL}$, the 3D joint loss $\mathcal{L}_\mathrm{joint}^{3D}$, and the 2D joint loss $\mathcal{L}_\mathrm{joint}^{2D}$ are defined as
\vspace{-3.5mm}
\begin{equation} \small
  \begin{split}
    & \mathcal{L}_\mathrm{SMPL} = \|\beta - \hat{\beta}\|_2 + \sum_{i=1}^{N} \|\theta_i - \hat{\theta}_i\|_2, \\
    &  \mathcal{L}_\mathrm{joint}^{3D} \mathrm{=} \sum_{i=1}^{N} \|X_i \mathrm{-} \hat{X}_i\|_2, 
    \quad 
    \mathcal{L}_\mathrm{joint}^{2D} \mathrm{=} \sum_{i=1}^{N} \|x_i \mathrm{-} \hat{x}_i\|_2.
  \end{split}
  \label{eq:vibe-loss}
  \vspace{-3.0mm}  
\end{equation}

\heading{Mask loss.} 
Since the ground-truth pose $\{\theta_i\}_{i=1}^N$, shape $\beta$, and 3D joint $\{X_i\}_{i=1}^N$ annotations are usually not available, using the 2D joint loss $\mathcal{L}_\mathrm{joint}^{2D}$ alone is insufficient to train the SPS-Net as there are numerous 3D meshes that can explain the same 2D projection.
To address this issue, we exploit the idea that the reprojection of the 3D mesh using the estimated camera parameters should be consistent with the segmentation mask obtained by directly segmenting the human from the input video frame.
We leverage an off-the-shelf instance segmentation model~\citep{YOLACT} to compile a pseudo ground-truth segmentation mask $m_i^\mathrm{pseudo}$ for each input video frame $I_i$.\footnote{We note that while other existing instance segmentation models can also be used for compiling segmentation masks, we leave the discussion of adopting different instance segmentation models as future work.}
Then, we use the pseudo ground-truth segmentation mask to supervise the reprojection of the 3D mesh with a mask loss:
\vspace{-1.5mm}  
\begin{equation}
  \mathcal{L}_\mathrm{mask} = - \sum_{i=1}^{N} m_i^\mathrm{pseudo} \log(m_i^\mathrm{proj}),
  \vspace{-1.5mm}  
  \label{eq:mask-loss}
\end{equation}
where $m_i^\mathrm{proj}$ denotes the reprojection of the 3D mesh using the estimated camera parameters.

\heading{Camera parameter consistency loss.}
Since there are no ground-truth camera parameter annotations for most datasets, existing methods~\citep{VIBE,HMR,SPIN} regularize the estimation of camera parameters via reprojecting the detected 3D keypoints onto 2D space and enforcing a 2D joint loss $\mathcal{L}_\mathrm{joint}^{2D}$ between the reprojected 2D joints and the corresponding ground-truth 2D joints.
This weaker form of supervision, however, is still under-constrained.
To address the absence of ground-truth camera parameter annotations, we exploit the idea that the overlapped video frames in different sequence segments from the same video should have the same camera parameter predictions.
Given two input sequence segments $S_1 = \{I_i^{S_1}\}_{i=n}^{k}$ and $S_2 = \{I_i^{S_2}\}_{i=n+1}^{k+1}$ from the same video $V$, the overlapped frames are $\{I_i\}_{i=n+1}^{k}$.
We enforce the camera parameter predictions of the overlapped frames $\{I_i\}_{i=n+1}^{k}$ to be the same in these two input sequence segments $S_1$ and $S_2$.
To achieve this, we propose a camera parameter consistency loss $\mathcal{L}_\mathrm{camera}$ which is defined as
\vspace{-2.0mm} 
\begin{equation}
  \mathcal{L}_\mathrm{camera} = \sum_{i=n+1}^{k}\|R_\mathrm{camera}(F_i^{S_1}) - R_\mathrm{camera}(F_i^{S_2})\|_2,
  \label{eq:camera-loss}
  \vspace{-1.5mm} 
\end{equation}
where $F_i^{S_1} \in \mathbb{R}^{d}$ and $F_i^{S_2} \in \mathbb{R}^{d}$ are the fused feature of frame $I_i^{S_1}$ and frame $I_i^{S_2}$, respectively.
Incorporating such consistency loss during training not only regularizes the prediction of camera parameters but also provides more supervisory signals to facilitate model training.

\heading{Adversarial loss.} 
In addition to the aforementioned loss functions, we also adopt an adversarial learning scheme that aims to encourage our method to recover a sequence of 3D meshes with realistic motions~\citep{VIBE}.
Similar to the VIBE~\citep{VIBE} method, we adopt the AMASS~\citep{AMASS} dataset and employ a discriminator $D$ that takes as input a sequence of pose parameters with the associated shape parameters $\hat{\Theta} = [\hat{\theta}_1, ..., \hat{\theta}_N, \hat{\beta}]$ estimated by the SPS-Net (treated as a fake example) and a sequence of those $\Theta = [\theta_1, ..., \theta_N, \beta]$ sampled from the AMASS~\citep{AMASS} dataset (treated as a real example), and aims to distinguish whether the input sequences are realistic or not.

\begin{figure}[t]
  \begin{center}
    \includegraphics[width=1.0\linewidth]{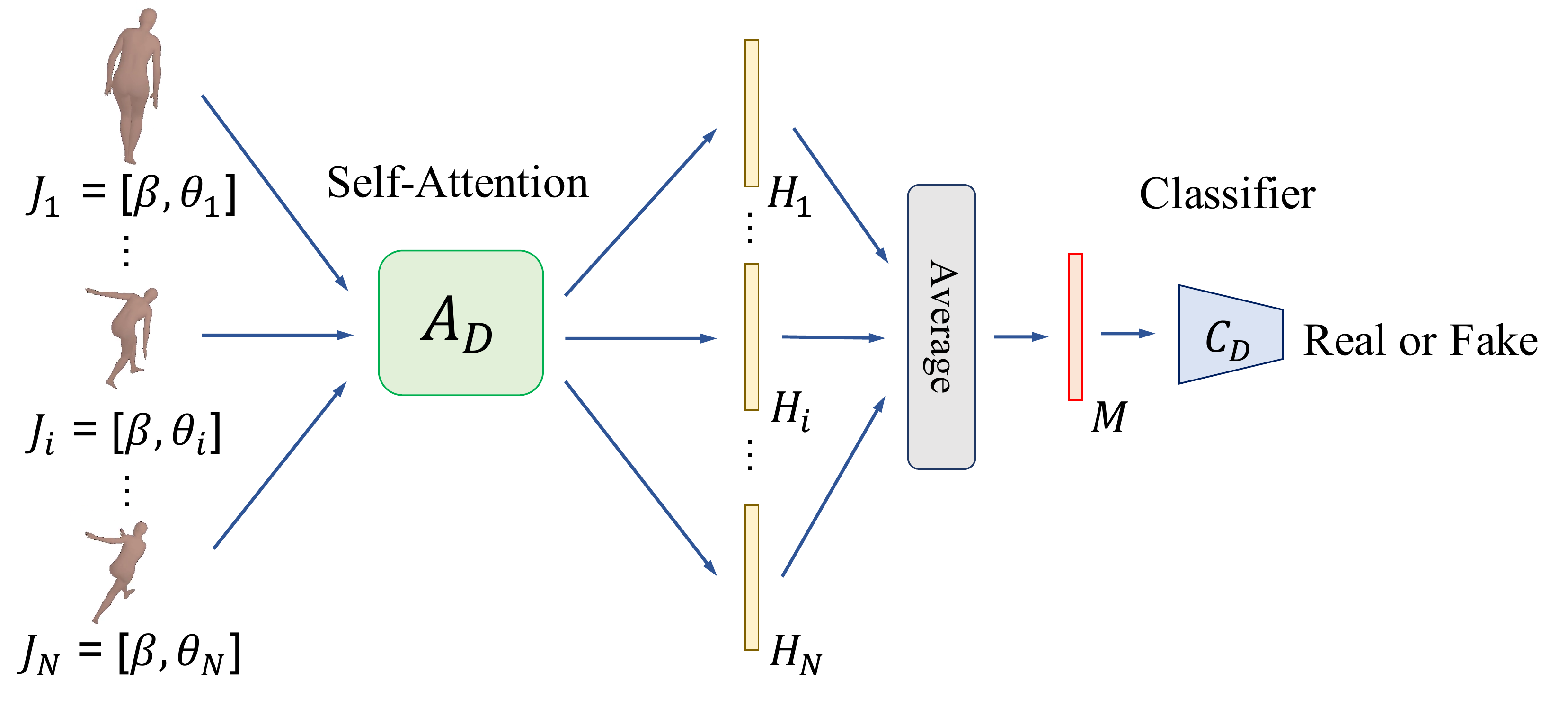}
  \end{center}
  \vspace{-5.0mm}
  \caption{
  \textbf{Overview of the discriminator $D$.}
  Our discriminator $D$ consists of a self-attention module $A_D$ and a classifier $C_D$.
  Given a sequence of pose parameters $\{\theta_i\}_{i=1}^{N}$ and the associated shape parameters $\beta$, we first derive the latent representations $\{H_i\}_{i=1}^{N}$ using the self-attention module $A_D$.
  We then average all the latent representations $\{H_i\}_{i=1}^{N}$ to derive the motion representation $M$.
  The classifier $C_D$ takes the motion representation $M$ as input and distinguishes whether the input motion representation $M$ is realistic or not.
  }
  \label{fig:discriminator}
  \vspace{-5.0mm} 
\end{figure}

As shown in Figure~\ref{fig:discriminator}, our discriminator $D$ is composed of a self-attention module $A_D$ and a classifier $C_D$.
We first concatenate the estimated shape parameters $\hat{\beta}$ with each of the estimated pose parameters $\{\hat{\theta}_i\}_{i=1}^N$ to form the joint representations $\{\hat{J}_i\}_{i=1}^N$, where $\hat{J}_i = [\hat{\beta}, \hat{\theta}_i] \in \mathbb{R}^{82}$.
We then pass all joint representations $\{\hat{J}_i\}_{i=1}^N$ to the self-attention module $A_D$ to derive the latent representations $\{\hat{H}_i\}_{i=1}^N$, where $\hat{H}_i \in \mathbb{R}^{82}$ is the latent representation of $\hat{J}_i$.
To derive the motion representation $\hat{M}$ of $\hat{\Theta}$, we average all the latent representations $\{\hat{H}_i\}_{i=1}^N$, i.e., $\hat{M} = \frac{1}{N}\sum_{i=1}^{N}\hat{H}_i \in \mathbb{R}^{82}$.
The motion representation $M \in \mathbb{R}^{82}$ of $\Theta$ can be derived similarly.
The classifier $C_D$ takes the motion representations $\hat{M}$ and $M$ as input and distinguishes whether the input motion representations are realistic or not.
Specifically, we have an adversarial loss $\mathcal{L}_\mathrm{adv}$ which is defined as
\vspace{-1.5mm}  
\begin{equation}
  \mathcal{L}_\mathrm{adv} = \mathbb{E}_{\Theta \sim p_\Theta}[\|D(\Theta) - 1\|_2] + \mathbb{E}_{\hat{\Theta} \sim p_{\hat{\Theta}}}[\|D(\hat{\Theta})\|_2].
  \vspace{-1.0mm}  
  \label{eq:adv-loss}
\end{equation}
Leveraging the unpaired data from the AMASS~\citep{AMASS} dataset serves as a weak supervision to encourage the SPS-Net to recover a sequence of 3D meshes with realistic motions.

We note that our discriminator $D$ is different from that of the VIBE~\citep{VIBE} method in two aspects.
First, our discriminator has a self-attention module, while the discriminator of the VIBE~\citep{VIBE} method has two GRU layers.
Second, we use self-attention to derive a representation for each frame that contains temporal information by jointly considering short-range and long-range dependencies across video frames, whereas the VIBE~\citep{VIBE} method leverages self-attention to derive a single representation for the entire pose sequence.

\heading{Self-supervised occlusion handling.}
While the aforementioned loss functions regularize the learning of the SPS-Net, the 2D and 3D joint losses and the mask loss are only enforced on the visible keypoints and regions of the human body.
That is, there is no explicit constraint imposed on the invisible keypoints and regions.
We develop a self-supervised learning scheme to allow our model to produce plausible predictions in order to account for the occlusion and out-of-view scenarios. 
For each input frame $I_i$, we first synthesize the occluded version $I_i'$ by randomly masking out some regions.
We then leverage the predictions of the original frames to \emph{supervise} those of the synthesized occluded or partially visible frames and develop a self-supervised parameter regression loss $\mathcal{L}_\mathrm{param}$ to exploit this property with 
\vspace{-4.0mm} 
\begin{equation} \small
  \begin{split}
    \mathcal{L}_\mathrm{param} & = \|\hat{\beta} - \hat{\beta}'\|_2 + \sum_{i=1}^{N} \|\hat{\theta}_i - \hat{\theta}_i'\|_2 \\
    & + \sum_{i=1}^{N} \|R_\mathrm{camera}(F_i) - R_\mathrm{camera}(F_i')\|_2.
  \end{split}
  \label{eq:param-loss}
  \vspace{-4.0mm} 
\end{equation}
By simulating the occlusion and out-of-view scenes, our model is able to predict plausible shape, pose, and camera parameters from the occluded or partially visible frames.

\vspace{-3.0mm}  
\section{Experimental Results}

In this section, we first describe the implementation details.
Next, we describe the datasets for model training and testing, followed by the evaluation metrics.
We then present the quantitative and visual comparisons to existing methods as well as the ablation study.

\vspace{-2.0mm} 
\subsection{Implementation details} 

We implement our model using PyTorch~\citep{paszke2019pytorch}.
Same as prior work~\citep{HMR,VIBE}, we adopt the ResNet-$50$~\citep{ResNet} pre-trained on single-image pose and shape estimation task~\citep{HMR,SPIN} to serve as our encoder $E$.
Our encoder $E$ is fixed and outputs a $2,048$-dimensional feature for each frame, i.e., $f_i \in \mathbb{R}^{2048}$.
We set the length of the input sequence to $32$ with a batch size of $16$.
Both the attention network $Q$ and the attention network $K$ in the self-attention module $A$ consist of $2$ fully connected layers, each of which has a hidden size of $2,048$, followed by a LeakyReLU layer.
As for the forecasting module $F$, unlike prior methods~\citep{PHD,TemporalHMR} that use 1D convolution layers, our forecasting module $F$ is composed of $2$ fully connected layers, each of which has a hidden size of $2,048$, followed by a LeakyReLU layer.
Both the attention network $Q$ and the attention network $K$ in the self-attention module $A_D$ also consist of $2$ fully connected layers, each of which has a hidden size of $82$, followed by a LeakyReLU layer.
The classifier $C_D$ in the discriminator $D$ is composed of a fully connected layer, followed by a sigmoid function.
The input and output dimensions of the classifier $C_D$ are $82$ and $1$, respectively.
Similar to the HMR~\citep{HMR}, the SMPL~\citep{SMPL} parameter regressor $\{R_\mathrm{pose}, R_\mathrm{shape}\}$ is composed of $2$ fully connected layers with a hidden size of $1,024$.
The shape $R_\mathrm{shape}$, pose $R_\mathrm{pose}$, and camera $R_\mathrm{camera}$ parameter regressors are initialized from the pre-trained weights of the HMR~\citep{HMR} approach.
The weights of the self-attention module $A$, the forecasting module $F$, the feature fusion module, and the discriminator $D$ are randomly initialized.
We use the ADAM~\citep{kingma2014adam} optimizer for training.
The learning rates for the SPS-Net and the discriminator $D$ are set to $5 \times 10^{-5}$ and $1 \times 10^{-4}$, respectively.
Following the VIBE~\citep{VIBE} method, we set the hyperparameters for the loss functions as follows: $\lambda_\beta \mathrm{=} 0.06$, $\lambda_\theta \mathrm{=} 60$, $\lambda_\mathrm{joint}^{3D} \mathrm{=} 300$, $\lambda_\mathrm{joint}^{2D} \mathrm{=} 300$, and $\lambda_\mathrm{adv} \mathrm{=} 2$.
For the other hyperparameters, we set $\lambda_\mathrm{feature} \mathrm{=} 1$, $\lambda_\mathrm{mask} \mathrm{=} 300$, $\lambda_\mathrm{camera} \mathrm{=} 0.1$, $\lambda_\mathrm{param}^\beta \mathrm{=} 0.06$, $\lambda_\mathrm{param}^\theta \mathrm{=} 60$, and $\lambda_\mathrm{param}^\mathrm{camera} \mathrm{=} 0.1$.
We train our model on a single NVIDIA V$100$ GPU with $32$GB memory for $120$ epochs.
For each epoch, there are $500$ iterations.

\heading{Camera parameter consistency loss $\mathcal{L}_\mathrm{camera}$.} 
To compute the camera parameter consistency loss $\mathcal{L}_\mathrm{camera}$, in each iteration we sample two consecutive sequence segments by shifting the starting index for data sampling by $1$.
Assuming that the starting index for data sampling is $n$, we first sample a sequence segment $S_1 = \{I_i\}_{i=n}^{n+31}$.
We then shift the starting index for data sampling by $1$ and sample another sequence segment $S_2 = \{I_i\}_{i=n+1}^{n+32}$.
Given these two sequence segments $S_1$ and $S_2$, the overlapped video frames are $\{I_i\}_{i=n+1}^{n+31}$.
We enforce the camera parameter predictions of the overlapped video frames $\{I_i\}_{i=n+1}^{n+31}$ to be the same in these two sequence segments with a camera parameter consistency loss.

\heading{Self-supervised occlusion handling.} 
Since the ground-truth 2D joint annotations are available, for each training image, we randomly sample $3$ to $5$ keypoints.
For each keypoint, we randomly sample a width offset between $25$ and $50$ pixels and a height offset between $25$ and $50$ pixels to determine the region to be masked out for synthesizing the occluded training data.
The shape, pose, and camera parameter predictions of the occluded training data are supervised by those of the original training data.
We note that for frames with ground-truth pose parameter annotations, the self-supervised parameter regression loss $\mathcal{L}_\mathrm{param}$ can be computed against the ground truth.
However, in our training set, only the MPI-INF-3DHP~\citep{MPII} and Human3.6M~\citep{human36m} datasets contain ground-truth pose parameter annotations.
For ease of implementation, we choose to compute the loss against the predictions of the original frames.
The formulation of the self-supervised parameter regression loss $\mathcal{L}_\mathrm{param}$ is applicable to all training data, with or without ground truth.

\heading{Multi-person tracking.} 
To recover human body from videos that contain multiple person instances, we first leverage a multi-person tracker to detect and track each person instance.
We then apply our SPS-Net to each person tracking result to estimate the 3D human pose and shape.
The multi-person tracker is composed of an object detector and an object tracker.
We adopt the YOLOv4~\citep{bochkovskiy2020yolov4} as the object detector and the SORT~\citep{bewley2016simple} as the object tracker.
The multi-person tracker first applies the YOLOv4~\citep{bochkovskiy2020yolov4} detector to each video frame to detect each person instance.
Then the person detection results are passed to the SORT~\citep{bewley2016simple} method to associate the detected person instances in the current frame to the existing ones.
Specifically, the SORT~\citep{bewley2016simple} first predicts the bounding box in the current frame for each existing person.
Then, we compute the intersection over union (IoU) between the detected bounding boxes and the predicted bounding boxes.
By using the Hungarian algorithm with a minimum IoU threshold, we can assign each detected person instance to an existing one or consider the detected person instance a new one.

\begin{table*}[t]
  \begin{center}
  \scriptsize
  \caption{
  \textbf{Experimental results of 3D human pose and shape estimation.}
  We present the experimental results with comparisons to existing methods.
  (\emph{Left}) Results on the 3DPW~\citep{3DPW} dataset.
  (\emph{Middle}) Results on the MPI-INF-3DHP~\citep{MPII} dataset.
  (\emph{Right}) Results on the Human3.6M~\citep{human36m} dataset.
  The \first{bold} and \second{underlined} numbers indicate the top two results, respectively.
  The ``-'' indicates the result is not available.
  }
  \vspace{-3.0mm}
  \label{exp:comparison}
  \resizebox{\linewidth}{!} 
  {
  \begin{tabular}{cl|c|cccc|ccc|cc}
    \toprule
    & \multirow{2}{*}{Method} & \revised{Number of} & \multicolumn{4}{c|}{3DPW~\citep{3DPW}} & \multicolumn{3}{c|}{MPI-INF-3DHP~\citep{MPII}} & \multicolumn{2}{c}{Human3.6M~\citep{human36m}} \\
    &  & \revised{parameters} & PA-MPJPE $\downarrow$ & MPJPE $\downarrow$ & PVE $\downarrow$ & Acceleration Error $\downarrow$ & PA-MPJPE $\downarrow$ & MPJPE $\downarrow$ & PCK $\uparrow$ & PA-MPJPE $\downarrow$ & MPJPE $\downarrow$ \\
    \midrule
    \multirow{13}{*}{\rotatebox{90}{Frame based}} & Yang~\etal~\citep{yang20183d} & - & - & - & - & - & - & - & 69.0 & - & - \\
    & Chen~\etal~\citep{chen2019unsupervised} & - & - & - & - & - & - & - & 71.1 & - & - \\
    & Mehta~\etal~\citep{mehta2017vnect} & 9.81M & - & - & - & - & - & - & 72.5 & - & - \\
    & EpipolarPose~\citep{kocabas2019self} & 34.28M & - & - & - & - & - & - & 77.5 & - & - \\
    & TCN~\citep{TCN} & - & - & - & - & - & - & - & 84.1 & - & - \\
    & RepNet~\citep{wandt2019repnet} & 10.03M & - & - & - & - & - & 97.8 & 82.5 & - & - \\
    & CMR~\citep{CMR} & 46.31M & 70.2 & - & - & - & - & - & - & 50.1 & - \\
    & STRAPS~\citep{sengupta2020synthetic} & 12.48M & 66.8 & - & - & - & - & - & - & 55.4 & - \\
    & NBF~\citep{omran2018neural} & 68.11M & 90.7 & - & - & - & - & - & - & 59.9 & - \\
    & ExPose~\citep{ExPose} & 47.22M & 60.7 & 93.4 & - & - & - & - & - & - & - \\
    & HUND~\citep{zanfir2020neural} & - & \second{56.5} & \second{87.7} & - & - & - & - & - & 53.0 & 72.0 \\
    & HMR~\citep{HMR} & 26.98M & 76.7 & 130.0 & - & 37.4 & 89.8 & 124.2 & 72.9 & 56.8 & 88.0 \\
    & SPIN~\citep{SPIN} & 26.98M & 59.2 & 96.9 & 116.4 & 29.8 & 67.5 & 105.2 & 76.4 & \second{41.1} & - \\
    \midrule
    \multirow{6}{*}{\rotatebox{90}{Video based}}  & Temporal 3D Kinetics~\citep{Temporal3DKinetics} & - & 72.2 & - & - & - & - & - & - & - & - \\
    & Motion to the Rescue~\citep{doersch2019sim2real} & - & 74.7 & - & - & - & - & - & - & - & - \\
    & DSD-SATN~\citep{sun2019human} & - & 69.5 & - & - & - & - & - & - & 42.4 & \second{59.1} \\
    & \revised{HMMR}~\citep{TemporalHMR} & 29.76M & 72.6 & 116.5 & 139.3 & \first{15.2} & - & - & - & 56.9 & - \\
    & VIBE~\citep{VIBE} & 48.30M & \second{56.5} & 93.5 & \second{113.4} & 27.1 & \second{63.4} & \second{97.7} & \second{89.0} & 41.5 & 65.9 \\
    & Ours & 51.43M & \first{50.4} & \first{85.8} & \first{100.6} & \second{22.1} & \first{60.7} & \first{94.3} & \first{90.1} & \first{38.7} & \first{58.9} \\
    \bottomrule
  \end{tabular}
  }
  \end{center}
  \vspace{-4.0mm}
\end{table*}

\begin{figure*}[t]
  \begin{center}
  \mpage{0.01}{\rotatebox{90}{SPIN}}
  \mpage{0.1485}{\includegraphics[width=\linewidth]{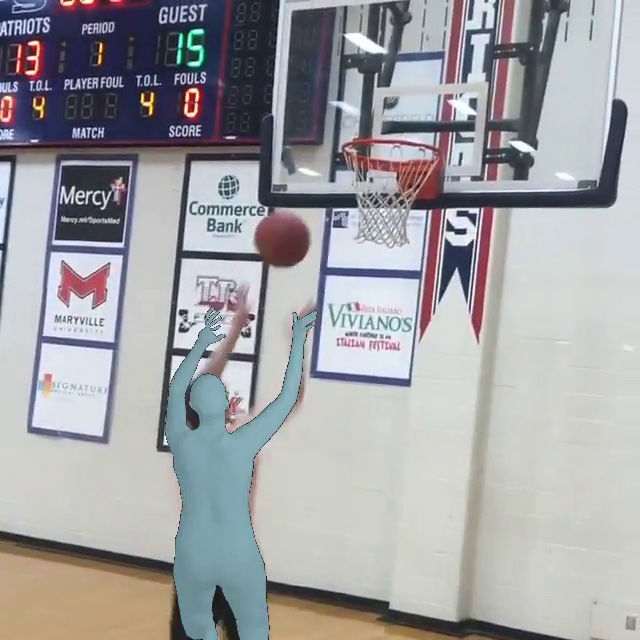}} \hfill
  \mpage{0.1485}{\includegraphics[width=\linewidth]{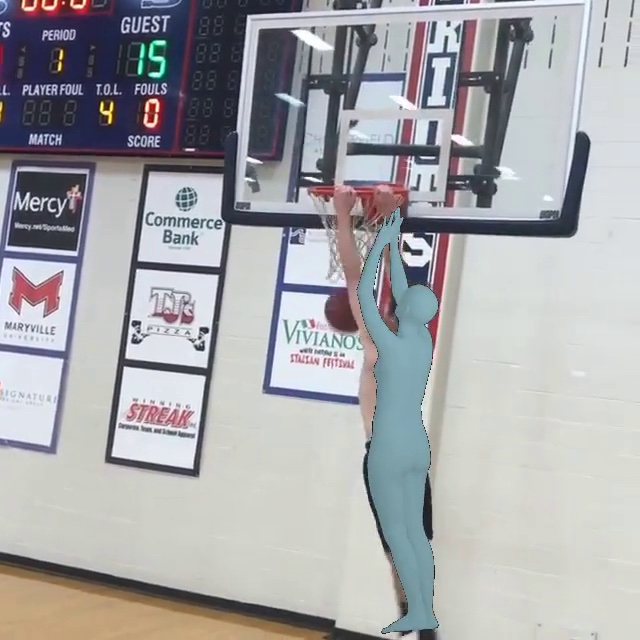}} \hfill
  \mpage{0.1485}{\includegraphics[width=\linewidth]{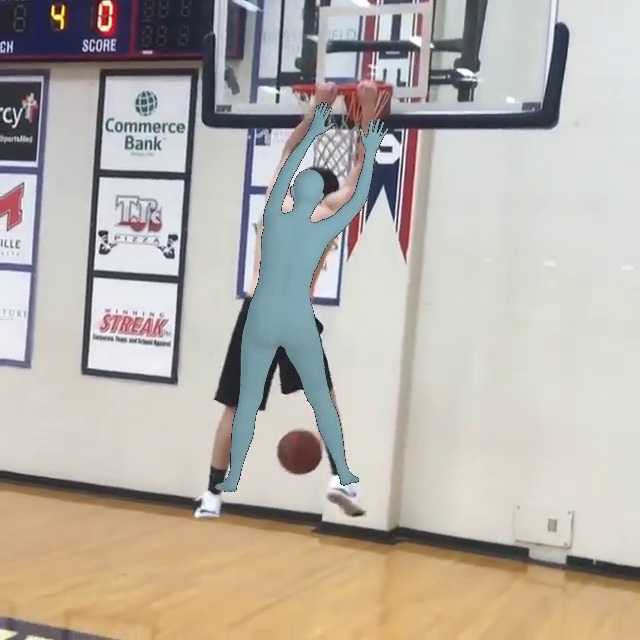}} \hfill
  \mpage{0.01}{} \hfill
  \mpage{0.1485}{\includegraphics[width=\linewidth]{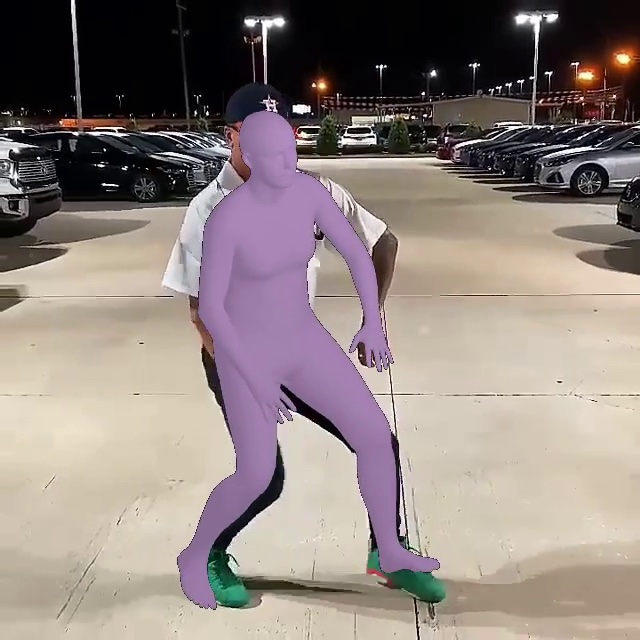}} \hfill
  \mpage{0.1485}{\includegraphics[width=\linewidth]{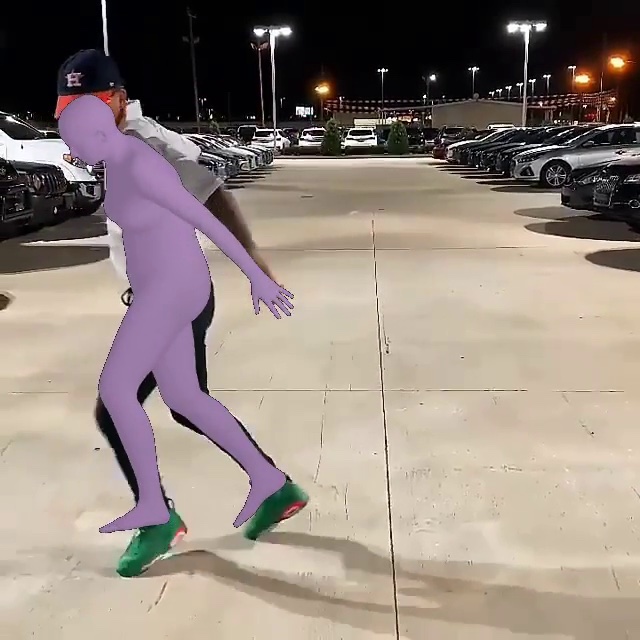}} \hfill
  \mpage{0.1485}{\includegraphics[width=\linewidth]{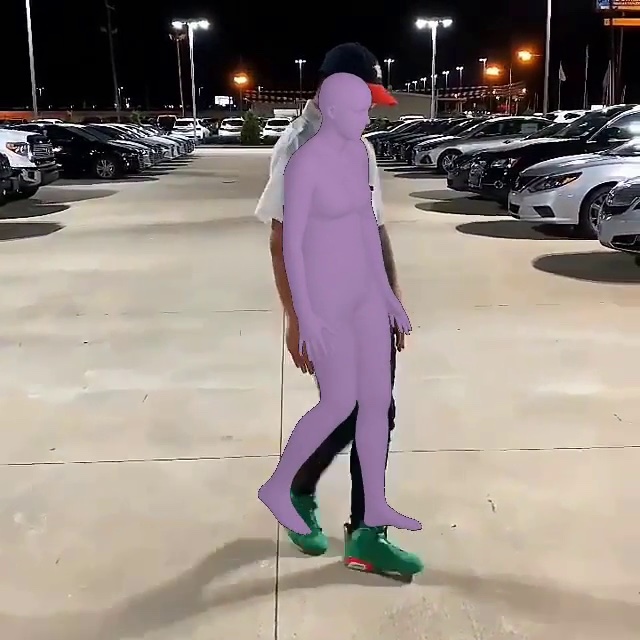}} \\
  \vspace{2.0mm}
  \mpage{0.01}{\rotatebox{90}{VIBE}}
  \mpage{0.1485}{\includegraphics[width=\linewidth]{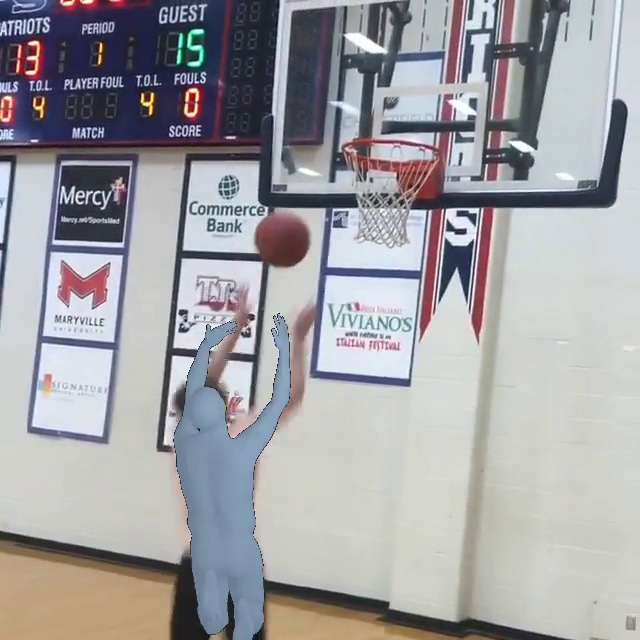}} \hfill
  \mpage{0.1485}{\includegraphics[width=\linewidth]{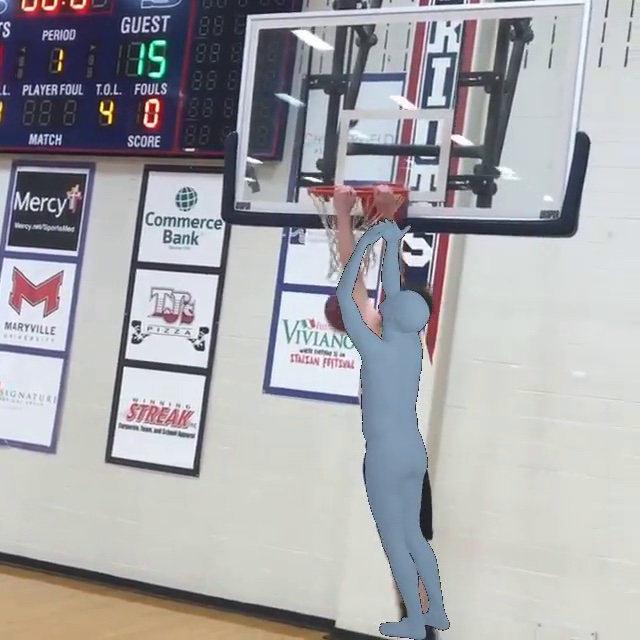}} \hfill
  \mpage{0.1485}{\includegraphics[width=\linewidth]{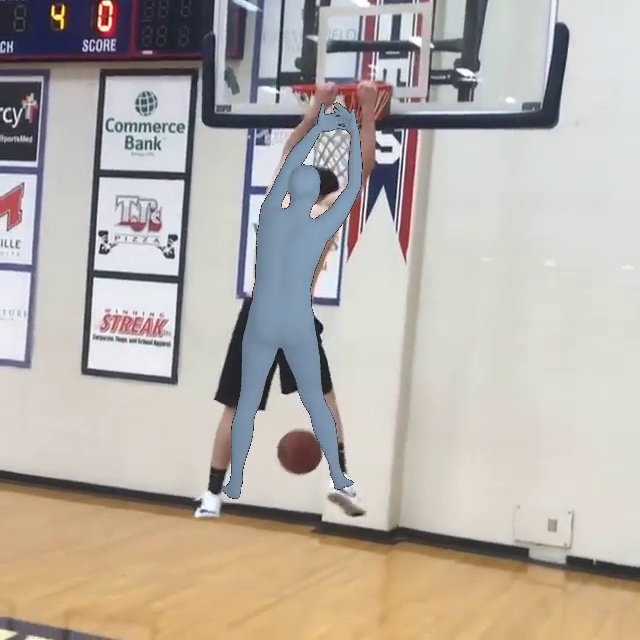}} \hfill
  \mpage{0.01}{} \hfill
  \mpage{0.1485}{\includegraphics[width=\linewidth]{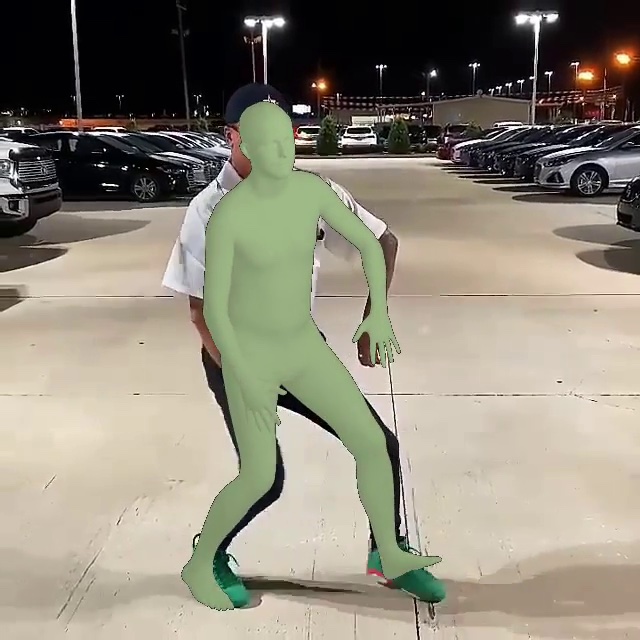}} \hfill
  \mpage{0.1485}{\includegraphics[width=\linewidth]{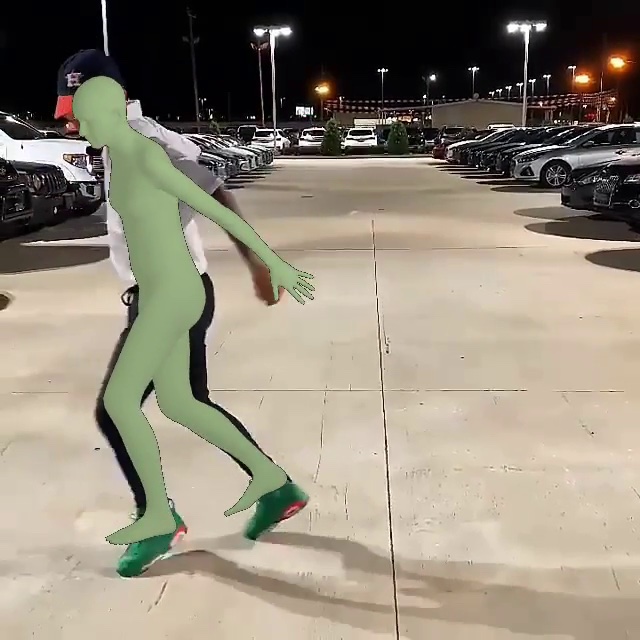}} \hfill
  \mpage{0.1485}{\includegraphics[width=\linewidth]{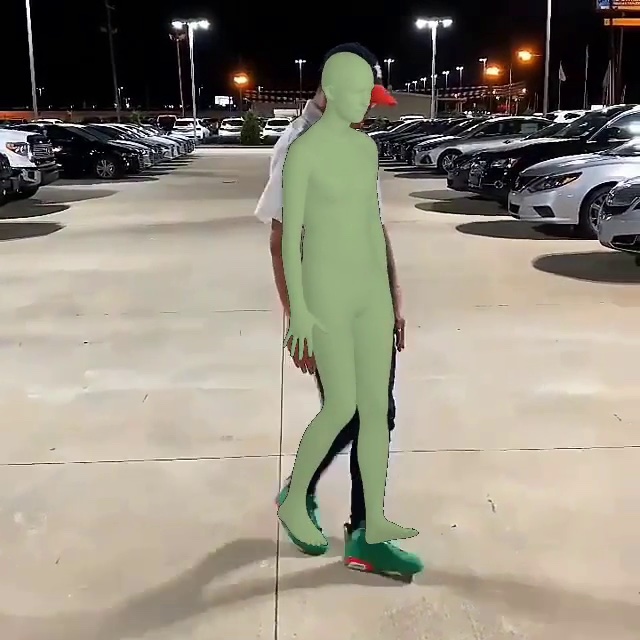}} \\
  \vspace{2.0mm}
  \mpage{0.01}{\rotatebox{90}{Ours}}
  \mpage{0.1485}{\includegraphics[width=\linewidth]{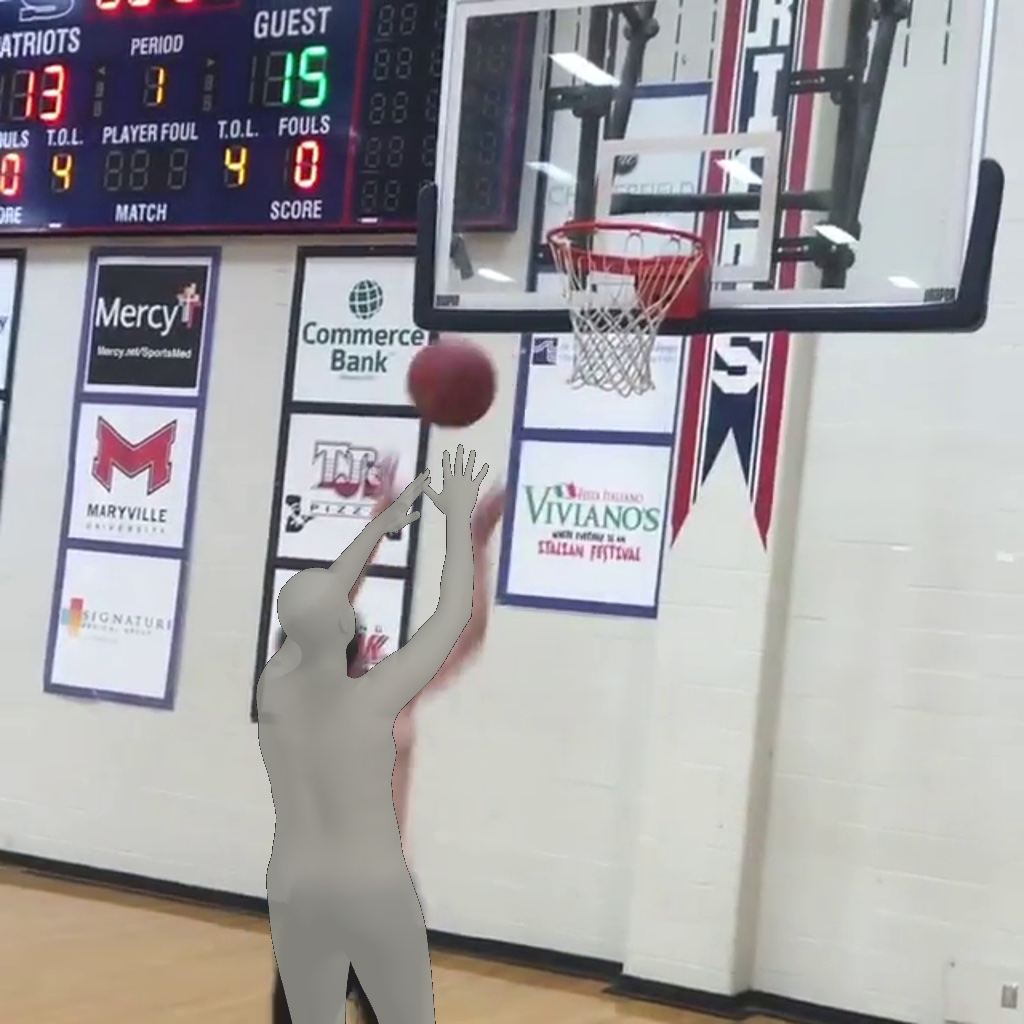}} \hfill
  \mpage{0.1485}{\includegraphics[width=\linewidth]{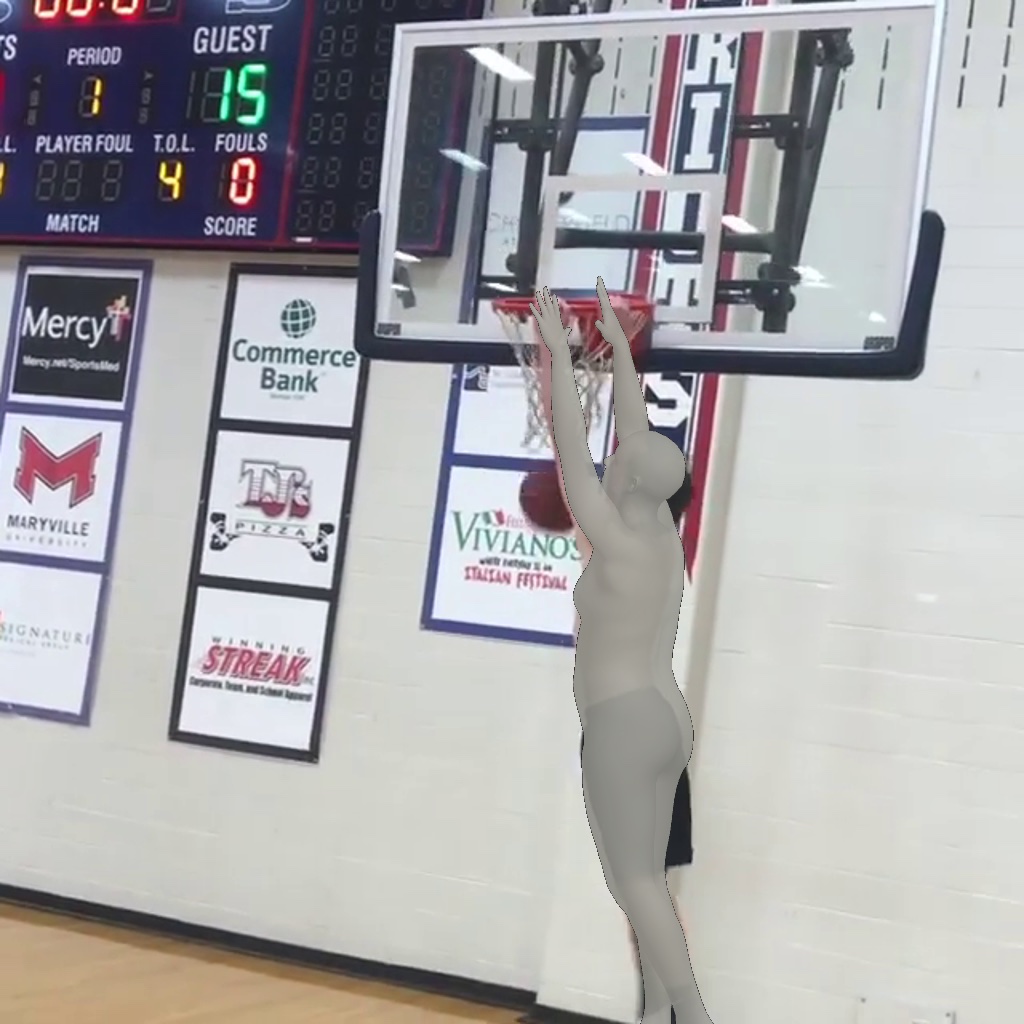}} \hfill
  \mpage{0.1485}{\includegraphics[width=\linewidth]{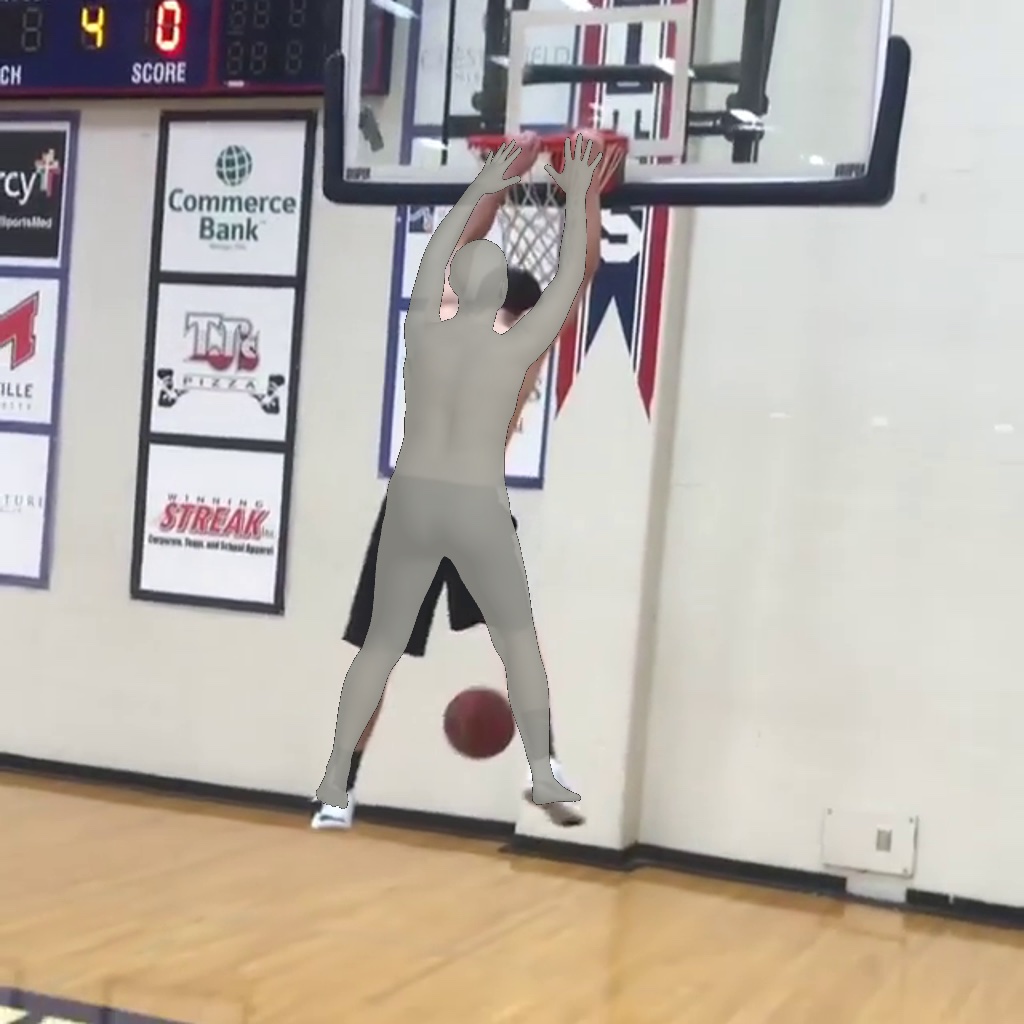}} \hfill
  \mpage{0.01}{} \hfill
  \mpage{0.1485}{\includegraphics[width=\linewidth]{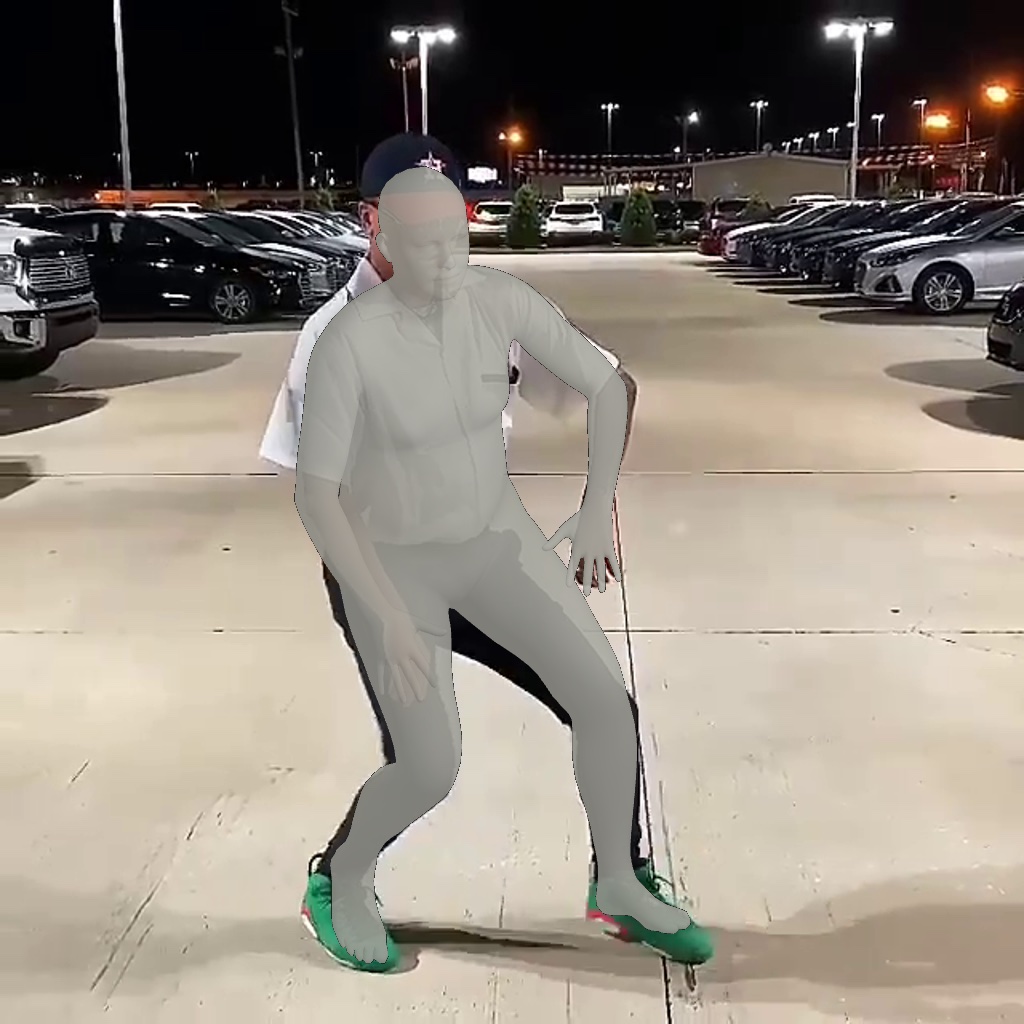}} \hfill
  \mpage{0.1485}{\includegraphics[width=\linewidth]{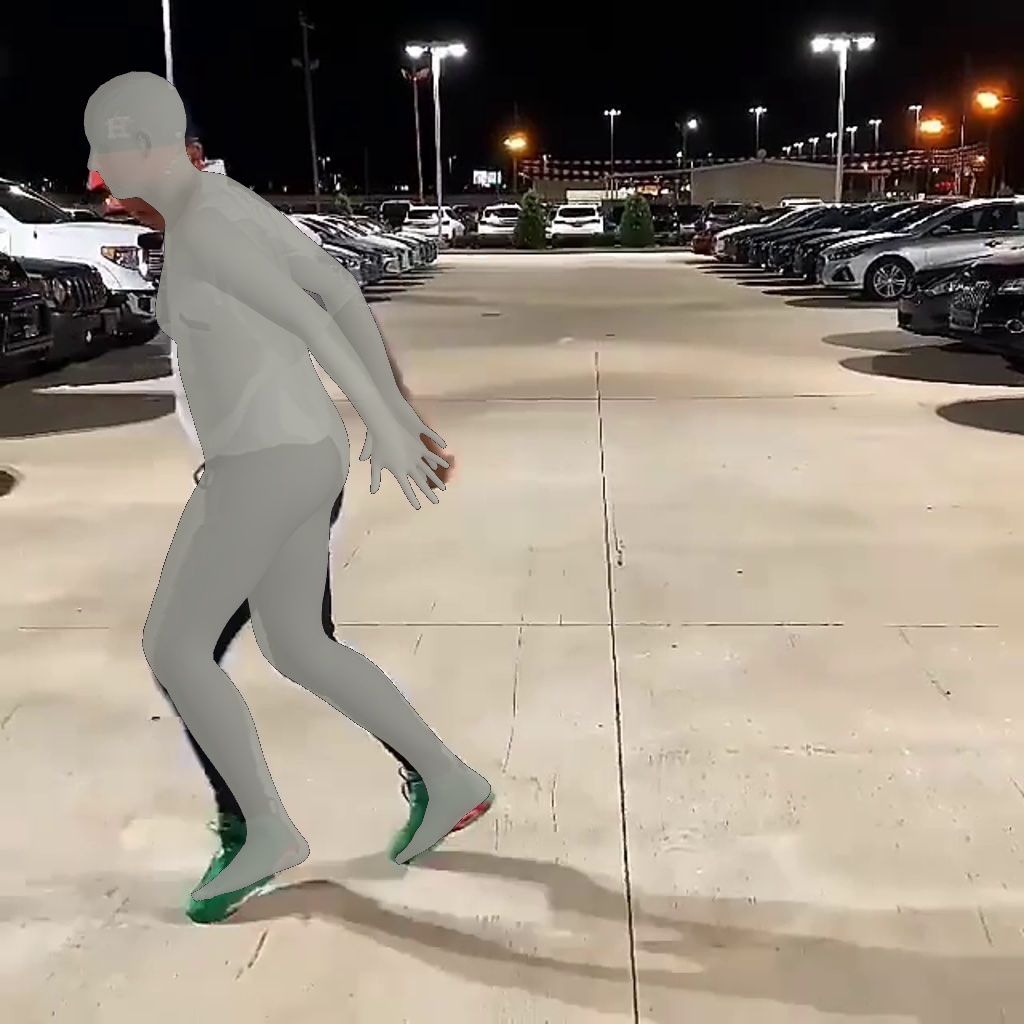}} \hfill
  \mpage{0.1485}{\includegraphics[width=\linewidth]{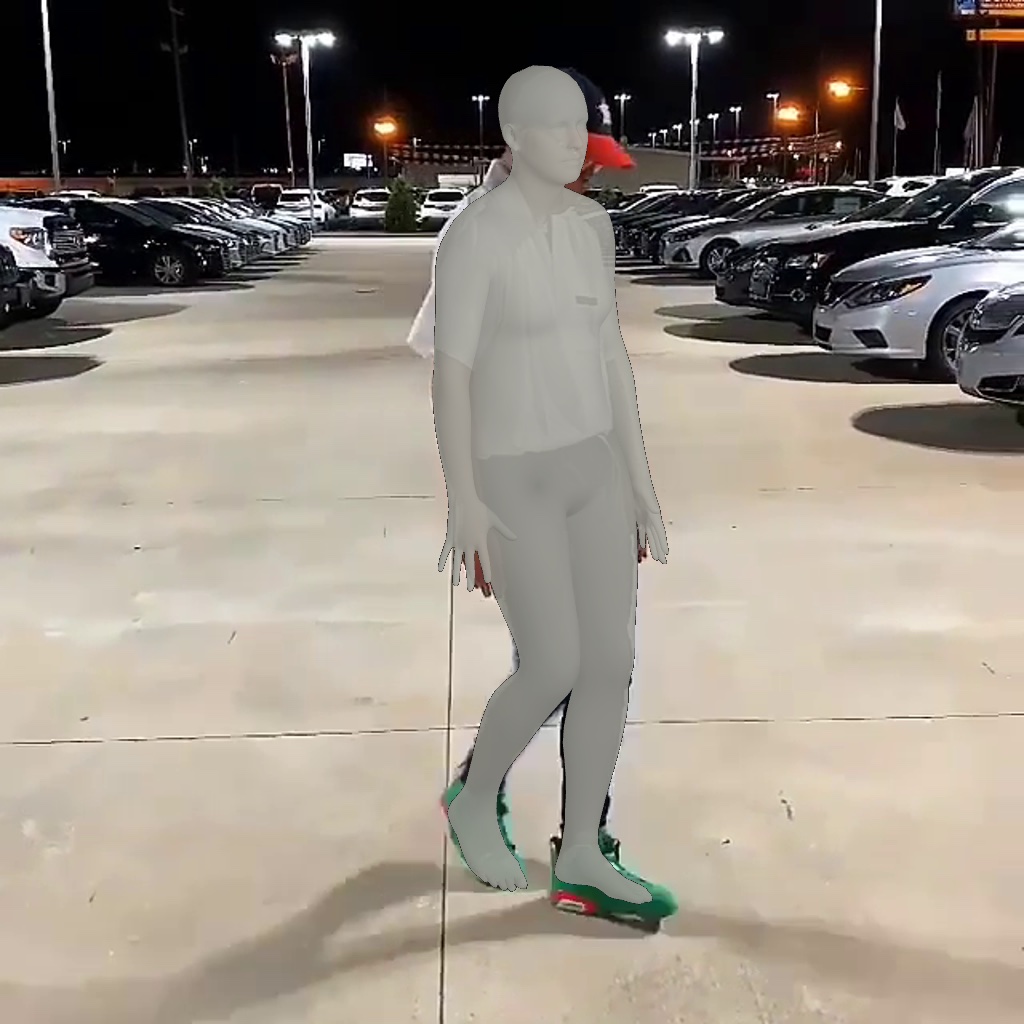}} \\
  \vspace{-1.5mm}
  \caption{
  \textbf{Visual comparisons.} 
  We present two visual comparisons with the SPIN~\citep{SPIN} and VIBE~\citep{VIBE} methods.
  Our method is capable of estimating shapes that cover human bodies well and predicting more accurate poses for limbs in particular.
  }
  \label{fig:comparison}
  \end{center}
  \vspace{-8.0mm}
\end{figure*}

\vspace{-3.0mm} 
\subsection{Experimental settings}

We describe the datasets and the evaluation metrics below.

\vspace{-3.0mm}  
\subsubsection{Datasets}

Similar to the state-of-the-art human pose and shape estimation methods~\citep{HMR,TemporalHMR,SPIN,VIBE}, we adopt a number of datasets that contain either 2D or 3D ground-truth annotations for training.
Specifically, we use the PennAction~\citep{PennAction}, InstaVariety~\citep{TemporalHMR}, PoseTrack~\citep{PoseTrack}, MPI-INF-3DHP~\citep{MPII}, and Human3.6M~\citep{human36m} datasets for training.
Same as the VIBE~\citep{VIBE} method, we use the Kinetics-$400$~\citep{kay2017kinetics} dataset to complement the missing parts of the InstaVariety~\citep{TemporalHMR} dataset.
We evaluate our method on the 3DPW~\citep{3DPW}, MPI-INF-3DHP~\citep{MPII}, and Human3.6M~\citep{human36m} datasets.
The details of each dataset are described below.

\heading{3DPW~\citep{3DPW}.}
The 3DPW dataset is an in-the-wild 3D dataset, containing $60$ videos of several in-the-wild and indoor activities. 
The training, validation, and test sets are composed of $24$, $12$, and $24$ video sequences, respectively.
We evaluate our method on the 3DPW test set.

\heading{MPI-INF-3DHP~\citep{MPII}.} 
The MPI-INF-3DHP dataset consists of multi-view videos captured in indoor environments.
The training set contains $8$ subjects, each of which has $16$ videos.
Following existing approaches~\citep{SPIN,VIBE}, we use the training set for model training and evaluate our SPS-Net on the test set.

\heading{Human3.6M~\citep{human36m}.}
The Human3.6M dataset is composed of $15$ sequences of several people performing different actions.
This dataset is collected in an indoor and controlled environment.
The training set contains $1.5$ million images, each of which has 3D ground-truth annotations.
Same as the VIBE~\citep{VIBE} method, we train our model on $5$ subjects (i.e., S$1$, S$5$, S$6$, S$7$, and S$8$) and evaluate our method on the remaining $2$ subjects (i.e., S$9$ and S$11$).

\heading{PennAction~\citep{PennAction}.} 
The PennAction dataset is composed of $2,326$ videos of $15$ actions.
Each video is annotated with 2D keypoints.
We use this dataset for training.

\heading{InstaVariety~\citep{TemporalHMR}.}
The InstaVariety dataset is composed of videos of $24$-hour long collected from Instagram.
Each video is annotated with 2D joints obtained by using the OpenPose~\citep{OpenPose} and Detect and Track~\citep{girdhar2018detect} methods.
We adopt this dataset for training.

\heading{PoseTrack~\citep{PoseTrack}.} 
The PoseTrack dataset consists of $1,337$ videos.
The training set is composed of $792$ videos.
The validation set contains $170$ videos.
The test set comprises $375$ videos.
Each video is annotated with $15$ keypoints.
We use the training set for model training.

\begin{figure*}[t]
  \begin{center}
  \mpage{0.23}{\includegraphics[width=\linewidth]{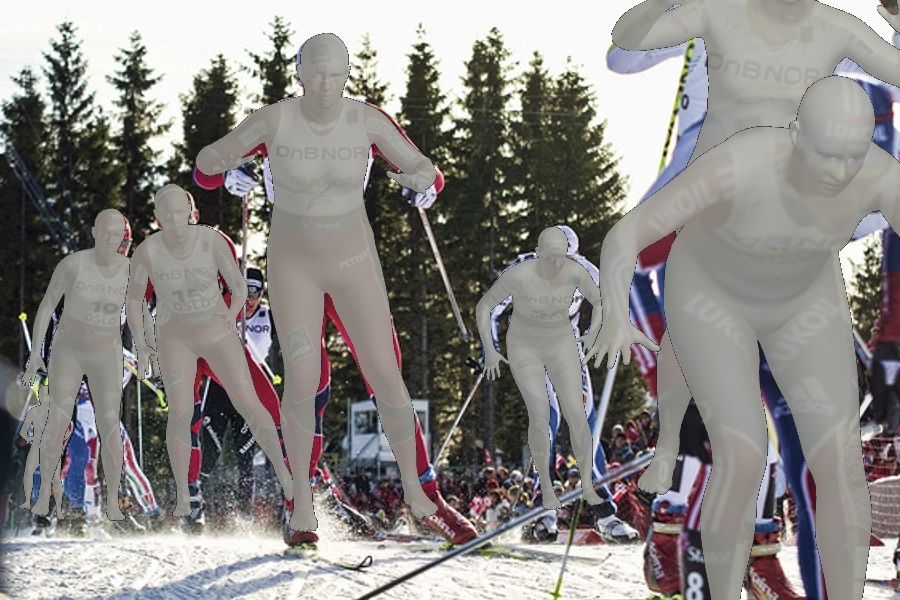}} \hfill
  \mpage{0.23}{\includegraphics[width=\linewidth]{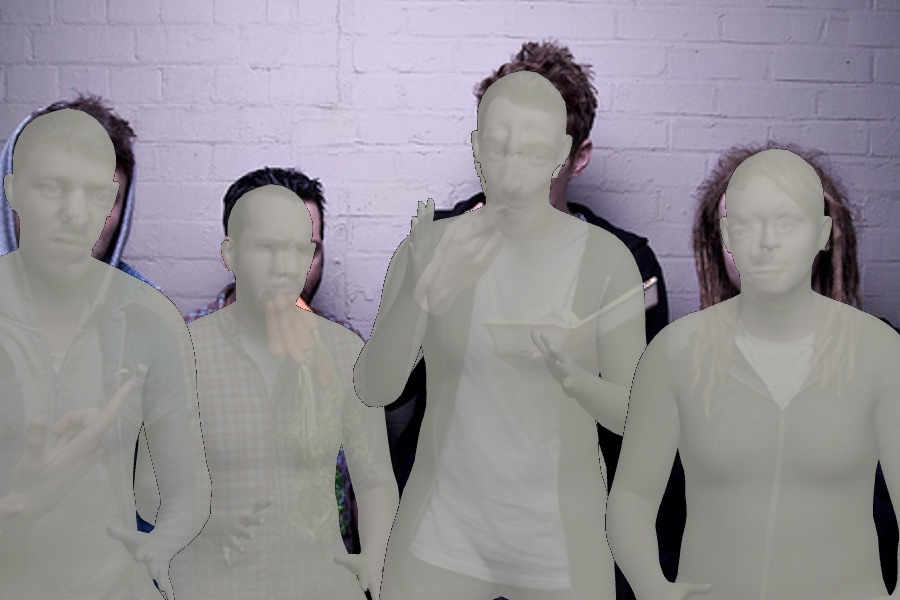}} \hfill
  \mpage{0.23}{\includegraphics[width=\linewidth]{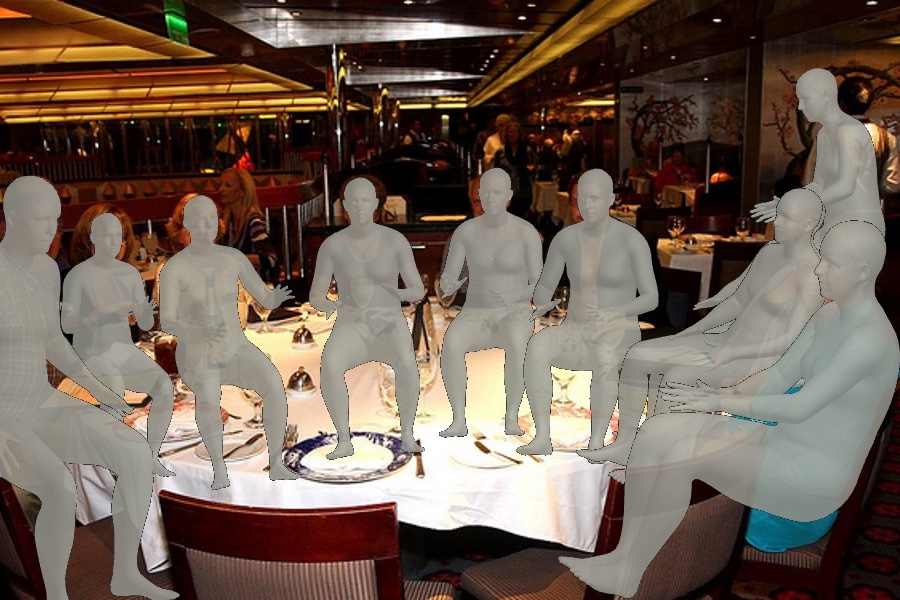}} \hfill
  \mpage{0.23}{\includegraphics[width=\linewidth]{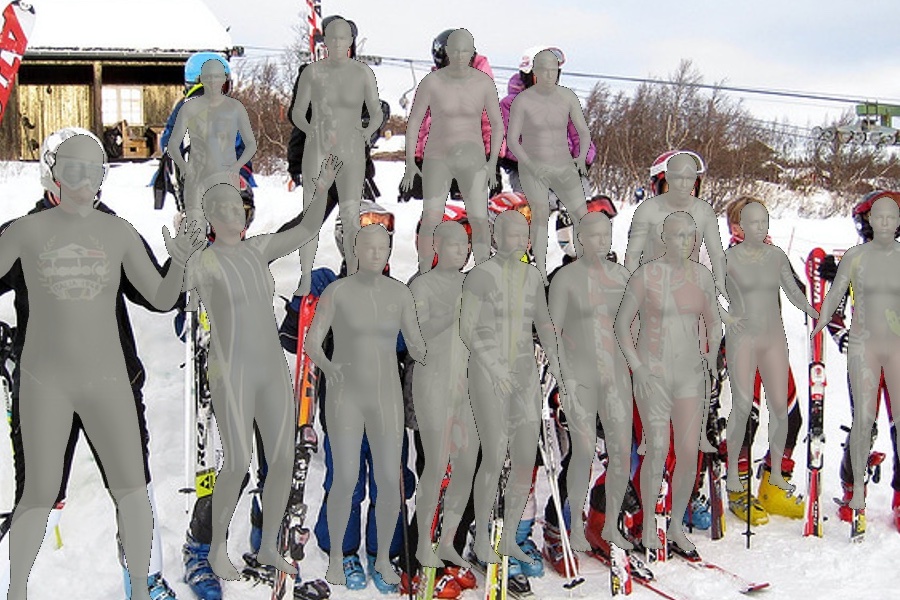}} \\
  \vspace{2.0mm}
  \mpage{0.23}{\includegraphics[width=\linewidth]{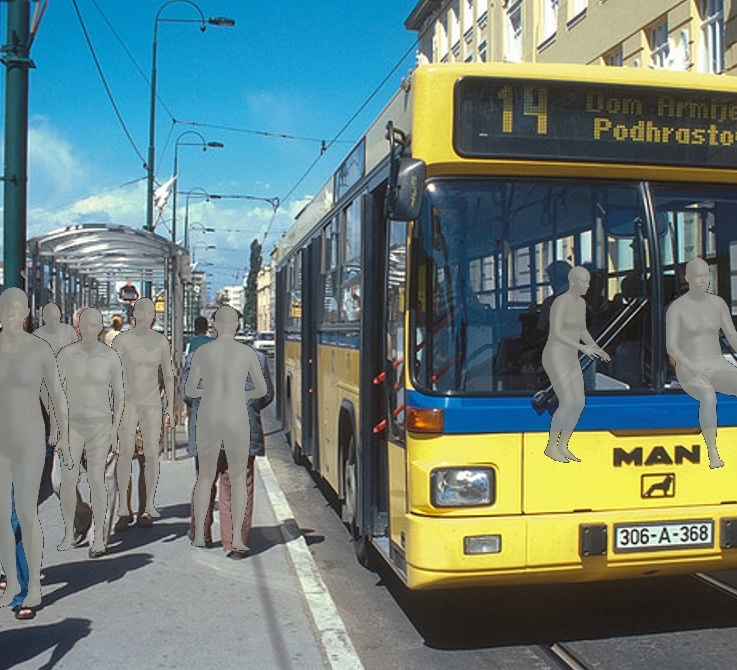}} \hfill
  \mpage{0.23}{\includegraphics[width=\linewidth]{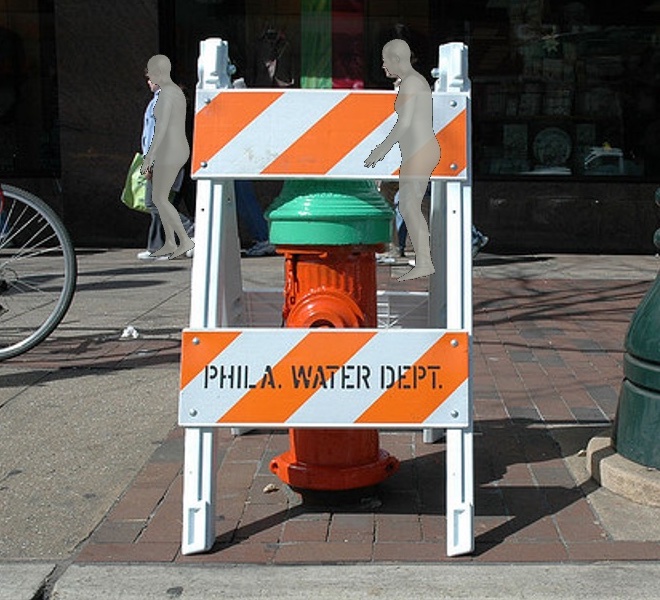}} \hfill
  \mpage{0.23}{\includegraphics[width=\linewidth]{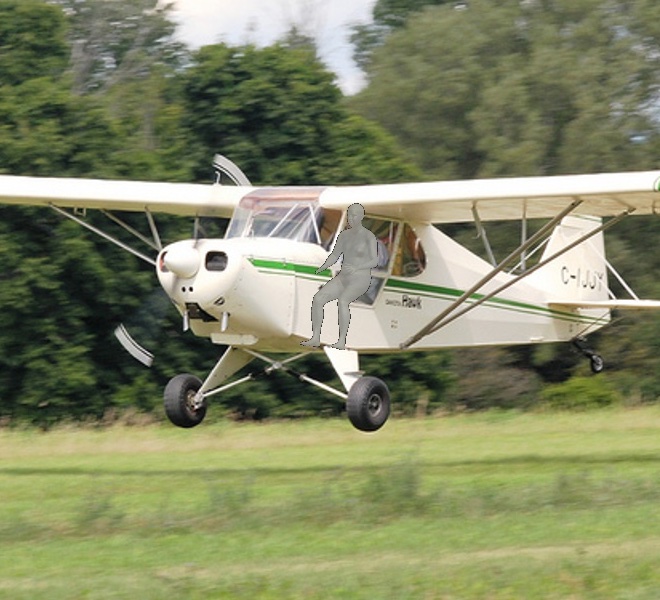}} \hfill
  \mpage{0.23}{\includegraphics[width=\linewidth]{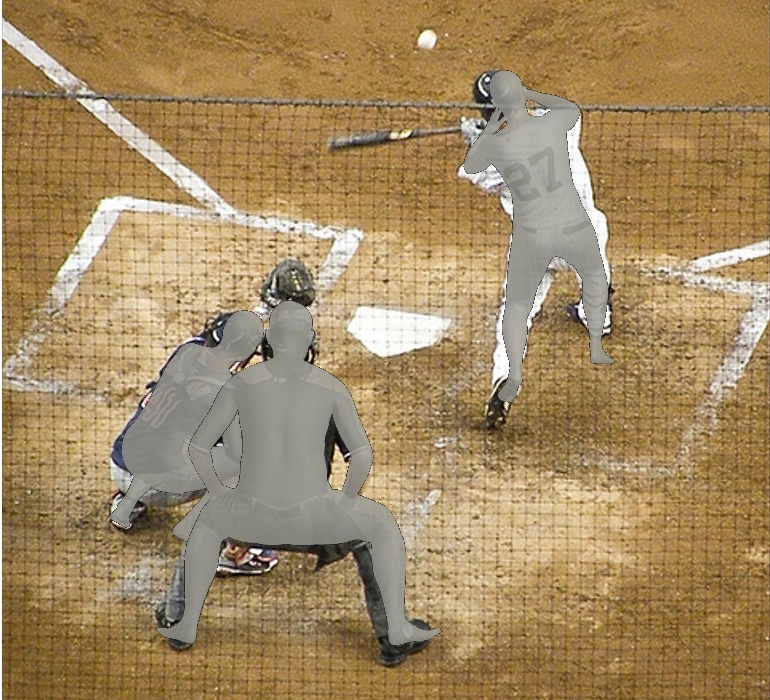}} \\
  \vspace{2.0mm}
  \mpage{0.23}{\includegraphics[width=\linewidth]{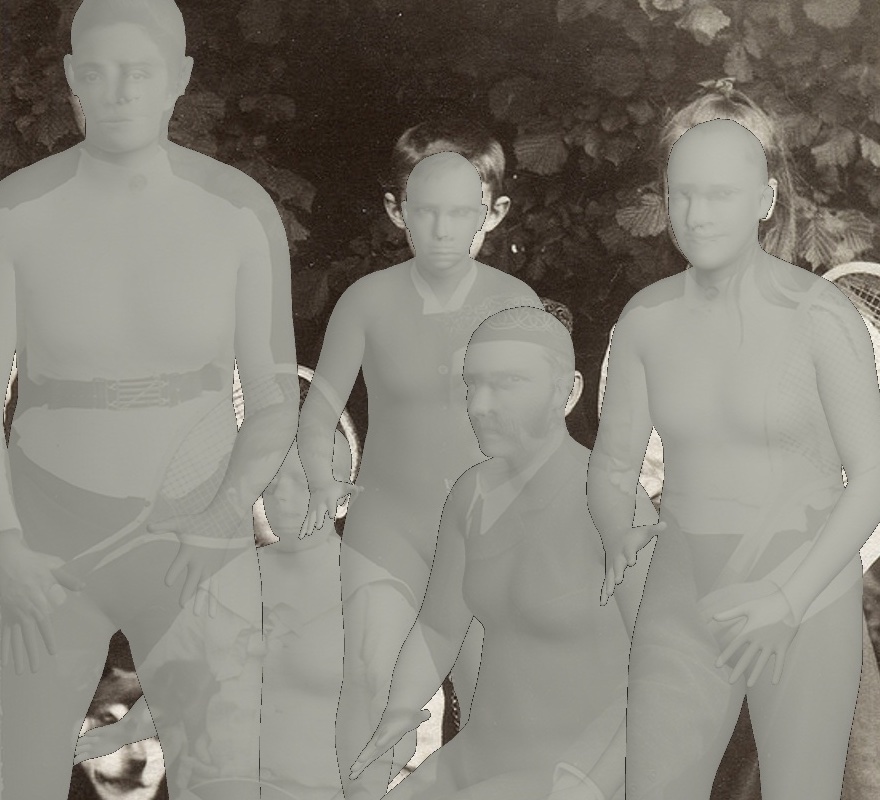}} \hfill
  \mpage{0.23}{\includegraphics[width=\linewidth]{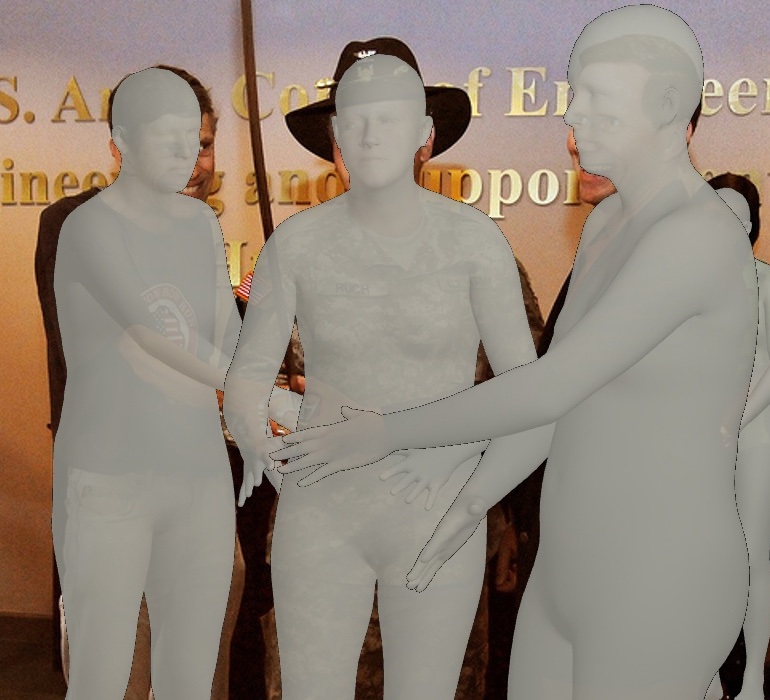}} \hfill
  \mpage{0.23}{\includegraphics[width=\linewidth]{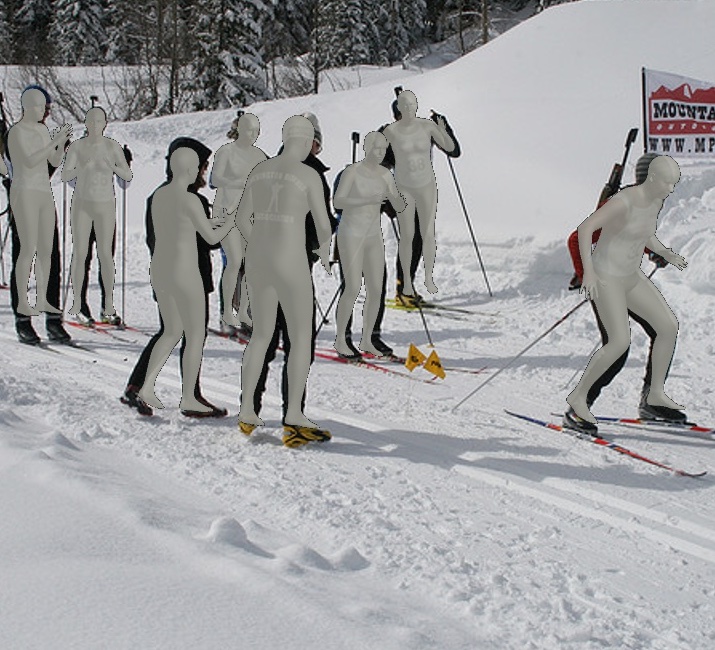}} \hfill
  \mpage{0.23}{\includegraphics[width=\linewidth]{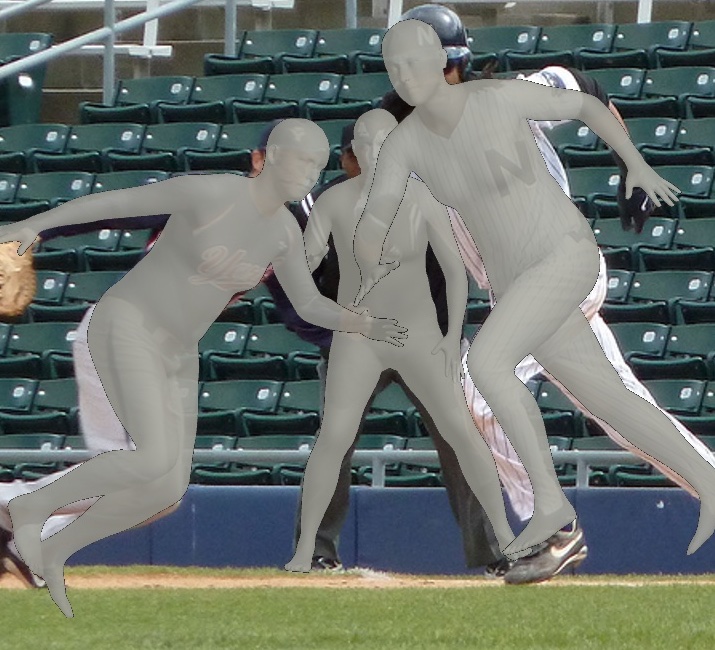}} \\
  \vspace{-1.5mm}
  \caption{
  \textbf{Visual results of occlusion handling.} 
  \revised{
  We present visual results on the CrowdPose dataset~\citep{li2019crowdpose}.
  Our results demonstrate the ability of the SPS-Net to recover plausible human bodies for the occluded person instances.
  }
  }
  \label{fig:occlusion}
  \end{center}
  \vspace{-8.0mm}
\end{figure*}

\begin{figure*}[t]
  \begin{center}
  \mpage{0.18}{\includegraphics[width=\linewidth]{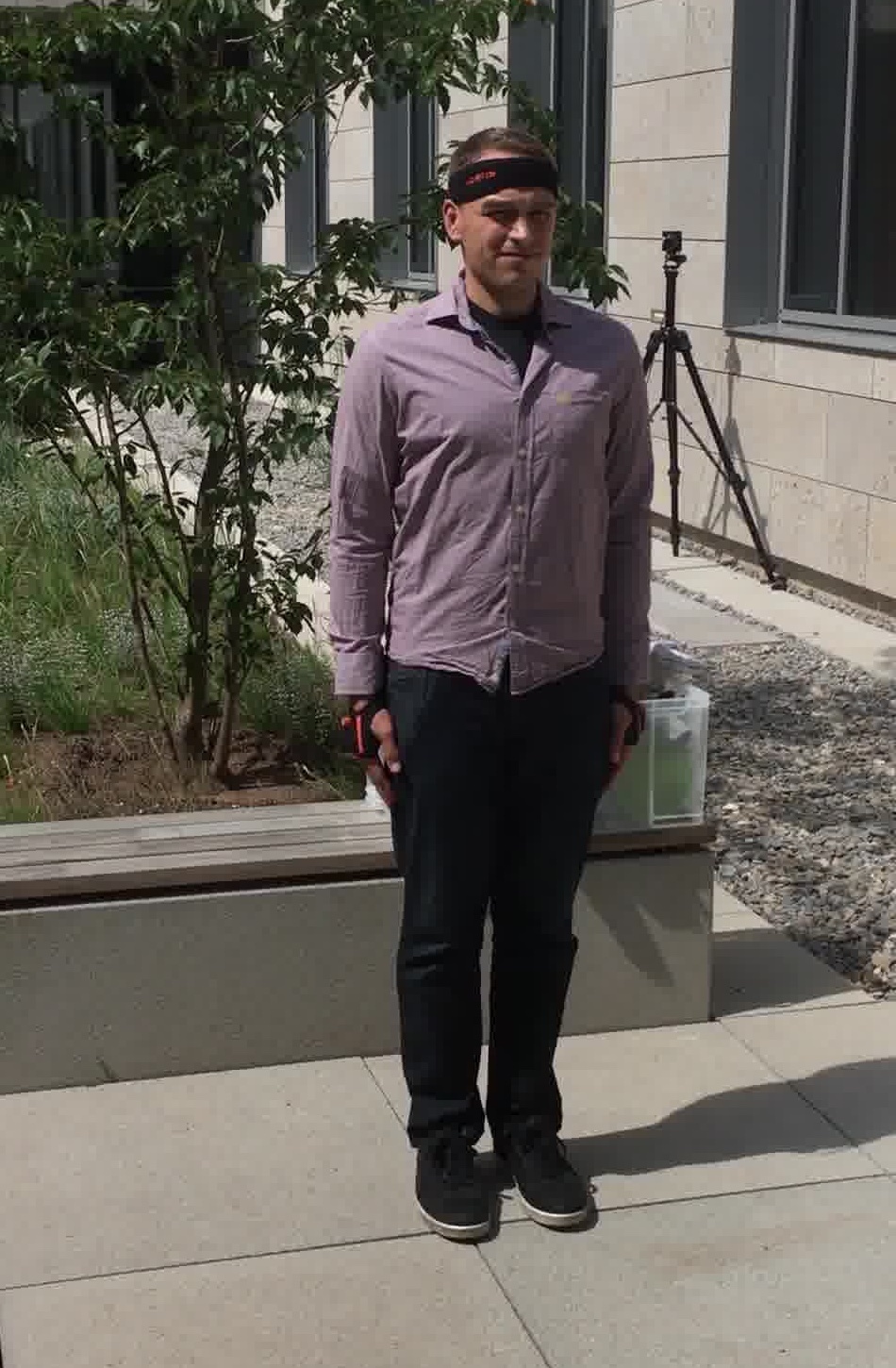}} \hfill
  \mpage{0.18}{\includegraphics[width=\linewidth]{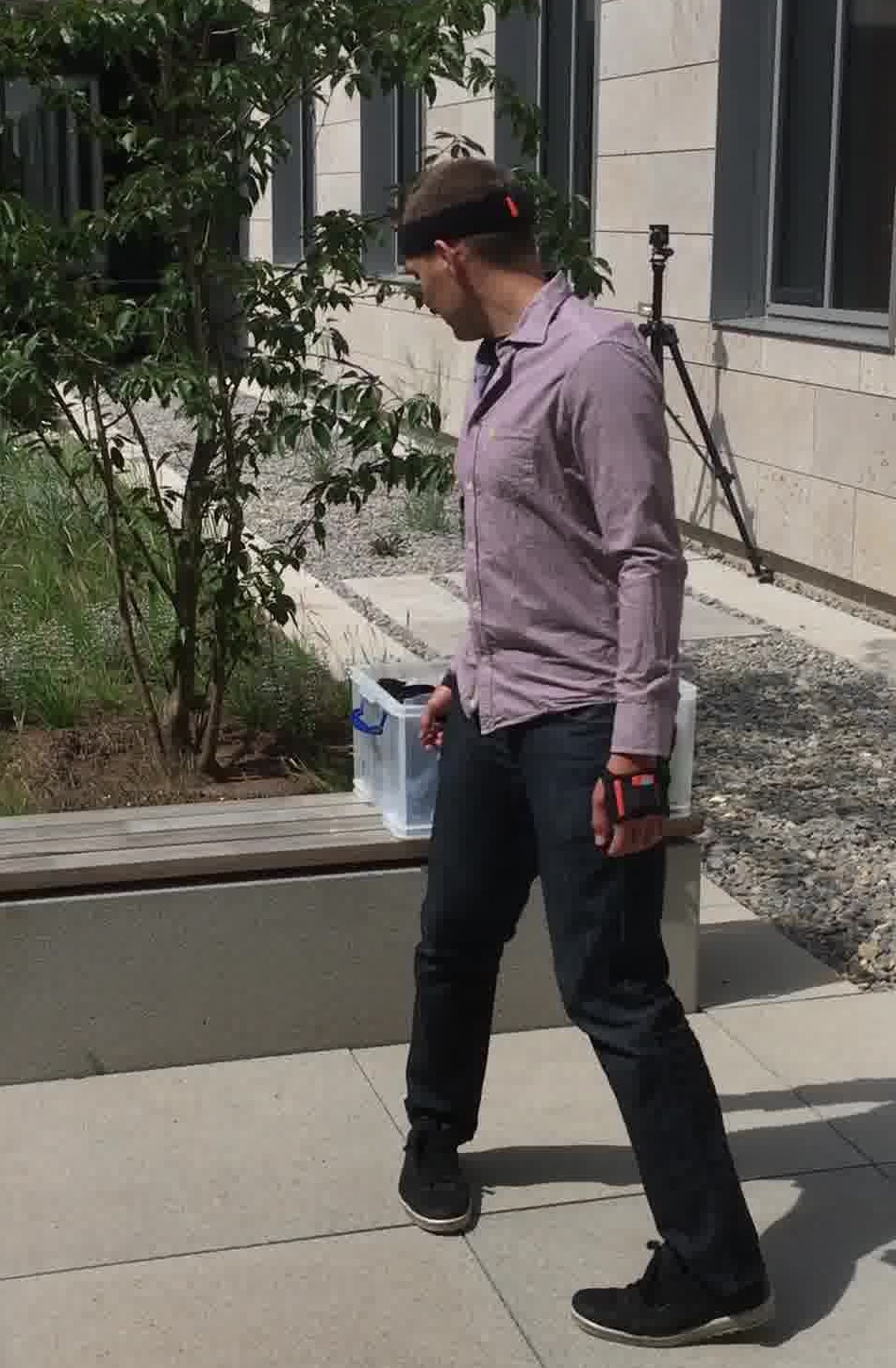}} \hfill
  \mpage{0.18}{\includegraphics[width=\linewidth]{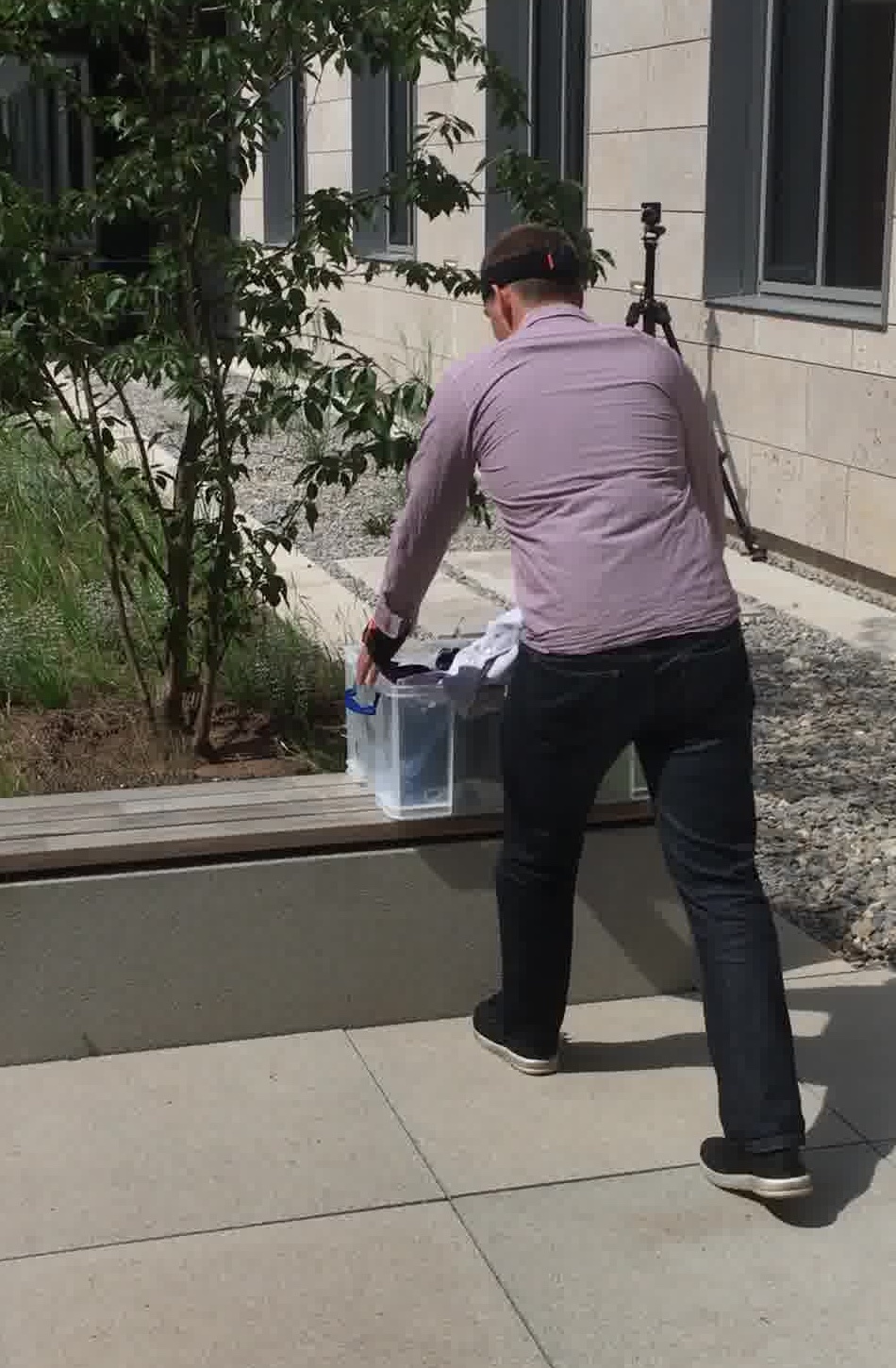}} \hfill
  \mpage{0.18}{\includegraphics[width=\linewidth]{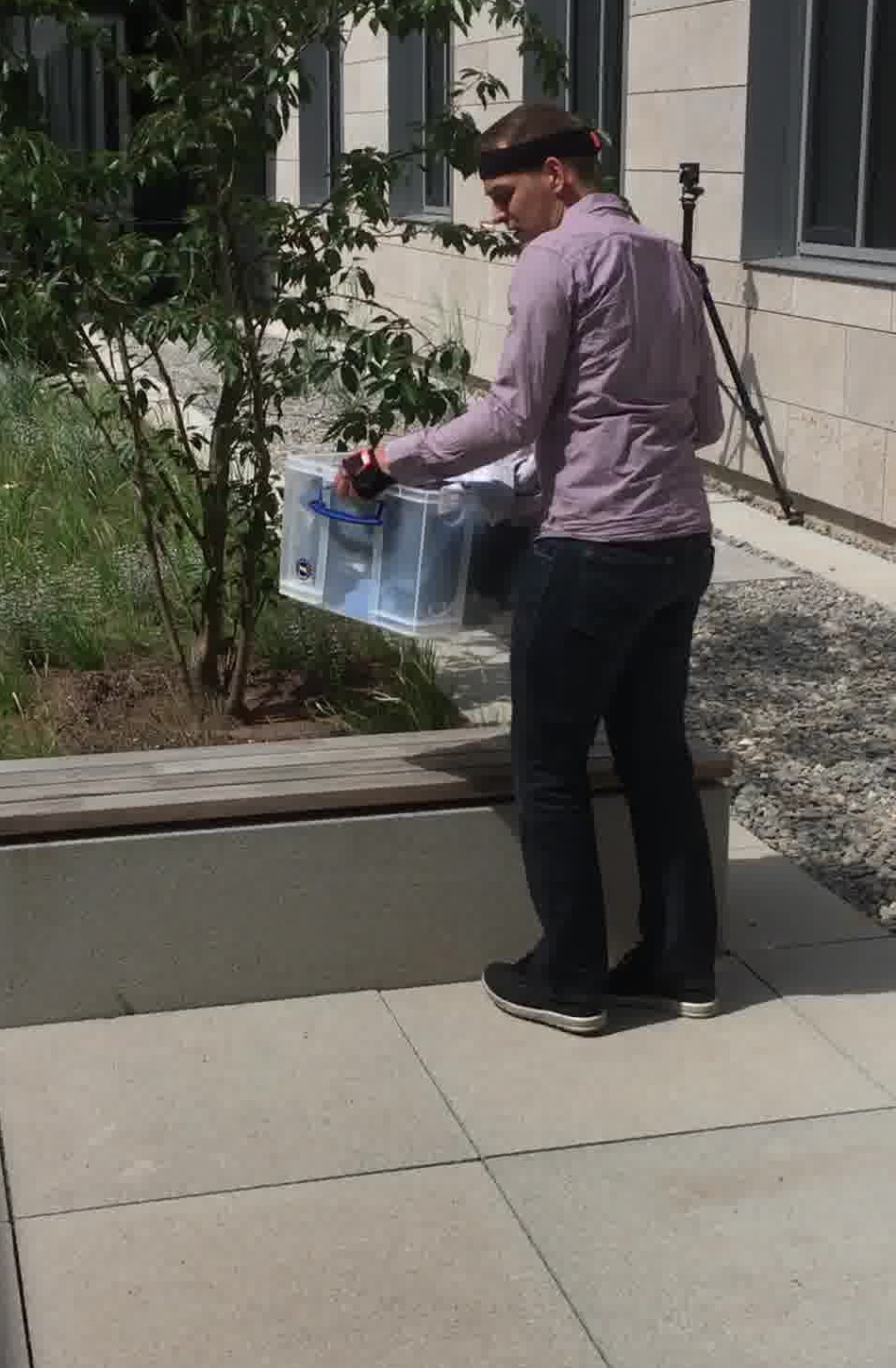}} \hfill
  \mpage{0.18}{\includegraphics[width=\linewidth]{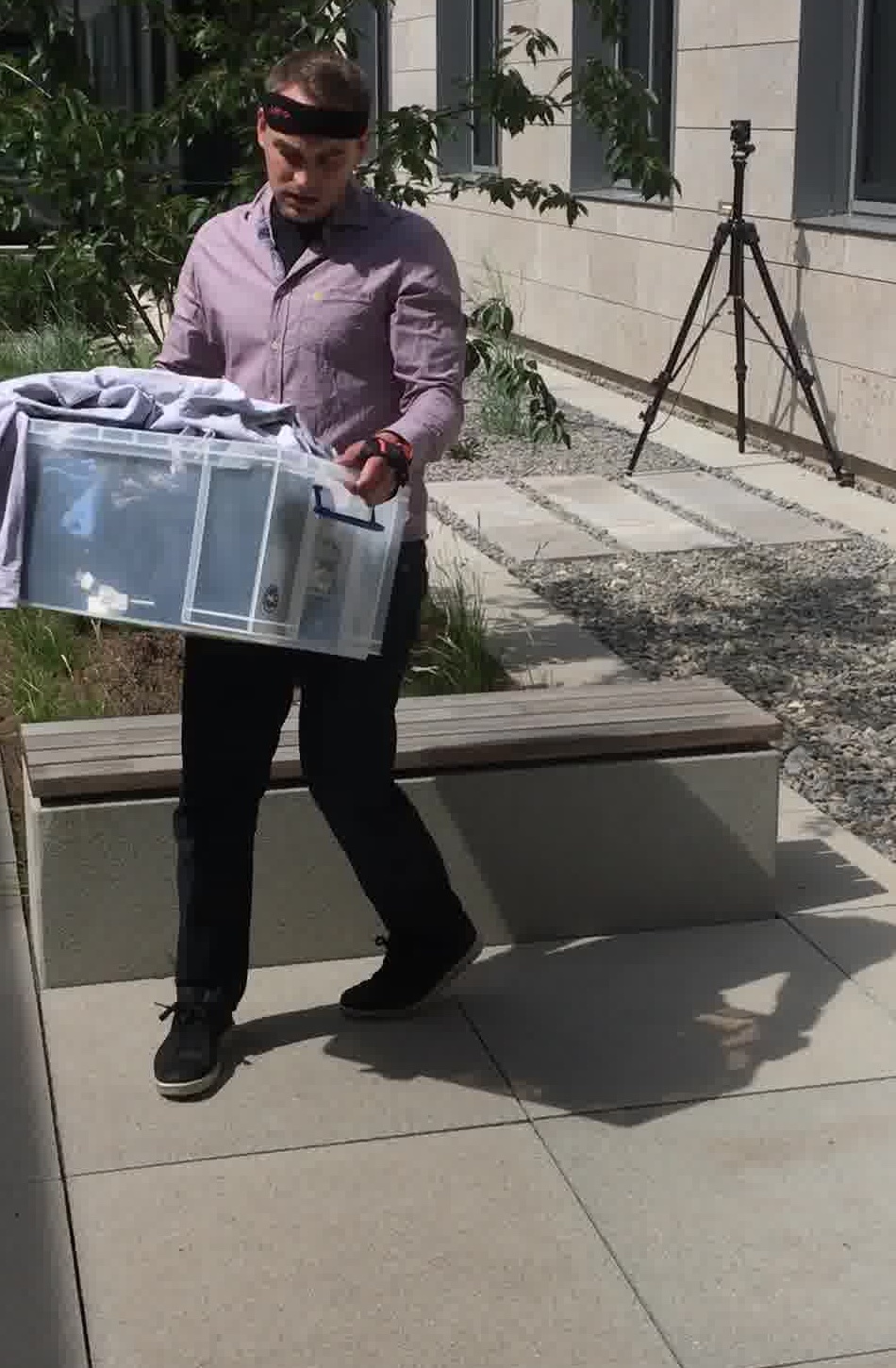}} \\
  \vspace{0.5mm}
  \mpage{0.18}{\includegraphics[width=\linewidth]{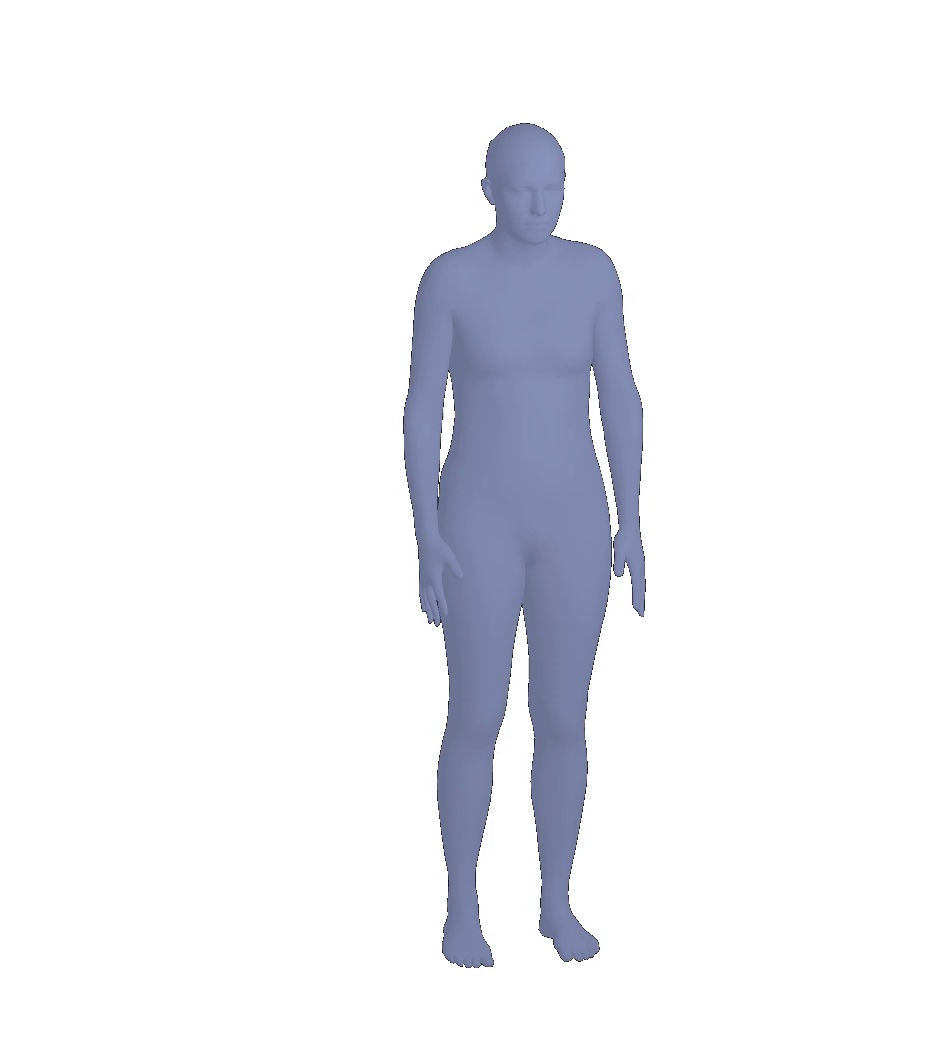}} \hfill
  \mpage{0.18}{\includegraphics[width=\linewidth]{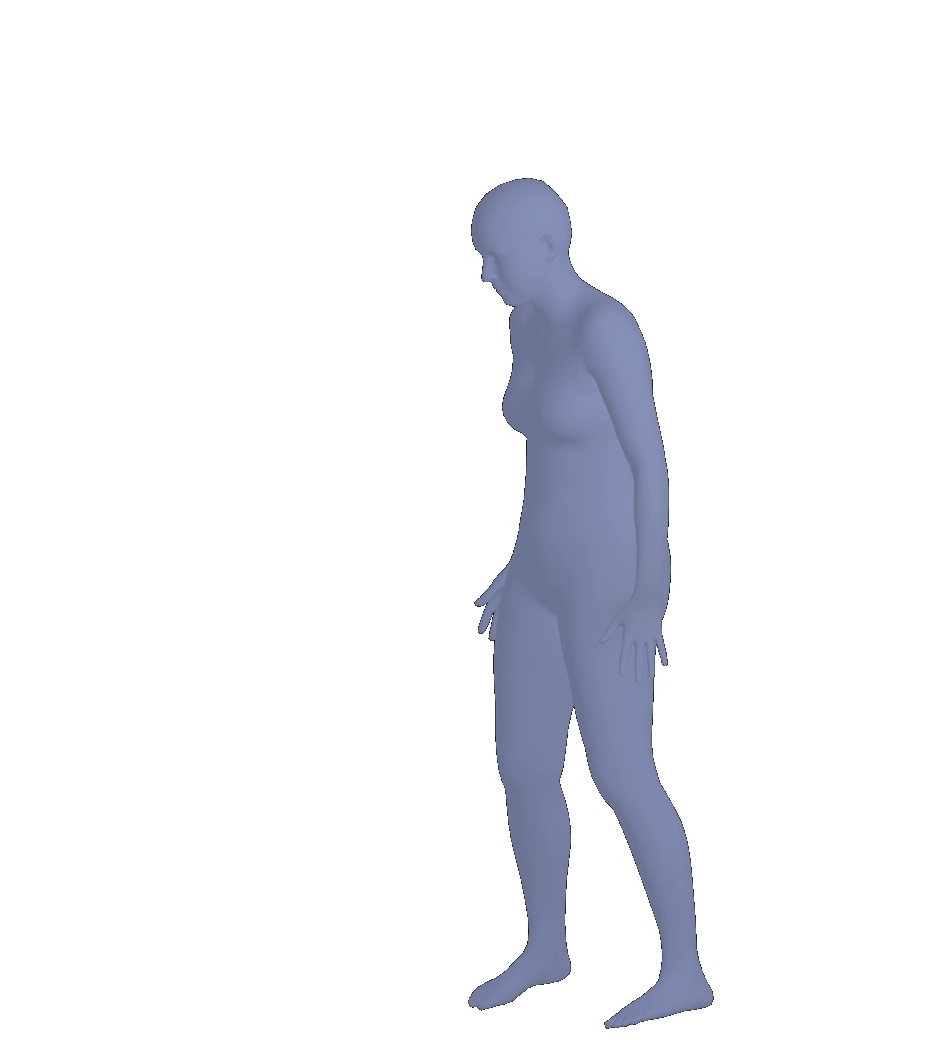}} \hfill
  \mpage{0.18}{\includegraphics[width=\linewidth]{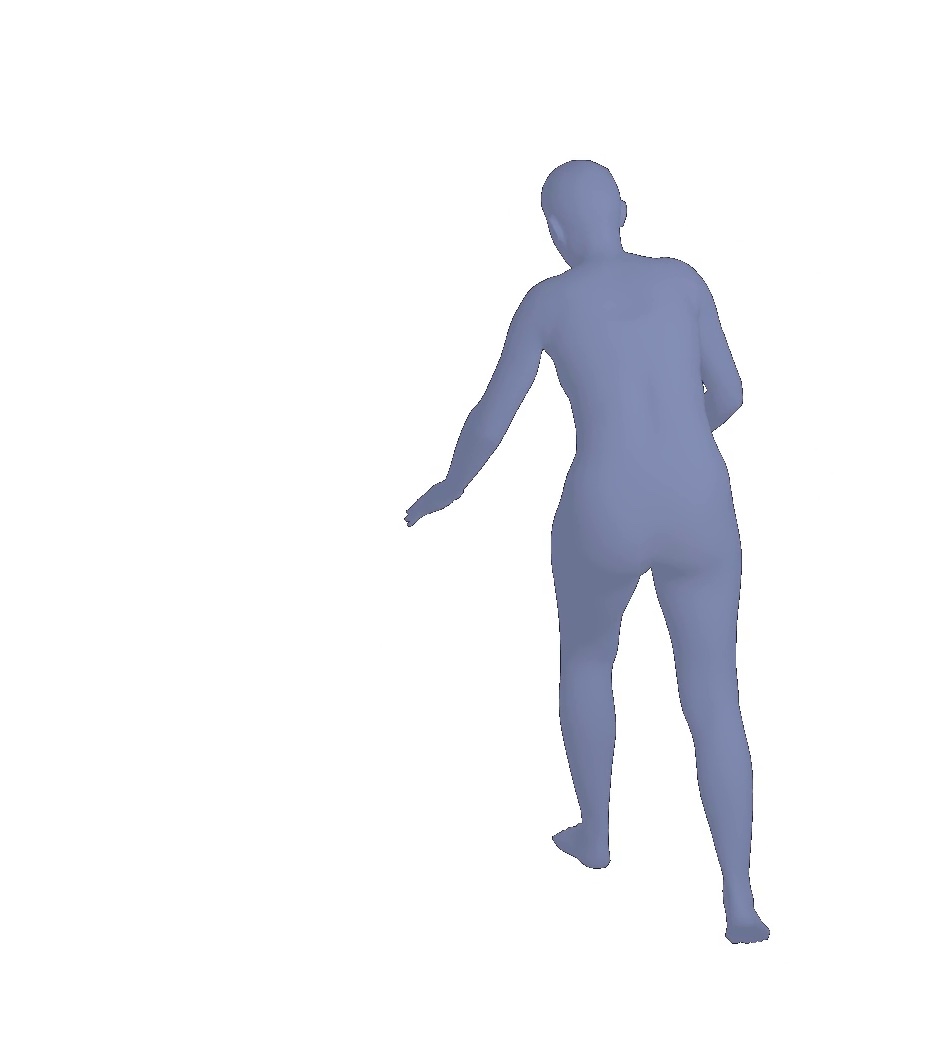}} \hfill
  \mpage{0.18}{\includegraphics[width=\linewidth]{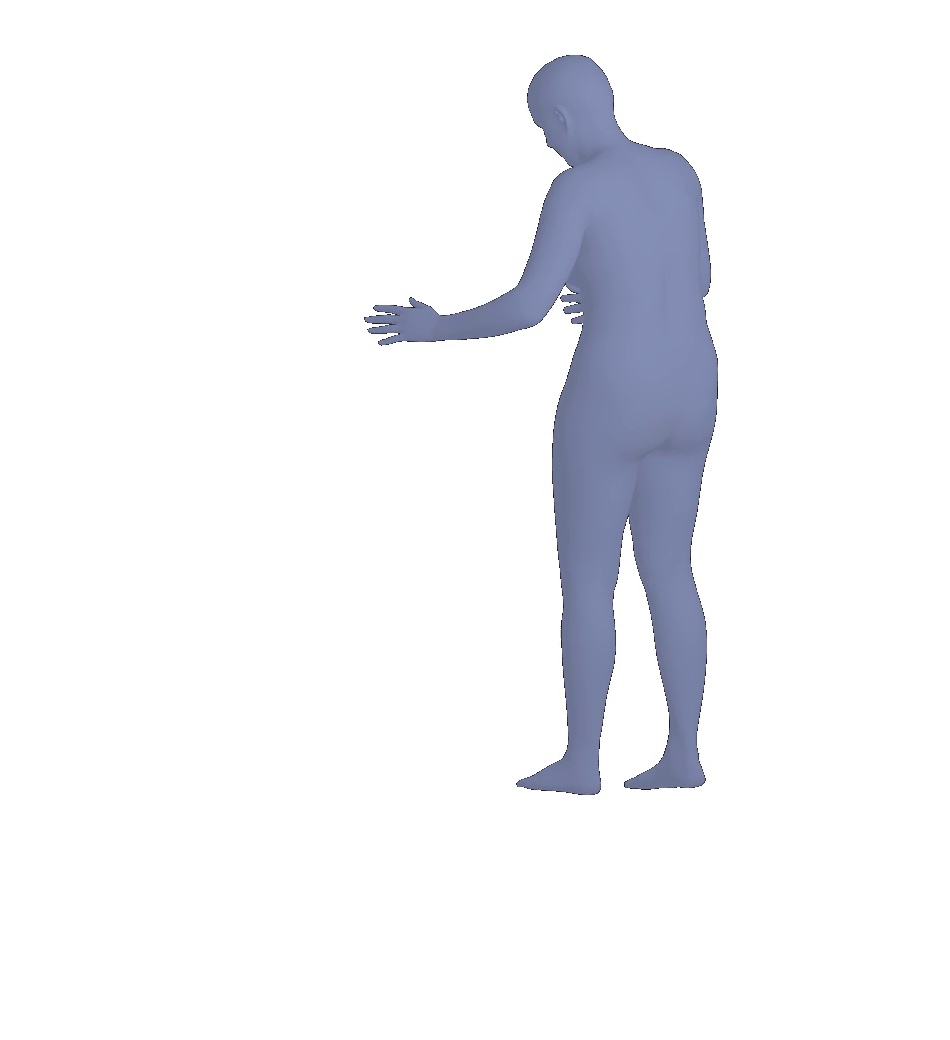}} \hfill
  \mpage{0.18}{\includegraphics[width=\linewidth]{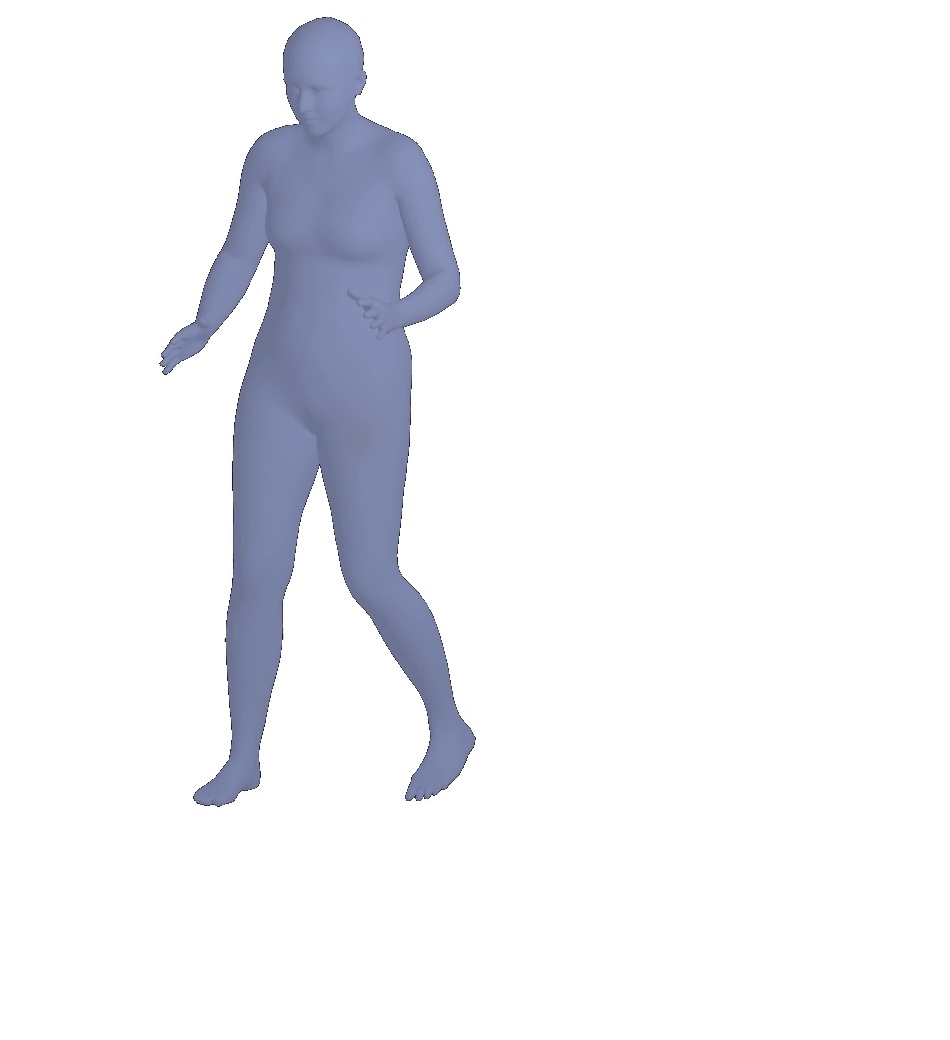}} \\
  \vspace{-1.5mm}
  \mpage{0.18}{\includegraphics[width=\linewidth]{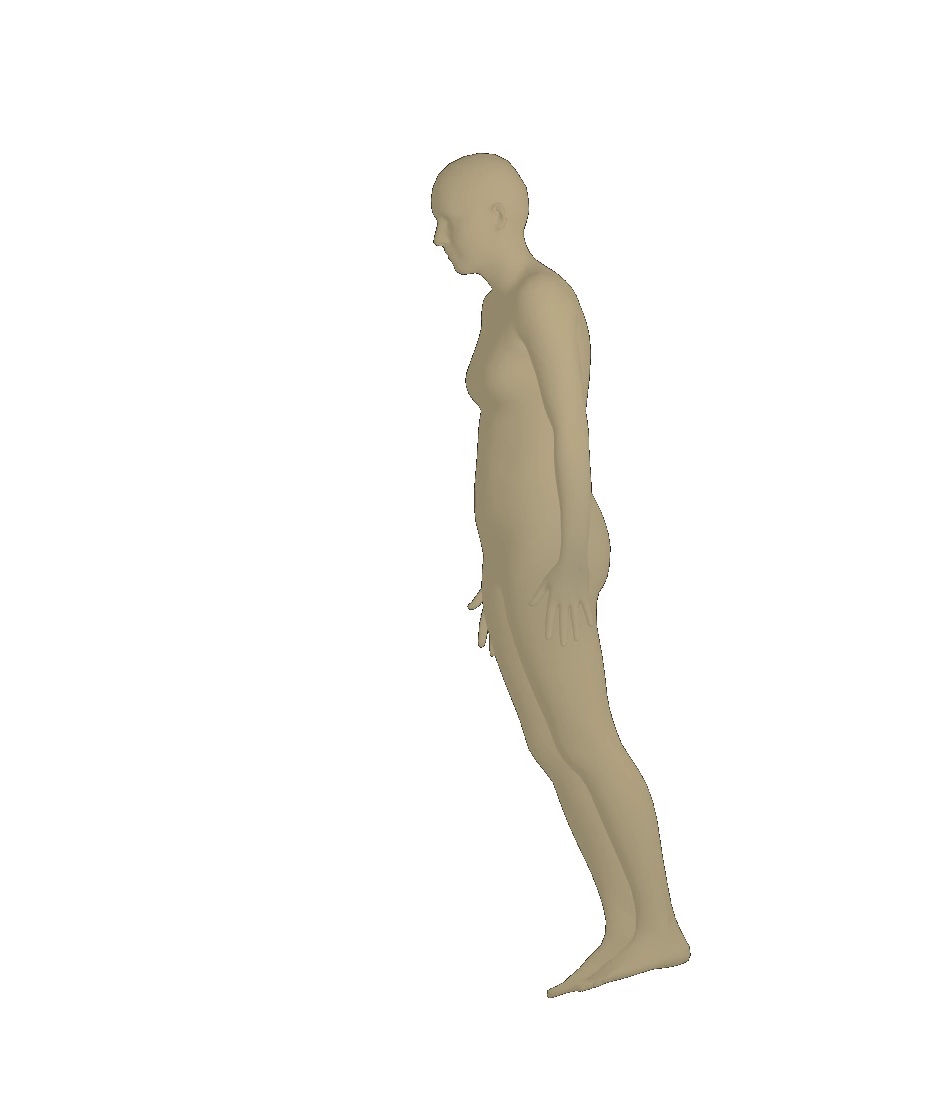}} \hfill
  \mpage{0.18}{\includegraphics[width=\linewidth]{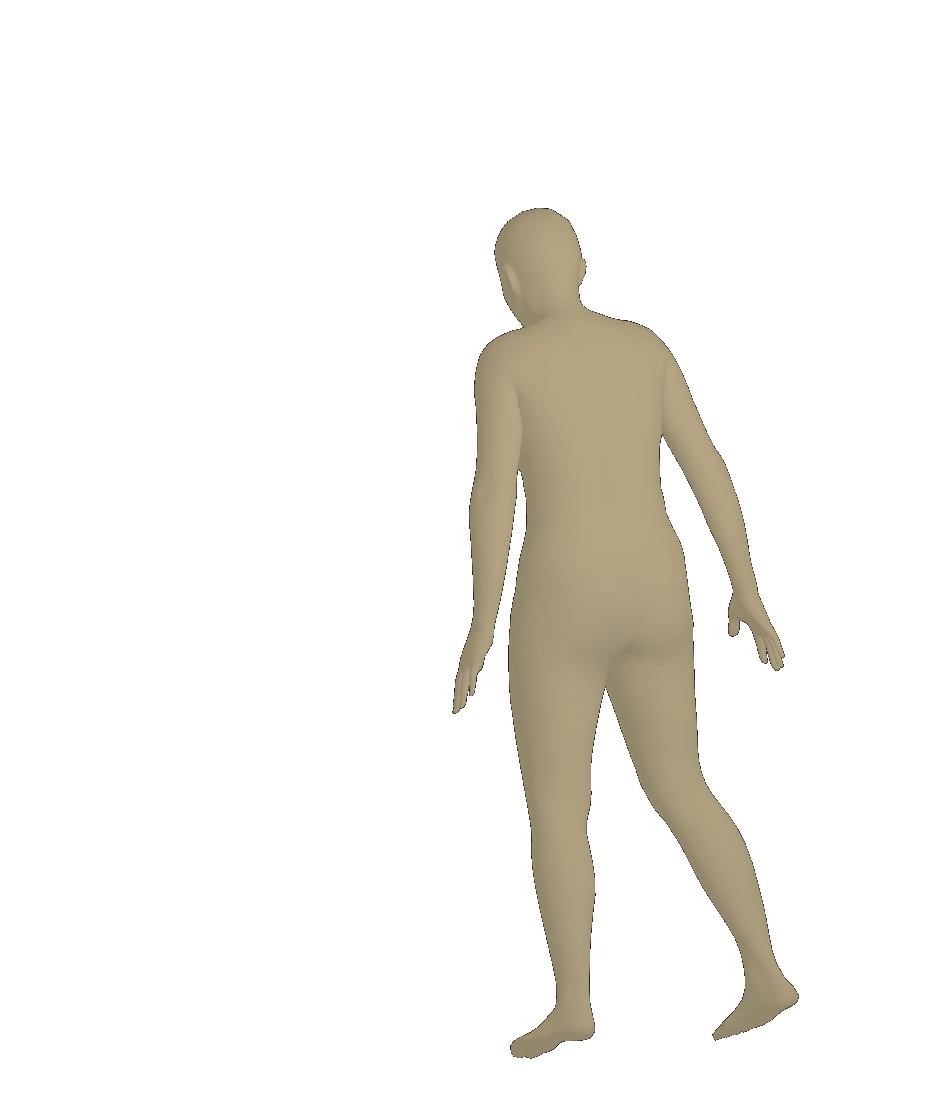}} \hfill
  \mpage{0.18}{\includegraphics[width=\linewidth]{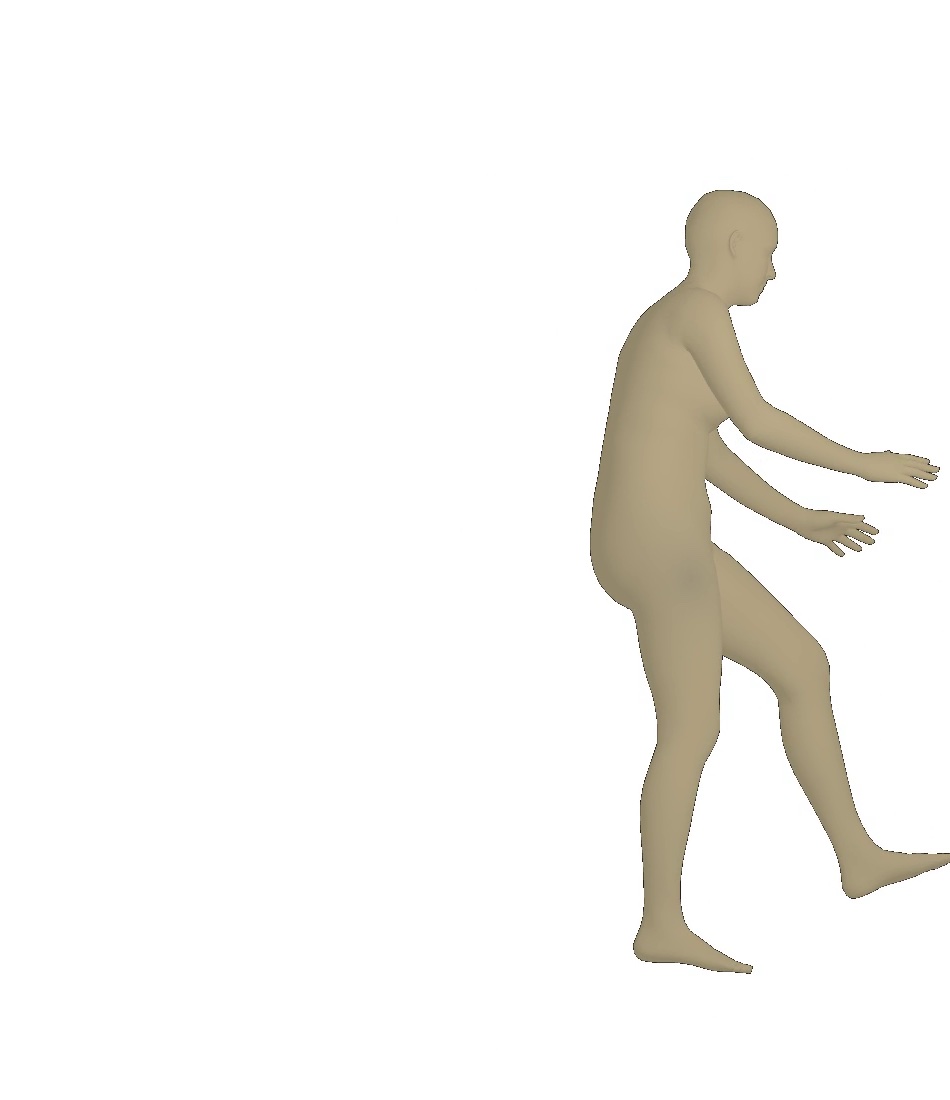}} \hfill
  \mpage{0.18}{\includegraphics[width=\linewidth]{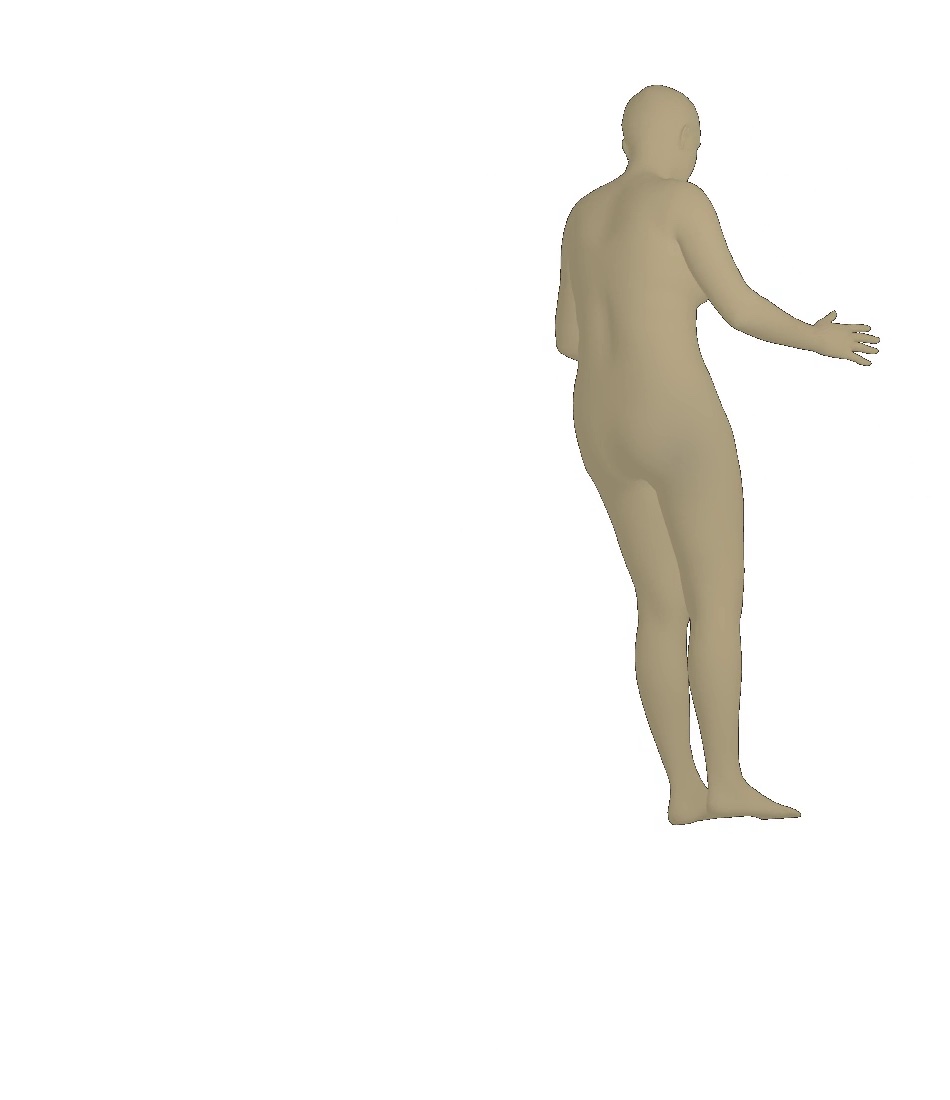}} \hfill
  \mpage{0.18}{\includegraphics[width=\linewidth]{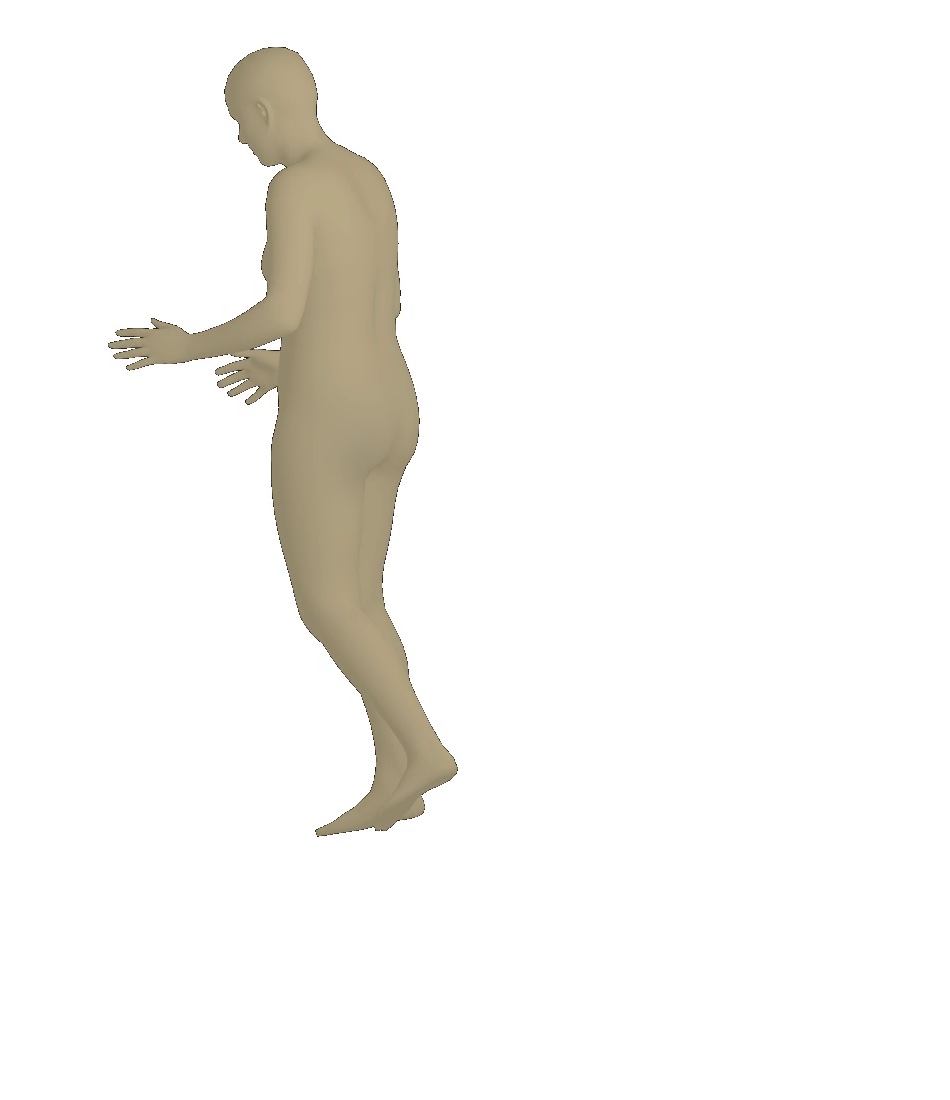}} \\
  \vspace{-3.0mm}
  \caption{
  \textbf{Qualitative results of 3D human pose and shape estimation.} 
  We visualize the 3D human body from different viewpoints recovered by our SPS-Net on the 3DPW~\citep{3DPW} test set.
  }
  \label{fig:mesh}
  \end{center}
  \vspace{-6.0mm}
\end{figure*}

\begin{figure*}[t]
  \begin{center}
  \mpage{0.01}{\raisebox{2pt}{\rotatebox{90}{Ours w/o $\mathcal{L}_\mathrm{camera}$}}} \hfill
  \mpage{0.14}{\includegraphics[width=\linewidth]{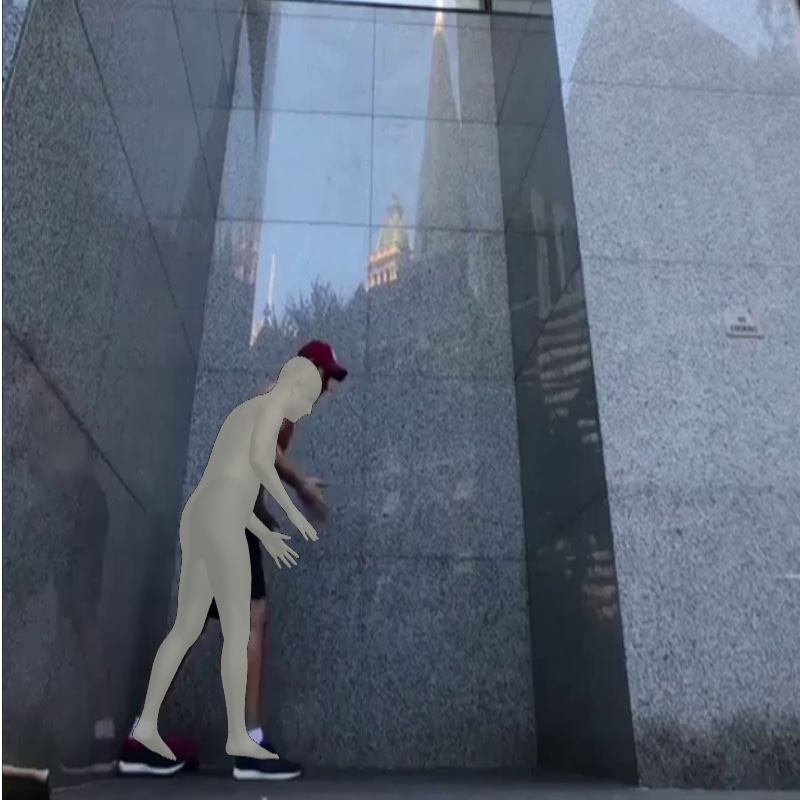}} \hfill
  \mpage{0.14}{\includegraphics[width=\linewidth]{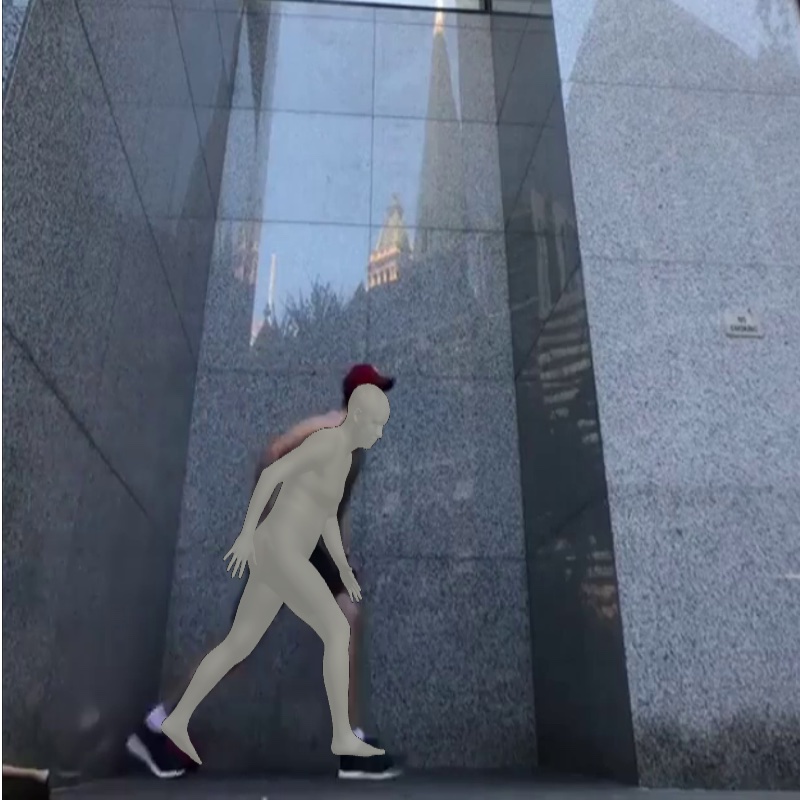}} \hfill
  \mpage{0.14}{\includegraphics[width=\linewidth]{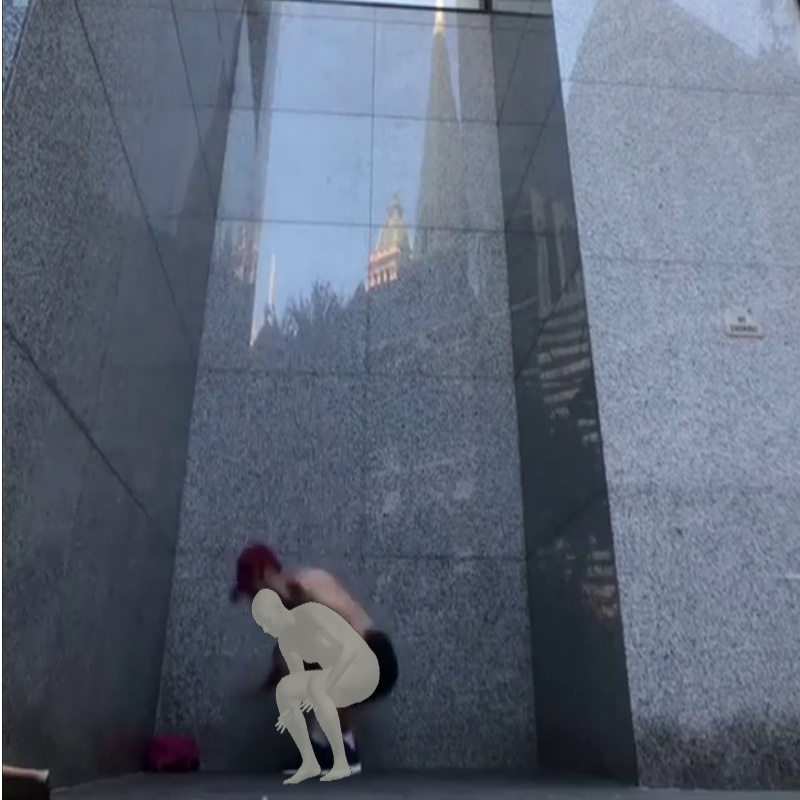}} \hfill
  \hspace{2.0mm}
  \mpage{0.01}{\raisebox{2pt}{\rotatebox{90}{Ours w/o $\mathcal{L}_\mathrm{mask}$}}} \hfill
  \mpage{0.14}{\includegraphics[width=\linewidth]{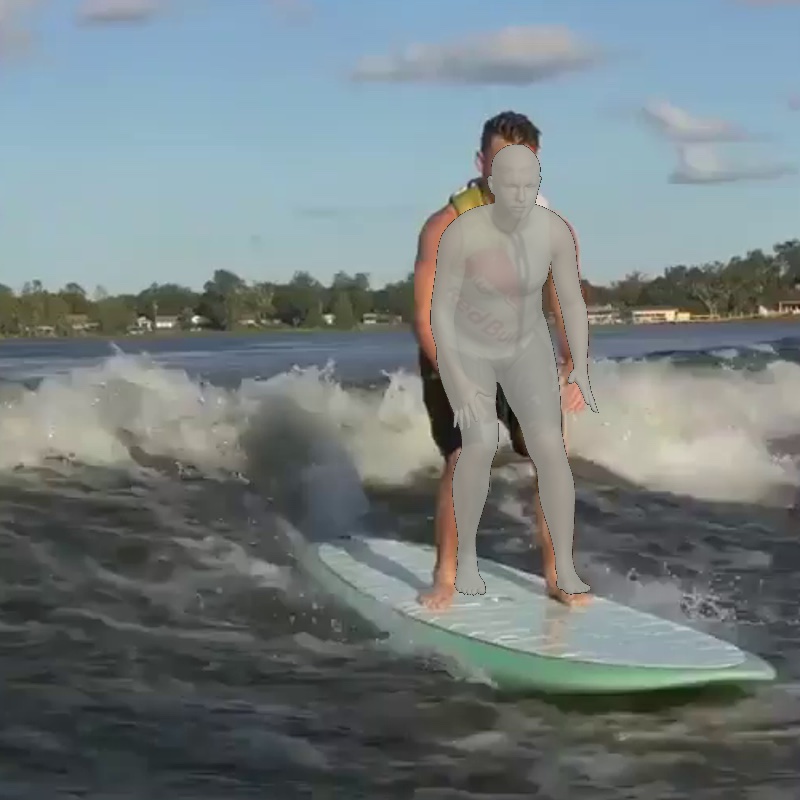}} \hfill
  \mpage{0.14}{\includegraphics[width=\linewidth]{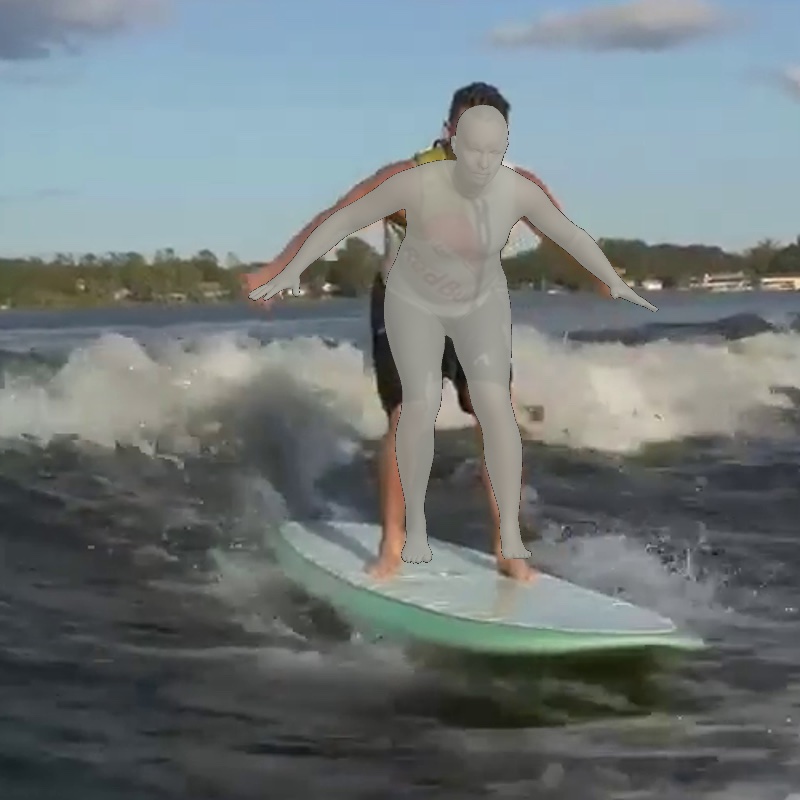}} \hfill
  \mpage{0.14}{\includegraphics[width=\linewidth]{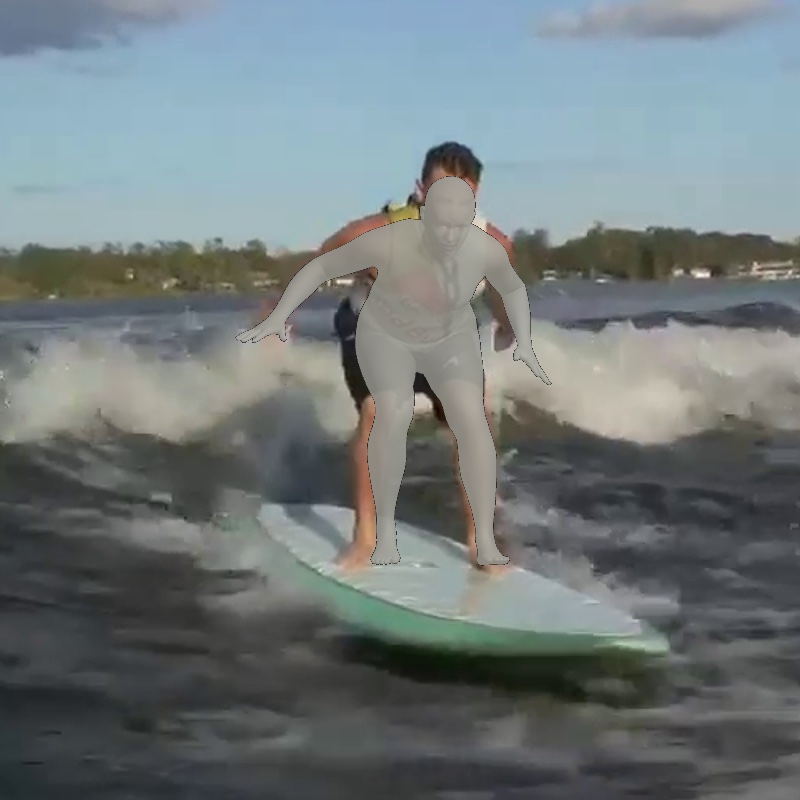}} \\
  \vspace{1.5mm}
  \mpage{0.01}{\raisebox{2pt}{\rotatebox{90}{Ours}}} \hfill
  \mpage{0.14}{\includegraphics[width=\linewidth]{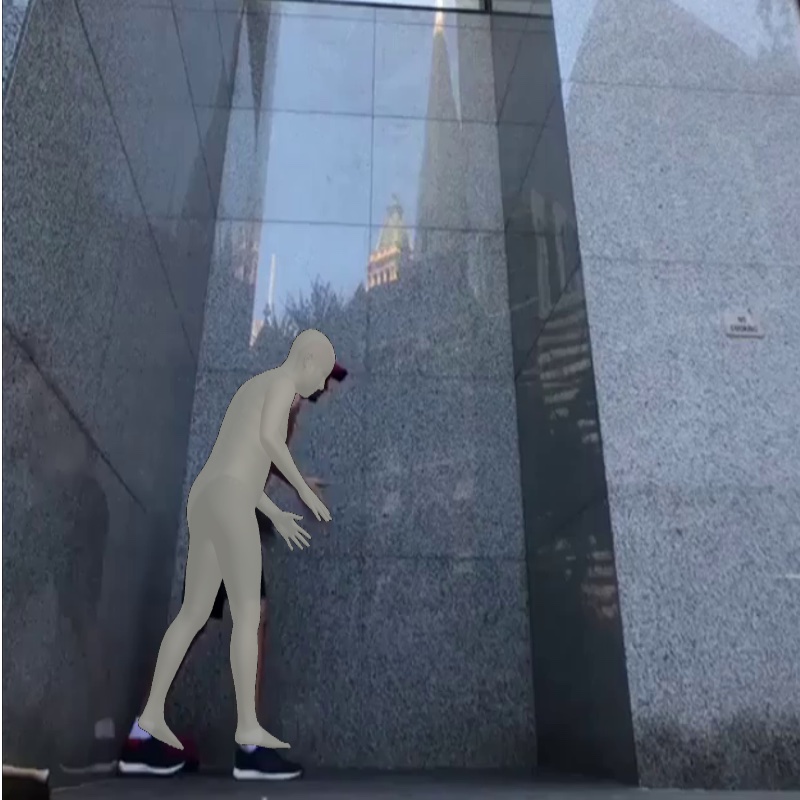}} \hfill
  \mpage{0.14}{\includegraphics[width=\linewidth]{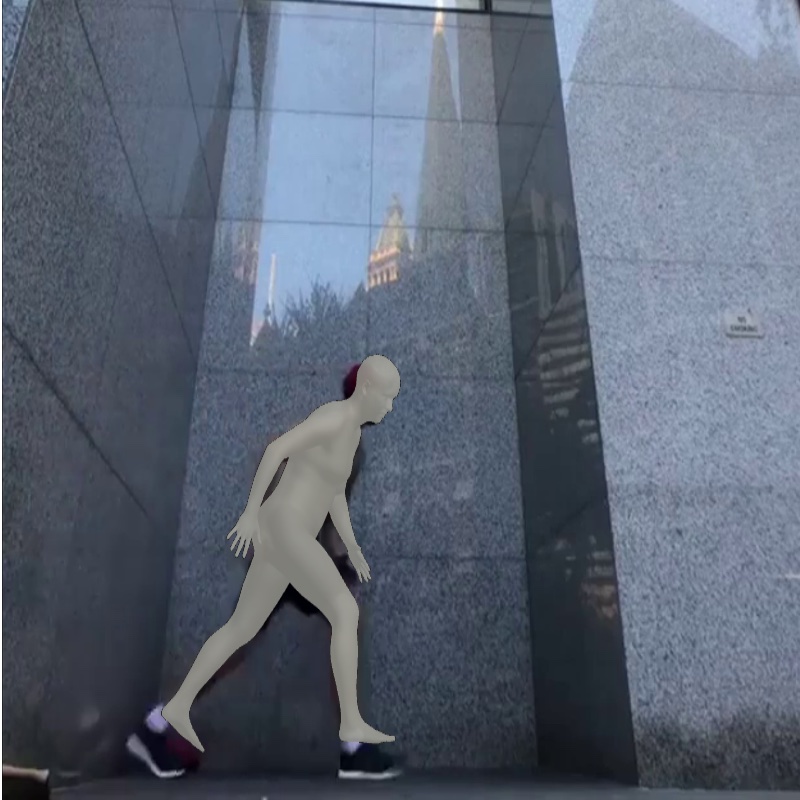}} \hfill
  \mpage{0.14}{\includegraphics[width=\linewidth]{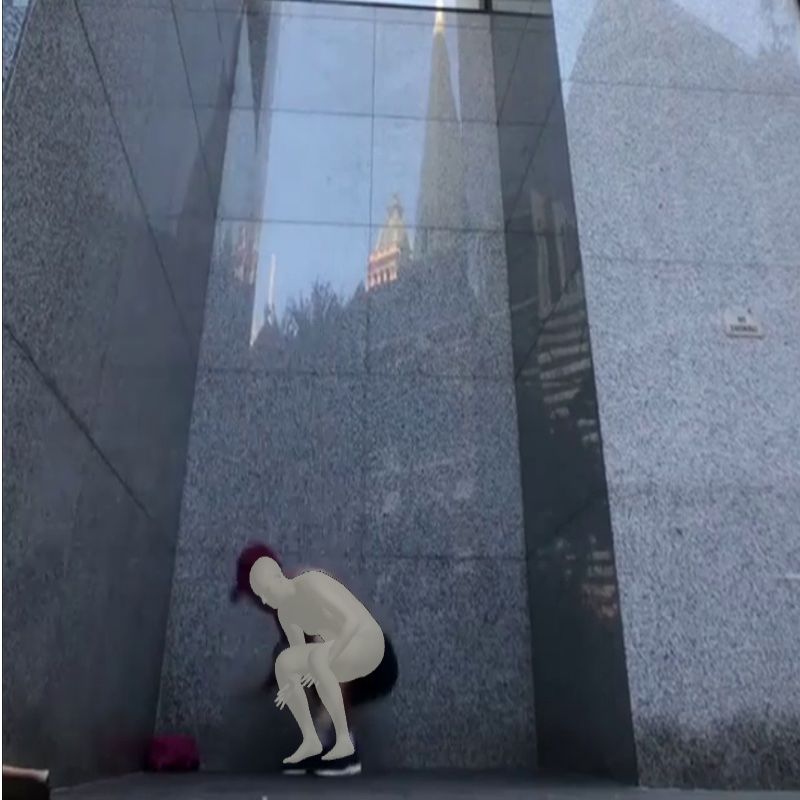}} \hfill
  \hspace{2.0mm}
  \mpage{0.01}{\raisebox{2pt}{\rotatebox{90}{Ours}}} \hfill
  \mpage{0.14}{\includegraphics[width=\linewidth]{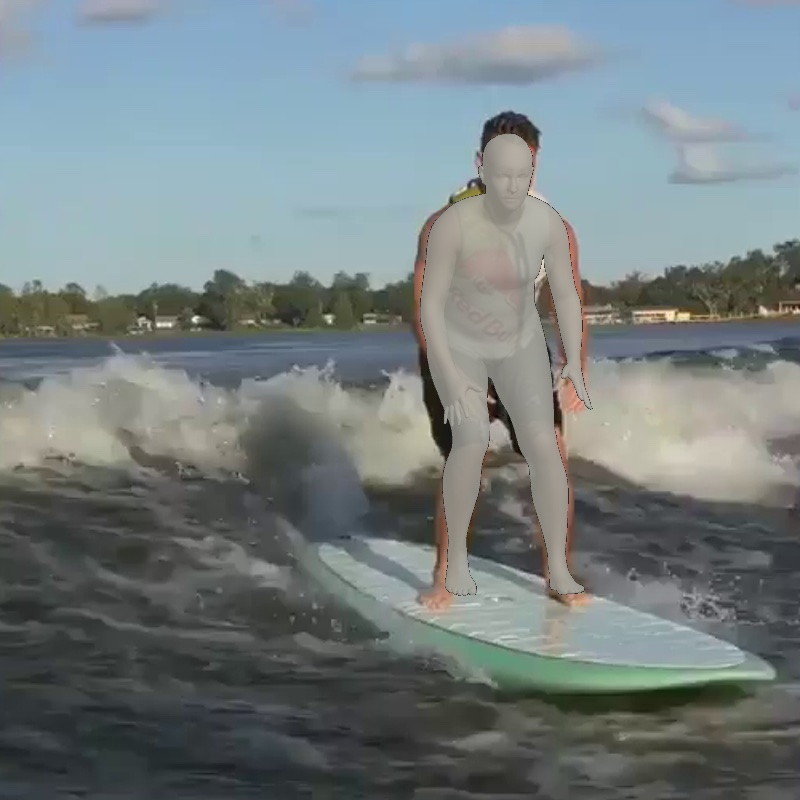}} \hfill
  \mpage{0.14}{\includegraphics[width=\linewidth]{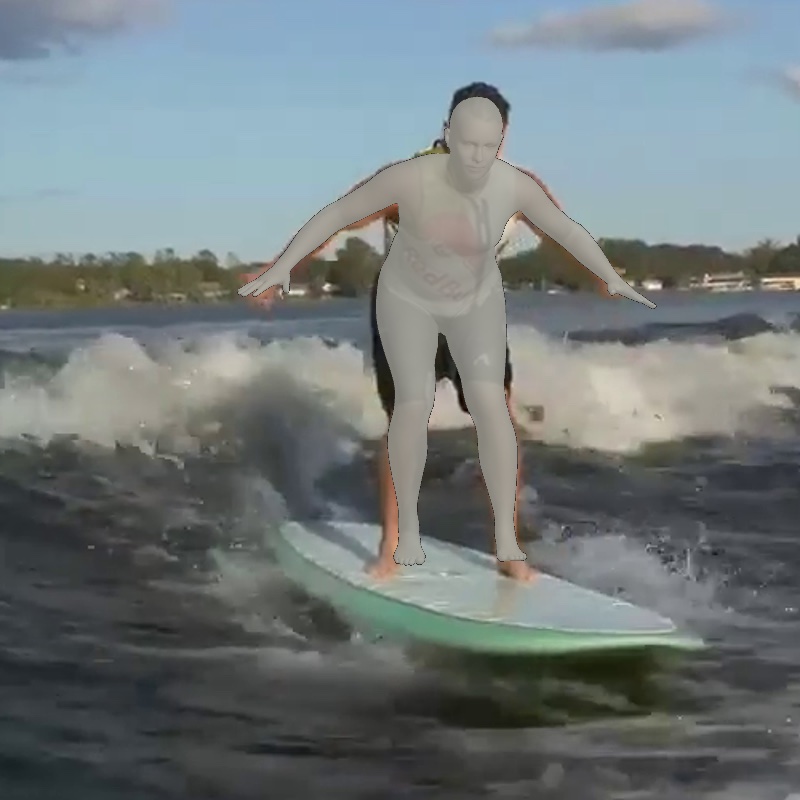}} \hfill
  \mpage{0.14}{\includegraphics[width=\linewidth]{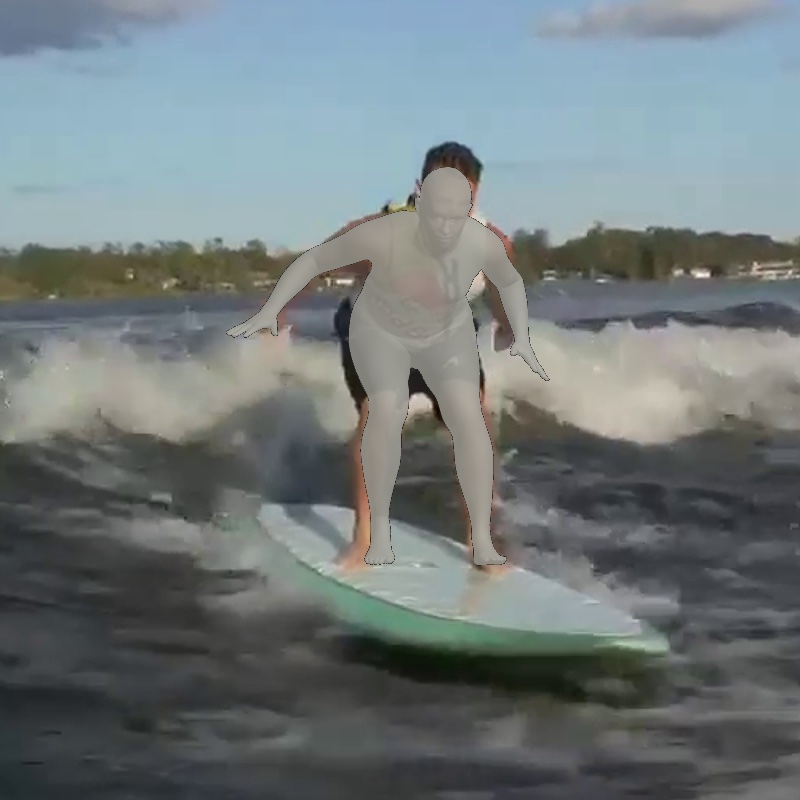}} \\
  \vspace{-1.5mm}
  \caption{
  \textbf{Visual comparisons with our variant methods.} 
  (\emph{Left}) Visual comparisons with the Ours w/o $\mathcal{L}_\mathrm{camera}$ method.
  (\emph{Right}) Visual comparisons with the Ours w/o $\mathcal{L}_\mathrm{mask}$ approach.
  }
  \label{fig:ablation}
  \end{center}
  \vspace{-6.0mm}
\end{figure*}

\setlength{\twoimg}{0.495\textwidth}
\begin{figure*}[t]
  \begin{subfigure}[b]{\twoimg}
    \begin{center}
      \includegraphics[width=0.95\textwidth]{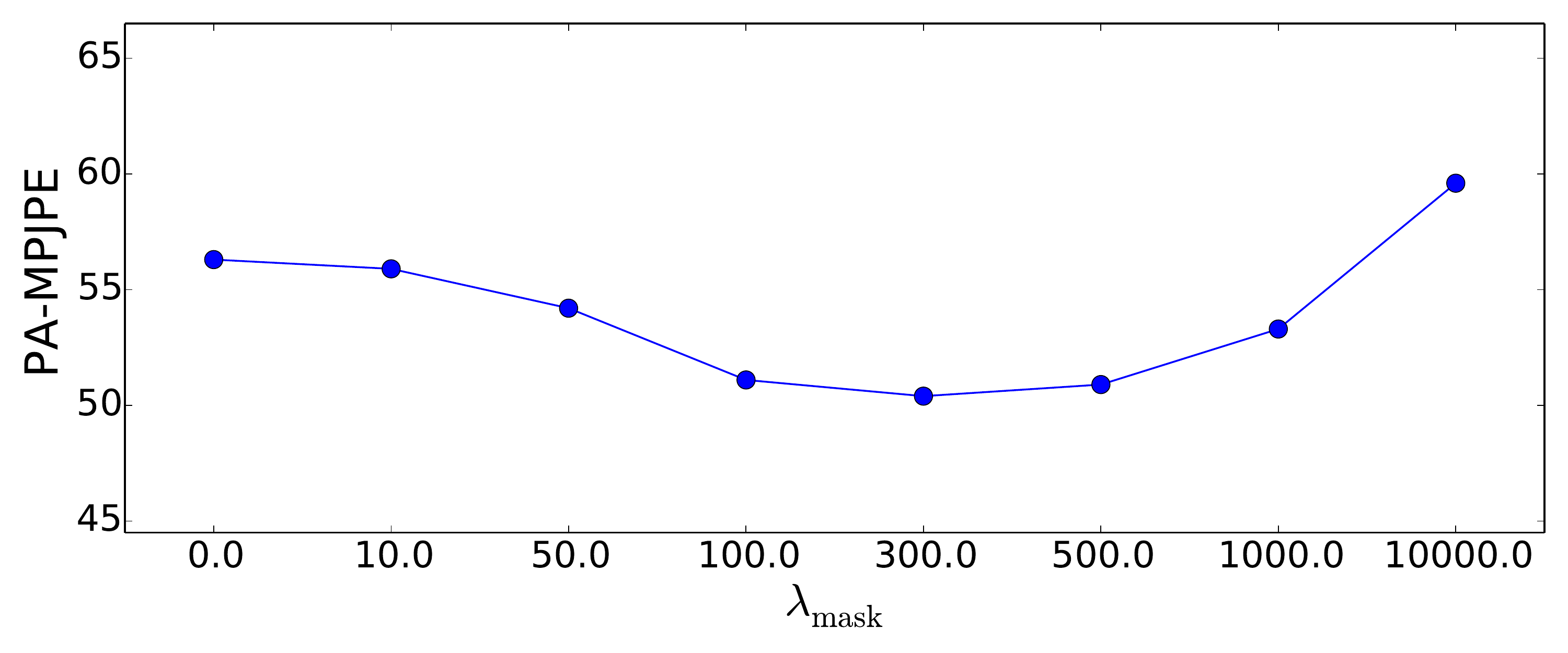}
      \vspace{-2.0mm}
    %   \caption*{Mask loss $\lambda_\mathrm{mask}$.}
    \end{center}
  \end{subfigure}
  \hfill
  \begin{subfigure}[b]{\twoimg}
    \begin{center}
      \includegraphics[width=0.95\textwidth]{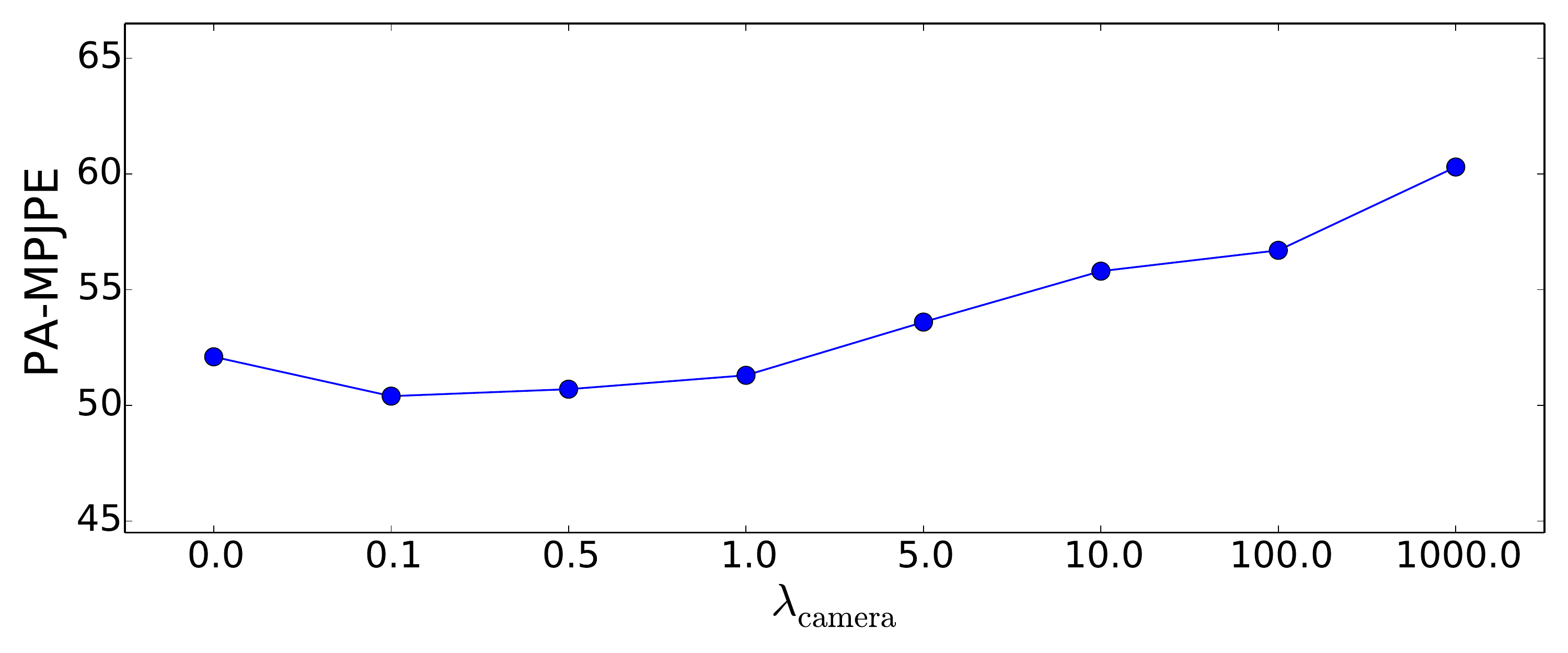}
      \vspace{-2.0mm}
    %   \caption*{Camera parameter consistency loss $\lambda_\mathrm{camera}$.}
    \end{center}
  \end{subfigure}
  \vspace{-2.0mm}
  \caption{
  \textbf{Sensitivity analysis of hyperparameters.}
  We report the PA-MPJPE results of our method on the 3DPW~\citep{3DPW} dataset.
  Experimental results show that the performance of our SPS-Net is stable when the hyperparameters are set within a suitable range.
  }
  \label{fig:hyperparameter}
  \vspace{-3.0mm}
\end{figure*}

\vspace{-3.0mm}
\subsubsection{Evaluation metrics}

We use the procrustes aligned mean per joint position error (PA-MPJPE), mean per joint position error (MPJPE), percentage of correct keypoints (PCK)~\citep{MPII}, per vertex error (PVE), and mean acceleration error of every joint in $\mathrm{mm}/\mathrm{s}^2$~\citep{TemporalHMR} for performance evaluation. 

\vspace{-3.0mm}  
\subsection{Performance evaluation and comparisons}

We compare the performance of our SPS-Net with existing frame-based methods~\citep{yang20183d,chen2019unsupervised,kocabas2019self,mehta2017vnect,TCN,wandt2019repnet,CMR,sengupta2020synthetic,omran2018neural,ExPose,zanfir2020neural,HMR,SPIN} and video-based approaches~\citep{TemporalHMR,Temporal3DKinetics,doersch2019sim2real,sun2019human,VIBE}.
Table~\ref{exp:comparison} presents the quantitative results on the 3DPW~\citep{3DPW}, MPI-INF-3DHP~\citep{MPII}, and Human3.6M~\citep{human36m} datasets.

Experimental results on all three datasets show that our method performs favorably against existing frame-based and video-based approaches on the PA-MPJPE, MPJPE, PVE, and PCK evaluation metrics.
However, the acceleration error of our method is inferior to that of the HMMR~\citep{TemporalHMR} approach.
The reason for the inferior performance is that the goal of the HMMR~\citep{TemporalHMR} method lies in predicting \emph{past and future motions} given a single image.
While we have a forecasting module $F$ that predicts the feature of the future frame based on past information, we do not aim to optimize the performance on the human motion prediction task but instead focus on learning to estimate 3D human pose and shape of the \emph{current} frame.
On the other hand, as noted by the VIBE~\citep{VIBE}, the HMMR~\citep{TemporalHMR} method applies smoothing to the predictions, leading to overly smooth pose predictions at the expense of sacrificing the accuracy of pose and shape estimation.

In addition to quantitative comparisons, we present 1) visual comparisons with the VIBE~\citep{VIBE} and SPIN~\citep{SPIN} methods, 2) visual results of occlusion handling, and 3) visual results of different viewpoints.

\heading{Visual comparisons with the VIBE and SPIN methods.}
Figure~\ref{fig:comparison} shows two visual comparisons with the VIBE~\citep{VIBE} and SPIN~\citep{SPIN}.
We observe that our model recovers bodies that well cover humans and estimates more accurate poses for limbs in particular.

\revised{
\heading{Visual results of occlusion handling.}
Figure~\ref{fig:occlusion} presents example visual results of occlusion handling on the CrowdPose dataset~\citep{li2019crowdpose}.
We observe that our model is able to recover plausible human bodies for the occluded person instances, demonstrating the robustness of our SPS-Net.
}

\heading{Visual results of different viewpoints.}
We visualize human bodies recovered by our SPS-Net from different viewpoints in Figure~\ref{fig:mesh}.
Our results show that our method estimates accurate rotation parameters.

\vspace{-3.0mm}  
\subsection{Ablation study}

\heading{Loss functions.}
To analyze the effectiveness of each loss function, we conduct an ablation study by removing one loss function at a time.
Specifically, we analyze how much performance gain each loss function contributes.
Table~\ref{exp:ablation-loss} shows the results on the 3DPW~\citep{3DPW} test set.

Without the camera parameter consistency loss $\mathcal{L}_\mathrm{camera}$, there is no explicit constraint imposed on the prediction of camera parameters, leading to performance drops of $1.7$ in PA-MPJPE and $3.5$ in PVE.
When removing the mask loss $\mathcal{L}_\mathrm{mask}$, our model does not have any constraints to regularize the 3D mesh.
Performance drops of $5.9$ in PA-MPJPE and $7.0$ in PVE occur.
Without the self-supervised parameter regression loss $\mathcal{L}_\mathrm{param}$, our model does not learn to produce plausible predictions when the occlusion or out-of-view issues occur, resulting in performance drops of $5.4$ in PA-MPJPE and $4.6$ in PVE.
When removing the adversarial loss $\mathcal{L}_\mathrm{adv}$, our model does not learn to render 3D meshes that have realistic motions.
Performance drops on all three evaluation metrics occur, which also concur with the findings in the HMR~\citep{HMR} and VIBE~\citep{VIBE}.

Figure~\ref{fig:ablation} presents two visual comparisons with the variant methods of our SPS-Net (i.e., Ours w/o $\mathcal{L}_\mathrm{camera}$ and Ours w/o $\mathcal{L}_\mathrm{mask}$).
Our visual results show that both the camera parameter consistency loss $\mathcal{L}_\mathrm{camera}$ and the mask loss $\mathcal{L}_\mathrm{mask}$ allow our model to predict more accurate pose and shape estimates.

The ablation study on loss functions shows that all four losses are crucial to the SPS-Net.

\begin{table}[t]
  \caption{
  \textbf{Ablation study on loss functions.}
  We report the experimental results on the 3DPW~\citep{3DPW} test set.
  The \first{bold} and \second{underlined} numbers indicate the top two results, respectively.
  }
  \vspace{-6.0mm}
  \begin{center}
    \scriptsize
    \label{exp:ablation-loss}
    %\resizebox{\linewidth}{!} 
    {
    \begin{tabular}{lccc}
      \toprule
      Method & PA-MPJPE $\downarrow$ & MPJPE $\downarrow$ & PVE $\downarrow$ \\
      \midrule
      Ours & \first{50.4} & \first{85.8} & \first{100.6} \\
      Ours w/o $\mathcal{L}_\mathrm{camera}$ & \second{52.1} & \second{88.2} & \second{104.1} \\
      Ours w/o $\mathcal{L}_\mathrm{mask}$ & 56.3 & 90.0 & 107.6 \\
      Ours w/o $\mathcal{L}_\mathrm{param}$ & 55.8 & 89.4 & 105.2 \\
      Ours w/o $\mathcal{L}_\mathrm{adv}$ & 56.2 & 93.4 & 112.5 \\
      \bottomrule
    \end{tabular}
    }
  \end{center}
  \vspace{-10.0mm}
\end{table}

\begin{table*}[t]
  \caption{
  \textbf{Ablation study on the self-attention and forecasting modules.}
  We report the experimental results on the 3DPW~\citep{3DPW} test set.
  The \first{bold} and \second{underlined} numbers indicate the top two results, respectively.
  }
  \vspace{-6.0mm}
  \begin{center}
    \scriptsize
    \label{exp:ablation-model}
    % \resizebox{\linewidth}{!} 
    {
    \begin{tabular}{lccccc}
      \toprule
      Method & \revised{Number of parameters} & PA-MPJPE $\downarrow$ & MPJPE $\downarrow$ & PVE $\downarrow$ & \revised{Acceleration Error $\downarrow$} \\
      \midrule
      Ours & 51.43M & \first{50.4} & \first{85.8} & \first{100.6} & \first{22.1} \\ 
      Ours w/o Forecasting $F$ & 47.23M & \second{54.2} & \second{91.9} & \second{104.3} & 23.3 \\
      Ours w/o Self-Attention $A$ & 34.64M & 57.6 & 96.6 & 104.7 & \second{22.9} \\ 
      \bottomrule
    \end{tabular}
    }
  \end{center}
  \vspace{-8.0mm}
\end{table*}

\heading{Self-attention and forecasting modules.}
We conduct an ablation study to analyze the contribution of the self-attention module $A$ and the forecasting module $F$ in the SPS-Net.
Specifically, we show the contribution of each component by disabling (removing) one at a time.
Table~\ref{exp:ablation-model} shows the results on the 3DPW~\citep{3DPW} test set.
Without either the forecasting module $F$ or the self-attention module $A$, the degraded method suffers from significant performance loss in all metrics.
When both modules are jointly utilized, our model achieves the best results, demonstrating the complementary importance of these two components.

\revised{
Figure~\ref{fig:visual-ablation} shows visual comparisons with the variants of our SPS-Net (i.e., Ours w/o Self-Attention $A$ and Ours w/o Forecasting $F$) on the CrowdPose dataset~\citep{li2019crowdpose}.
Our visual results show that without either the self-attention module $A$ or the forecasting module $F$, the degraded model cannot recover accurate poses.
}

\begin{table}[t]
  \caption{
  \textbf{Ablation study on different temporal modules.}
  We report the experimental results on the 3DPW~\citep{3DPW} test set.
  The \first{bold} and \second{underlined} numbers indicate the top two results, respectively.
  }
  \vspace{-6.0mm}
  \begin{center}
    \scriptsize
    \label{exp:ablation-self-attention}
    \resizebox{\linewidth}{!} 
    {
    \begin{tabular}{lcccc}
      \toprule
      Method & \revised{Number of parameters} & PA-MPJPE $\downarrow$ & MPJPE $\downarrow$ & PVE $\downarrow$ \\
      \midrule
      Ours (Self-Attention) & 51.43M & \first{50.4} & \first{85.8} & \first{100.6} \\ 
      Ours (GRU) & 50.88M & \second{52.8} & \second{87.7} & \second{103.2} \\
      \bottomrule
    \end{tabular}
    }
  \end{center}
  \vspace{-8.0mm}
\end{table}

\heading{Self-attention module vs. GRU.}
To analyze the effectiveness of employing different temporal modules, we conduct an ablation study by swapping the self-attention module $A$ in the SPS-Net with a two-layer GRU module as in the VIBE~\citep{VIBE} model, i.e., comparing the performance between the ``Ours (Self-Attention)'' method and the ``Ours (GRU)'' approach.
Table~\ref{exp:ablation-self-attention} presents the results on the 3DPW~\citep{3DPW} test set.
We observe that employing the self-attention module results in performance improvement over adopting the GRU on all three evaluation metrics.

\begin{table}[t]
  \begin{center}
  \scriptsize
  \caption{
  \textbf{Ablation study of the input sequence length.}
  We present the experimental results on the 3DPW~\citep{3DPW} test set.
  The \first{bold} and \second{underlined} numbers indicate the top two results, respectively.
  }
  \vspace{-2.0mm}
  \label{exp:ablation}
  \centering
  % \resizebox{\linewidth}{!} 
  {
  \begin{tabular}{lccc}
  \toprule
  Input sequence length & PA-MPJPE $\downarrow$ & MPJPE $\downarrow$ & PVE $\downarrow$ \\
  \midrule
  8 & 55.3 & 92.4 & 110.8 \\
  16 & 53.1 & 87.6 & 105.5 \\
  32 & \second{50.4} & \second{85.8} & \second{100.6} \\ 
  48 & \first{50.2} & \first{85.1} & \first{100.2} \\
  \bottomrule
  \end{tabular}
  }
  \end{center}
  \vspace{-8.0mm}
\end{table}

\begin{table}[t]
  \begin{center}
  \scriptsize
  \caption{
  \textbf{Run time analysis.}
  \revised{
  We report the GPU platform, the model training time in hours, and the inference time for processing an image in seconds.
  The ``-'' indicates the result is not available.
  }
  }
  \vspace{-3.0mm}
  \label{exp:inference-time}
   \resizebox{\linewidth}{!} 
  {
  \begin{tabular}{l|ccc}
    \toprule
    Method & Platform & Training & Inference \\
    \midrule
    Yang~\etal~\citep{yang20183d} & Titan X & - & 1.1 \\
    %
    % & Chen~\etal~\citep{chen2019unsupervised} & - & - & - \\
    %
    Mehta~\etal~\citep{mehta2017vnect} & Titan X & - & 3.3 \\
    %
    % & EpipolarPose~\citep{kocabas2019self} & - & - & - \\
    %
    % & TCN~\citep{TCN} & - & - & - \\
    %
    RepNet~\citep{wandt2019repnet} & Titan X & - & 10 \\
    CMR~\citep{CMR} & RTX 2080Ti & - & 3.3 \\
    STRAPS~\citep{sengupta2020synthetic} & RTX 2080Ti & 120 & 0.25 \\
    NBF~\citep{omran2018neural} & V100 & 18 & - \\
    ExPose~\citep{ExPose} & Quadro P5000 & - & 0.16 \\
    HUND~\citep{zanfir2020neural} & P100 & 72 & 0.055 \\
    HMR~\citep{HMR} & Titan 1080Ti & 120 & 0.04  \\
    SPIN~\citep{SPIN} & - & - & 3 \\
    \midrule
    Temporal 3D Kinetics~\citep{Temporal3DKinetics} & - & - & 2 \\
    %
    % & Motion to the Rescue~\citep{doersch2019sim2real} & - & - & - \\
    % %
    % & DSD-SATN~\citep{sun2019human} & - & - & - \\
    % %
    % & HMMR~\citep{TemporalHMR} & - & - & - \\
    %
    VIBE~\citep{VIBE} & RTX 2080Ti & 1 & 0.07 \\
    Ours & V100 & 12 & 0.09 \\
    \bottomrule
  \end{tabular}
  }
  \end{center}
  \vspace{-8.0mm}
\end{table}

\heading{Input sequence length.} 
We conduct an ablation study to analyze the effect of the input sequence length.
Table~\ref{exp:ablation} presents the results on the 3DPW~\citep{3DPW} dataset.
Our results show that the performance on all three metrics improves as the input sequence length increases.
When the input sequence length increases from $32$ (the default setting in our experiments) to $48$, our results can be further improved.
However, due to GPU memory constraints, we are not able to experiment with longer input sequence lengths.

\heading{Sensitivity analysis.}
To analyze the sensitivity of the SPS-Net with respect to the hyperparameters, we perform a sensitivity analysis on the hyperparameters $\lambda_\mathrm{mask}$ and $\lambda_\mathrm{camera}$.
We report the PA-MPJPE results on the 3DPW~\citep{3DPW} test set.
Figure~\ref{fig:hyperparameter} presents the experimental results.

We observe that when the hyperparameter is set to $0$ (i.e., the corresponding loss function is removed), our SPS-Net suffers from performance drops.
When the hyperparameters are set within a suitable range (i.e., around $300$ for $\lambda_\mathrm{mask}$ and around $0.1$ for $\lambda_\mathrm{camera}$), the performance of our SPS-Net is improved, demonstrating the effectiveness of the corresponding loss function.
When the hyperparameters are set to large values (e.g., $1 \times 10^{4}$ for $\lambda_\mathrm{mask}$ and $1 \times 10^{3}$ for $\lambda_\mathrm{camera}$), our model training will be dominated by optimizing the corresponding loss, leading to performance drops.

The sensitivity analysis of hyperparameters shows that when each hyperparameter is set within a suitable range, the performance of our method is improved and remains stable.
\revised{
\vspace{-3.0mm}
\subsection{Run-time analysis}
We report the model training time in hours, the inference time for processing an image in seconds, and the GPU platform used by each method in Table~\ref{exp:inference-time}. 
First, the training time of our method is shorter than that of the HMR~\citep{HMR}, STRAPS~\citep{sengupta2020synthetic}, and HUND~\citep{zanfir2020neural}.
Second, the inference time of our method is comparable to that of the VIBE method~\citep{VIBE}, and shorter than that of the  ExPose~\citep{ExPose} and STRAPS~\citep{sengupta2020synthetic} approaches.
Third, our method performs favorably against existing frame-based and video-based approaches on all three datasets as shown in Table~\ref{exp:comparison}.
}

\vspace{-3.0mm}
\subsection{Failure modes}

\begin{figure}[t]
  \begin{center}
  \mpage{0.234}{\includegraphics[width=\linewidth]{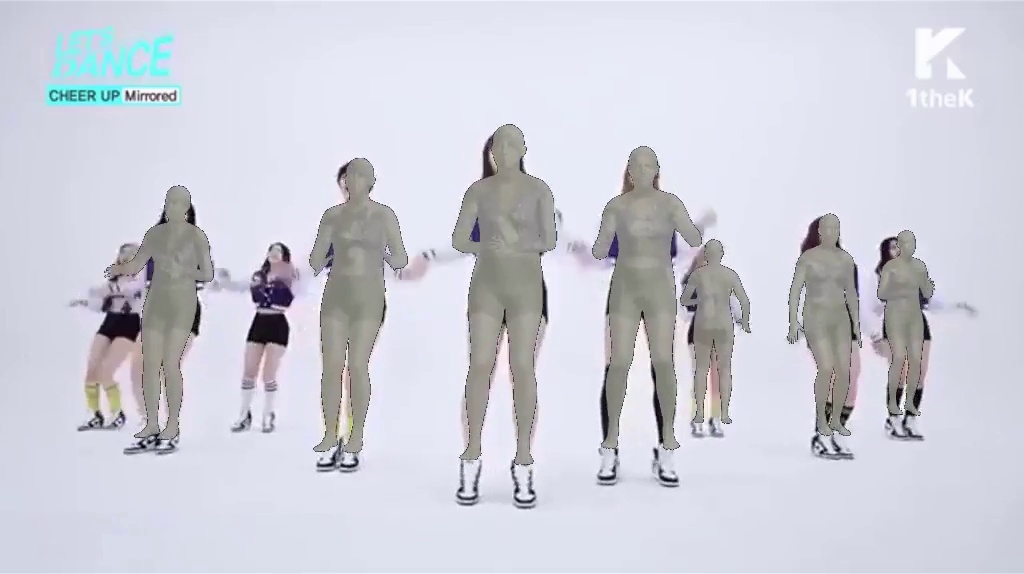}} \hfill
  \mpage{0.234}{\includegraphics[width=\linewidth]{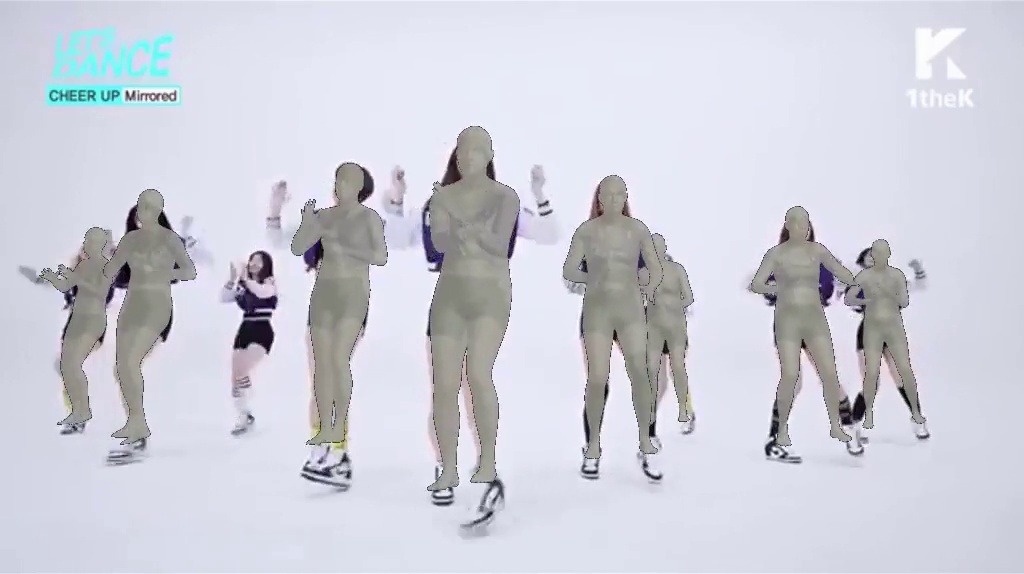}} \\
  \vspace{-1.0mm}
  \caption{
  \textbf{Failure cases.} 
  If missing detection happens, our SPS-Net will not be able to predict 3D human body.
  }
  \label{fig:failure}
  \end{center}
  \vspace{-8.0mm}
\end{figure}

\begin{figure*}[t]
  \begin{center}
  \mpage{0.32}{\includegraphics[width=\linewidth]{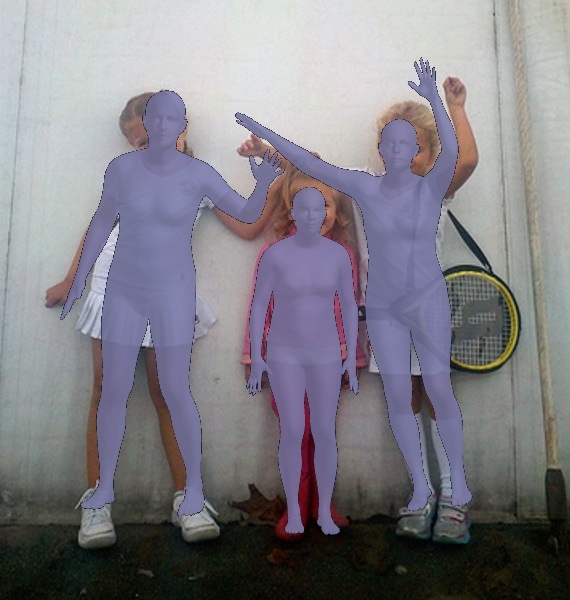}} \hfill
  \mpage{0.32}{\includegraphics[width=\linewidth]{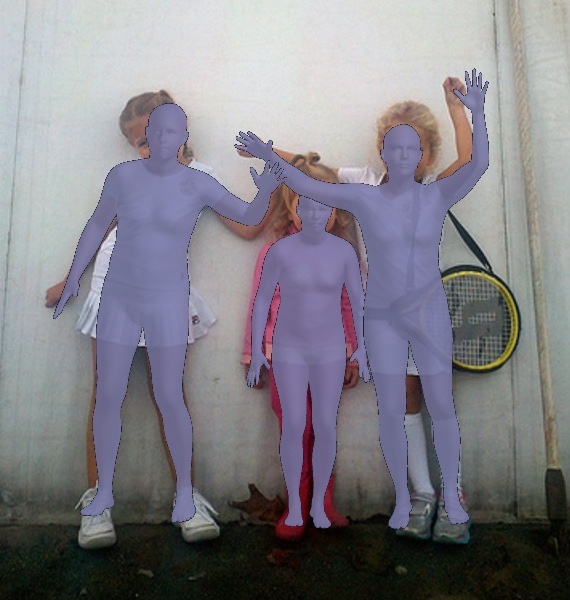}} \hfill
  \mpage{0.32}{\includegraphics[width=\linewidth]{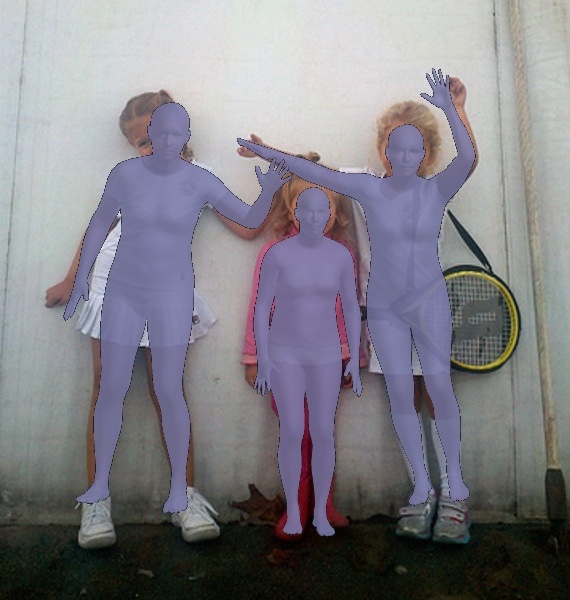}} \\
  \vspace{2.0mm}
  \mpage{0.32}{\includegraphics[width=\linewidth]{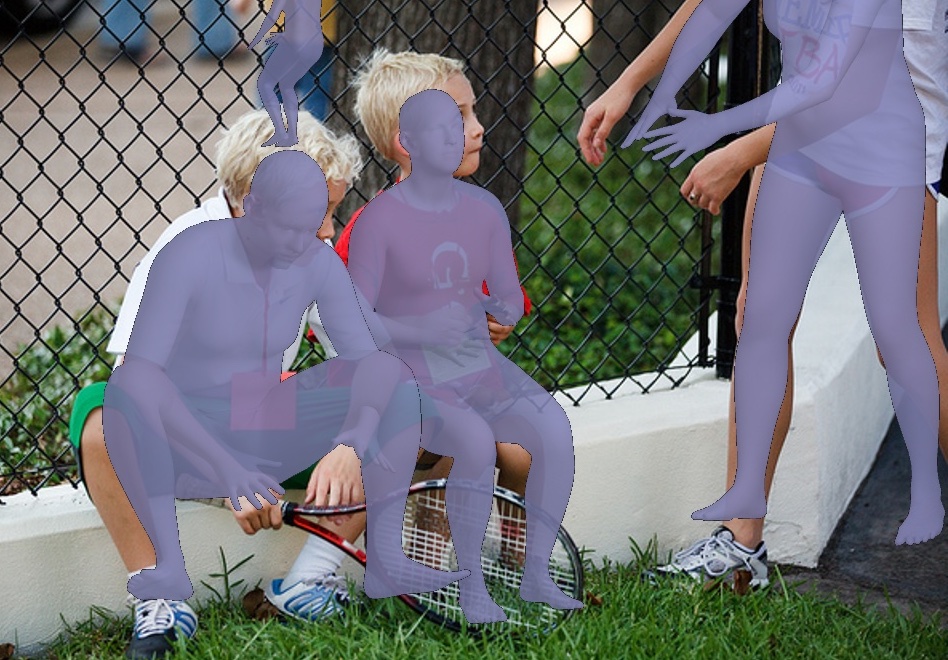}} \hfill
  \mpage{0.32}{\includegraphics[width=\linewidth]{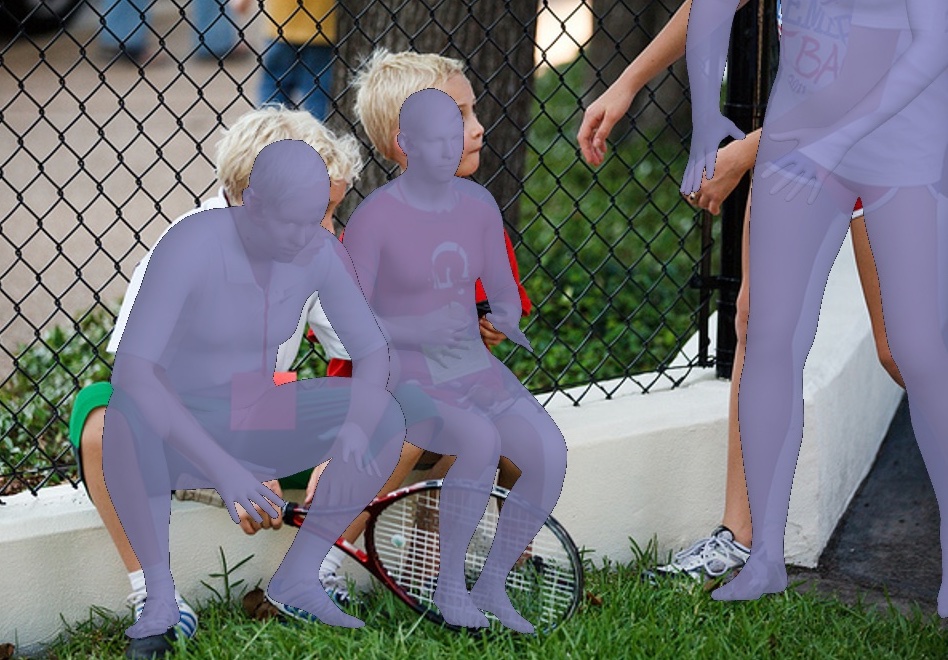}} \hfill
  \mpage{0.32}{\includegraphics[width=\linewidth]{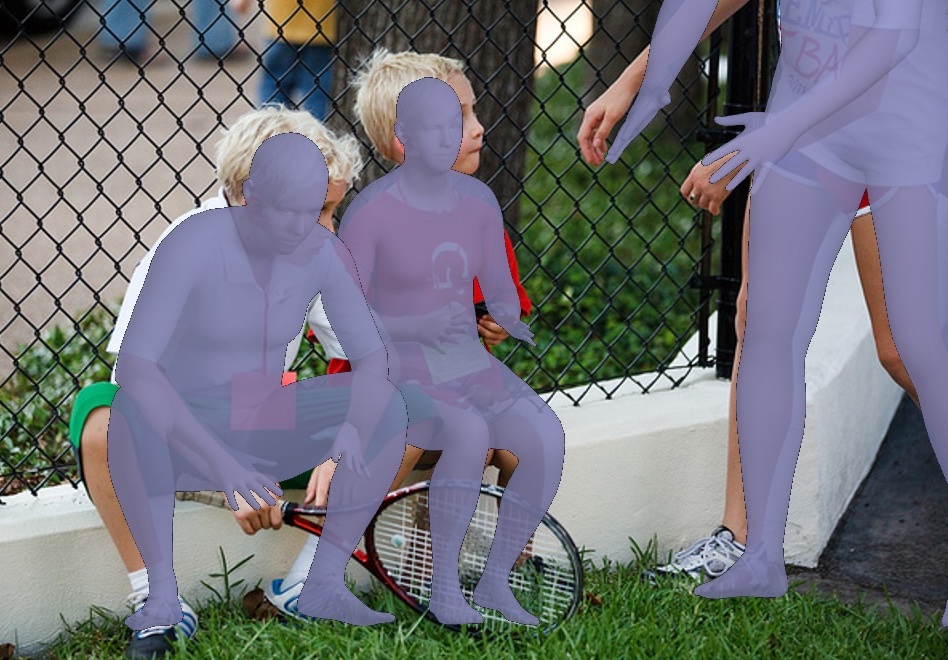}} \\
  \vspace{2.0mm}
  \mpage{0.32}{\includegraphics[width=\linewidth]{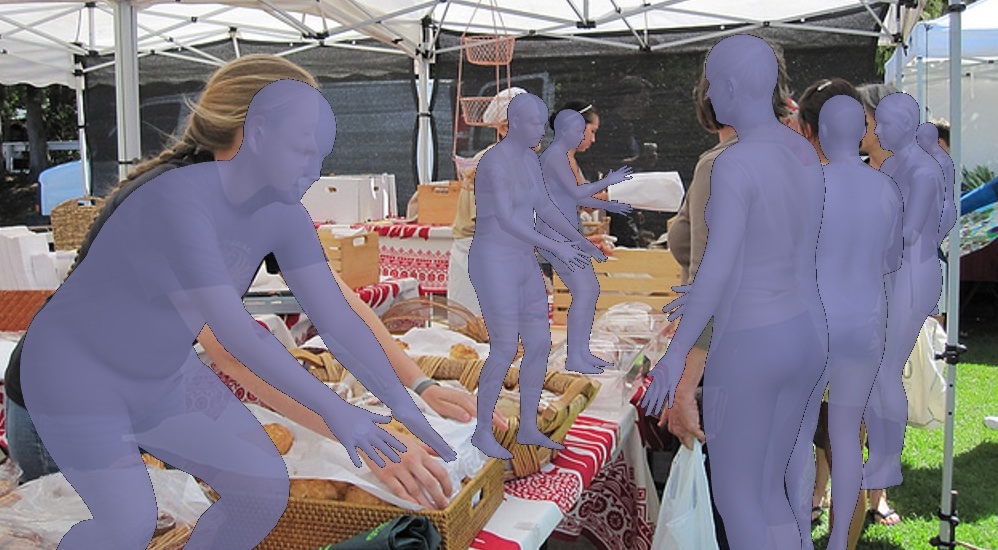}} \hfill
  \mpage{0.32}{\includegraphics[width=\linewidth]{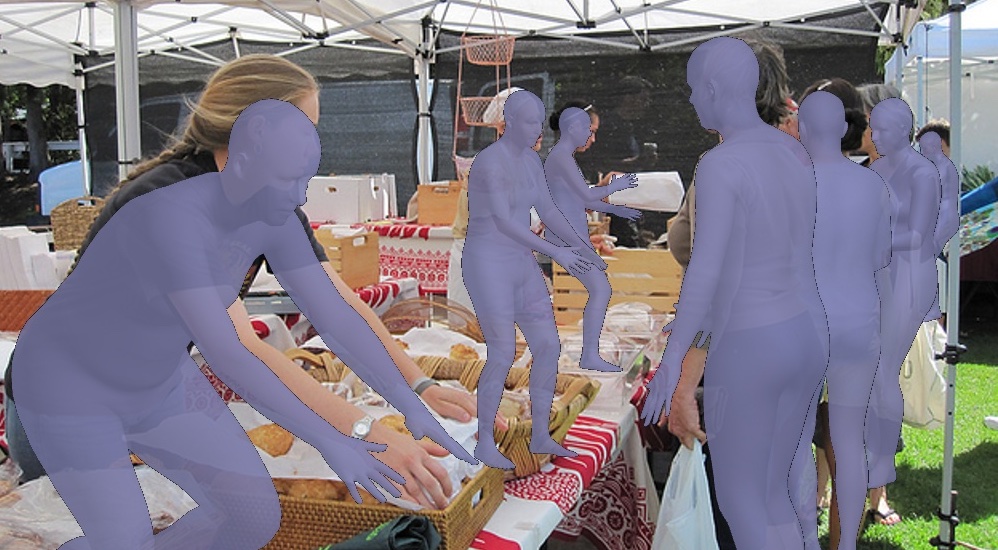}} \hfill
  \mpage{0.32}{\includegraphics[width=\linewidth]{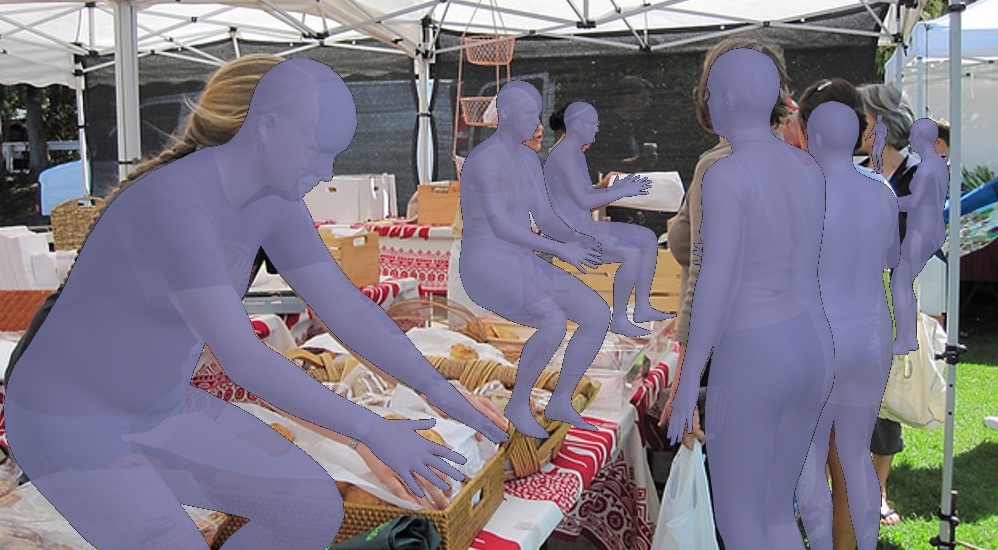}} \\
  \vspace{2.0mm}
  \mpage{0.32}{\includegraphics[width=\linewidth]{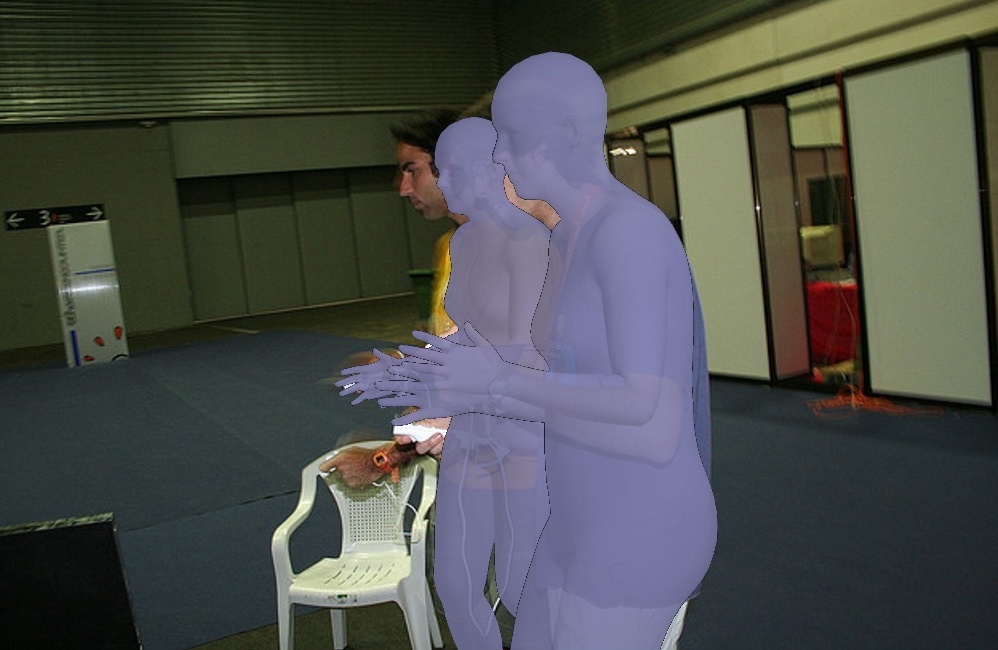}} \hfill
  \mpage{0.32}{\includegraphics[width=\linewidth]{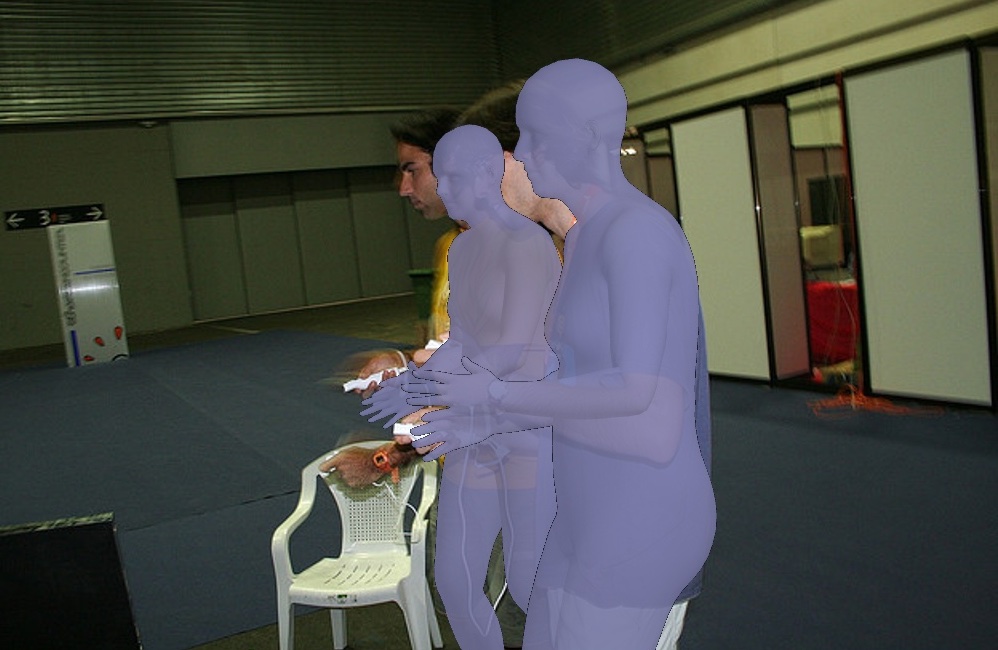}} \hfill
  \mpage{0.32}{\includegraphics[width=\linewidth]{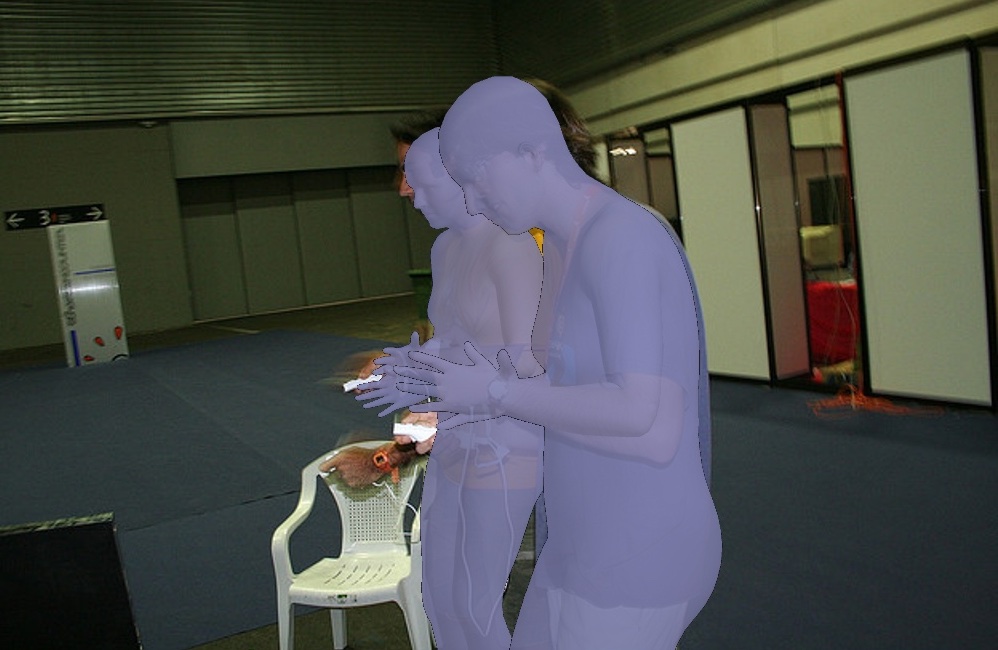}} \\
  \vspace{2.0mm}
  \mpage{0.32}{Ours w/o Self-Attention $A$} \hfill
  \mpage{0.32}{Ours w/o Forecasting $F$} \hfill
  \mpage{0.32}{Ours} \\
  \vspace{-2.0mm}
  \caption{
  \textbf{Visual comparisons with our variant methods.} 
  \revised{
  We present visual comparisons with the Ours w/o Self-Attention $A$ and Ours w/o Forecasting $F$ methods on the CrowdPose dataset~\citep{li2019crowdpose}.
  }
  }
  \label{fig:visual-ablation}
  \end{center}
  \vspace{-7.0mm}
\end{figure*}

We present the failure cases of our method in Figure~\ref{fig:failure}.
As our SPS-Net assumes that the input video frames contain a single person, if missing detection happens, our method will not be able to perform human pose and shape estimation.

\vspace{-3.5mm}
\section{Conclusions}

We propose the SPS-Net for estimating 3D human pose and shape from videos.
The main contributions of this work lie in the design of the self-attention module that captures short-range and long-range dependencies across video frames and the forecasting module that allows our model to exploit visual cues from human motion for producing temporally coherent predictions.
To address the absence of ground-truth camera parameter annotations, we propose a camera parameter consistency loss that not only regularizes the learning of camera parameter prediction but also provides additional supervisory signals to facilitate model training.
We develop a self-supervised learning scheme that explicitly models the occlusion and out-of-view scenarios by masking out some regions in the video frames.
By leveraging the predictions of the original video frames to supervise those of the synthesized occluded or partially visible data, our model learns to predict plausible estimations.
Extensive experimental results on three challenging datasets show that our SPS-Net performs favorably against the state-of-the-art 3D human pose and shape estimation methods.

\bibliographystyle{model2-names}
\bibliography{reference}

\end{document}